\documentclass{article}

     \usepackage[final]{neurips_2025}

\setcitestyle{numbers,square,comma}
\usepackage{pdfpages}
\usepackage[utf8]{inputenc} 
\usepackage[T1]{fontenc}    
\usepackage{hyperref}       
\usepackage{url}            
\usepackage{booktabs}       
\usepackage{amsfonts}       
\usepackage{nicefrac}       
\usepackage{microtype}      
\usepackage{enumitem}
 \usepackage{comment}
 \usepackage{tabularray}
 \usepackage{tabularx}
 \usepackage{pifont}
 \usepackage{verbatim}
 \usepackage{amsmath}
 \usepackage{mdframed}

\usepackage[dvipsnames]{xcolor}
\usepackage{multirow}

\definecolor{set1red}{HTML}{C44E52}
\definecolor{set1blue}{HTML}{4C72B0}
\definecolor{set1green}{HTML}{55A868}
\definecolor{set1orange}{HTML}{DD8452}
\definecolor{set1purple}{HTML}{8172B3}
\definecolor{set1yellow}{HTML}{CCB974}
\definecolor{set1white}{HTML}{FAF9F6}

\newcommand{\xmark}{\ding{55}}%

\usepackage{pgf-pie}
 \usepackage[table,xcdraw,dvipsnames]{xcolor}
\usepackage{subcaption}

\definecolor{osured}{HTML}{CE5140}
\definecolor{osublue}{HTML}{6385EC}
\definecolor{iccvblue}{rgb}{0.21,0.49,0.74}
\title{Reading Recognition in the Wild}

\author{Charig Yang\textsuperscript{1,2}, Samiul Alam\textsuperscript{3}, Shakhrul Iman Siam\textsuperscript{3}, Michael J. Proulx\textsuperscript{1}, Lambert Mathias\textsuperscript{1}, \\ \bf Kiran Somasundaram\textsuperscript{1}, Luis Pesqueira\textsuperscript{1}, James Fort\textsuperscript{1}, Sheroze Sheriffdeen\textsuperscript{1}, Omkar Parkhi\textsuperscript{1}, \\ \bf
Carl Ren\textsuperscript{1}, Mi Zhang\textsuperscript{3}, Yuning Chai\textsuperscript{1}, Richard Newcombe\textsuperscript{1}, Hyo Jin Kim\textsuperscript{1} \\ [5pt]
\textsuperscript{1}Meta Reality Labs Research \quad \textsuperscript{2}VGG, University of Oxford  \quad \textsuperscript{3}The Ohio State University
 \\ 
 \texttt{charig@robots.ox.ac.uk, kimhyojin@meta.com} \\ [5pt]
\href{https://www.projectaria.com/datasets/reading-in-the-wild/}{\texttt{https://www.projectaria.com/datasets/reading-in-the-wild/}} 
} 

\usepackage{titlesec}

\titlespacing{\section}
  {0pt}    
  {2pt}    
  {2pt}    
\titlespacing{\subsection}
  {0pt}    
  {2pt}    
  {2pt}    
\usepackage[font=footnotesize]{caption}

\begin{document}

\maketitle
\begin{abstract}
To enable egocentric contextual AI in always-on smart glasses, it is crucial to be able to keep a record of the user's interactions with the world, including during reading.
In this paper, we introduce a new task of \emph{reading recognition} to determine \emph{when} the user is reading.
We first introduce the first-of-its-kind large-scale multimodal \emph{Reading in the Wild} dataset, containing 100 hours of reading and non-reading videos in diverse and realistic scenarios.
We then identify three modalities (egocentric RGB, eye gaze, head pose) that can be used to solve the task, and present a flexible transformer model that performs the task using these modalities, either individually or combined.
We show that these modalities are relevant and complementary to the task, and investigate how to efficiently and effectively encode each modality.
Additionally, we show the usefulness of this dataset towards classifying types of reading, extending current reading understanding studies conducted in constrained settings to larger scale, diversity and realism. 
\end{abstract}
\begin{figure}[h]
    \centering
\includegraphics[width=0.38\textwidth]{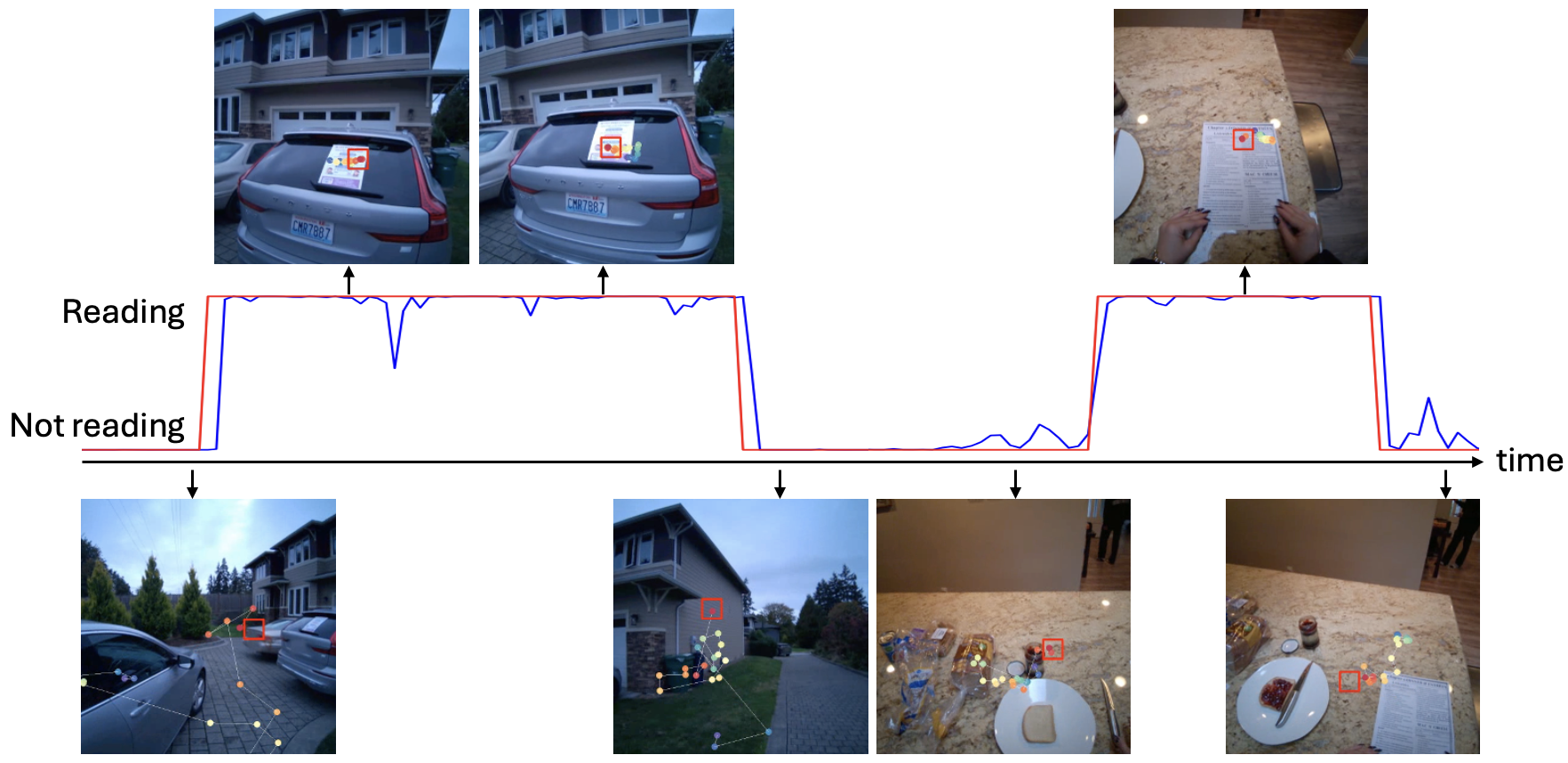}
\includegraphics[width=0.58\textwidth]{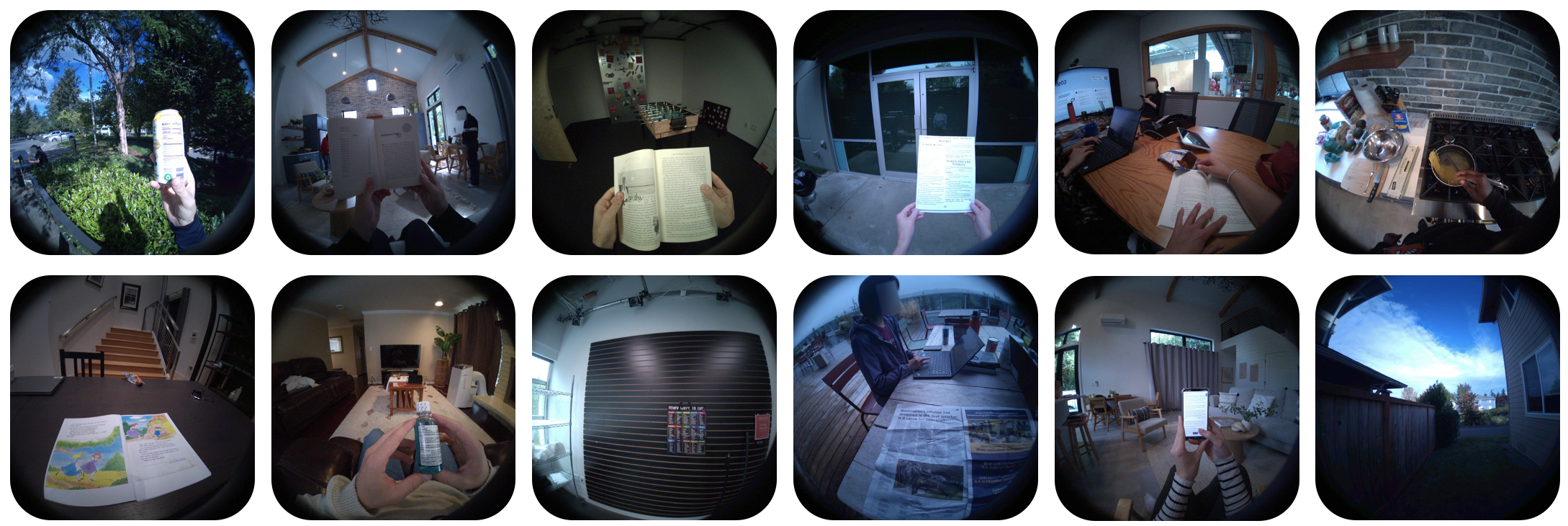}
    \caption{\textbf{Am I reading?} The left figure shows a timeline as the user navigates the world. We aim to solve the task of reading recognition to enable AI assistants in always-on wearables. We identify three modalities: eye gaze (in colored dot patterns), RGB crop around gaze (in red box), and inertial sensors performs the task to high accuracy (with \textcolor{blue}{Prediction} and \textcolor{red}{GT} shown).
    Images from our \textit{Reading in the Wild} dataset, which features 100 hours of diverse reading and non-reading activities in real-world settings, with examples shown in the right. }
    \label{fig:teaser}

\end{figure}
\section{Introduction}
\vspace{-0.1cm}
\label{sec:intro}


The potential future of AI personal assistants depends on its ability to understand the physical context of the user. Smart glasses are becoming a promising device form factor capable of linking visual AI capabilities to the real world. Recently, there has been a sharp rise in the development of smart glasses, both products (Meta Ray-Ban, Amazon Echo Frames) and prototypes (Snapchat Spectacles, Halliday AI Glasses, Xreal One Pro). These all-day wearable devices enable proactive, personalized, and contextualized AI agents to perceive the world like humans do by understanding the users' context.

However, for always-on wearable glasses, due to both \textit{hardware} (power, bandwidth, heat) and \textit{software} (perception capability of AI agents, especially with heavy models) constraints, it is impractical to record and process every single frame over long periods of time. One solution is to have a proxy signal, so that the device can record and process \textit{key} frames only \textit{when} relevant. The question becomes: what forms important context of the user that the AI assistant needs to know, and how do we know when to capture them?

The ability to read underpins one of, if not the most important unique modalities by which modern humans communicate, entertain each other, and learn. Reading is a key mechanism humans use to communicate with high fidelity and high information density. Reading spans a broad array of mediums, from handwritten and printed text on paper and digital displays to environmental signposts. The act of reading occurs within real-time communication with one another and today's AI chatbots, through to reading long-form articles in books or online. Enabling AI with the ability to recognize reading is hence clearly one of the most important context signals a future AI can be enabled with to unlock truly personalized and contextually relevant AI.

Given this, we ask: how can we provide future AI with the ability to know when someone is reading? This apparently simple idea underpins the ability to efficiently enable devices to know what the user has and has not read, and hence where they can assist given what it understands that the user has read. 

This task of \textit{reading recognition} is challenging for two main reasons. First, the problem can often be ill-posed: just because a text exists in the field of view does not mean that the user is reading it (or even looking at it), which is ambiguous to solve using visual information alone. Also, the method should be efficient for real-time, always-on computation subject to the practical constraints of a wearable device. Both of these challenges render OCR-based text detection methods impractical, given the inability to solve the ambiguity and the requirements for high-resolution capture and processing. Instead, reading recognition can be used as an efficient proxy to indicate \textit{when} and \textit{where} it is relevant to invoke heavier models (OCR and VLMs) instead of running these models all the time.

Motivated by this question, we introduce a new dataset created with Project Aria~\cite{project_aria} glasses, which enables us to develop the contextual AI capability of detecting \textit{when} a wearer is reading. We present the first-of-its-kind large-scale multimodal "Reading in the Wild" dataset, containing 100 hours of reading and non-reading videos in diverse and realistic scenarios. This dataset allows us to identify three modalities (egocentric RGB, eye gaze, head pose) that can be used to solve the task. We then present a flexible transformer model that performs the task using these modalities, either individually or combined. We show that these modalities are relevant and complementary to the task and investigate how to efficiently and effectively encode each modality, as well as the model's ability to generalize towards unseen scenarios and perform real-time reading detection. 


Achieving reading recognition makes it feasible to keep a record of a user's reading interactions with the world to build a contextually aware AI. It also enables several other applications: it allows reading assistant tools \cite{thaqi2024sara} in children with learning difficulties \cite{caldani2020visual} and people with low vision \cite{wang2024gazeprompt} to operate in the real world; it can also be used to track whether a user has read crucial information (\textit{e.g.} signs during driving) and to measure attention and distraction while performing a task.

Additionally, the dataset and method contributed in this paper can be extended to classifying different types of reading. This has been of interest in cognitive studies in reading comprehension, but they are often limited to controlled environments \cite{kunze2013know, landsmann2019classification, kelton2019reading, ahn2020towards, castilla2024improving}, hence limiting its usefulness. We show that our dataset allows for reading mode and medium classification to be performed in unconstrained settings, and provide experimental results in this direction.

In summary, we make the following contributions:
\begin{itemize}[left=0pt, nosep]
    \item First, we introduce a new task of reading recognition \textit{in the wild}, and demonstrate its usefulness. Unlike previous studies, we focus on in-the-wild settings and practicality towards wearable glasses.
    \item Second, we present the first-of-its-kind large-scale egocentric multimodal \textit{Reading in the Wild} dataset, which will be made publicly available, alongside a scalable protocol for data collection.
    \item Third, we identify three modalities relevant and complementary to the task (RGB, gaze, and IMU), and develop a lightweight, flexible model that inputs these modalities either individually or in combination for reading recognition, resulting in a strong and efficient baseline for this task.
    \item Fourth, we show that our method and dataset extend towards reading understanding, including classifying reading mode and medium, demonstrating usefulness towards cognitive studies.
\end{itemize}

\section{Related Work}
\vspace{-0.1cm}
\noindent\textbf{Reading recognition} has been a long studied task with rich literature. Eye gaze has been used as the primary signal \cite{kelton2019reading, campbell2001robust, ahn2020towards, landsmann2019classification}, however, it relied on handcrafted feature engineering methods such as detecting fixations and saccades, which we show are unnecessary. Moreover, the experiments are usually constrained, and not performed \textit{in the wild}.
Other modalities have also been considered, such as electrooculography (EOG) signals \cite{bulling2008robust}, though the usage of electrodes can be invasive and hence less practical towards building user-friendly wearable glasses. In this paper, we steer this towards practical usage in modern smart glasses, where we show that gaze can be used in combination with visual information and IMU sensors.
With recent advances in wearable devices, reading recognition expands to tasks such as word recognition and reading order prediction \cite{jahagirdar2024icdar}. While this is relevant, it concerns the reading content, and assumes the user is already reading, which differs from the task of detecting whether the user is reading in this paper.
Applications include reading comprehension \cite{meziere2023using, just1980theory, copeland2014predicting, rataj2018understanding}, understanding user behavior \cite{castilla2024improving}, and in building reading assistants \cite{thaqi2024sara,caldani2020visual,wang2024gazeprompt}. However, the literature is largely constrained to controlled environments. 


\noindent\textbf{Egocentric activity recognition} is a popular vision task that usually require computationally heavy solutions using video input \cite{Kazakos_2019_ICCV, zhao2022lavila}.
In terms of data, reading is only a subset of activities in some common datasets such as EGTEA Gaze+ \cite{li2018eye} and Ego-Exo4D \cite{grauman2024ego}. However, not all datasets contain reading \cite{Damen_2018_ECCV}. For those which include reading, its nature is very restricted to activities such as reading recipes (in \cite{li2018eye}), reading covid testing manuals, climbing instructions, and music sheets (in \cite{grauman2024ego}). Ego4D  \cite{Grauman_2022_CVPR} offers a more diverse range of reading activities, but only less than 1\% of the data includes eye gaze. 
In contrast, our paper focuses on efficient reading recognition, and the proposed dataset contains large-scale and diverse reading and non-reading examples with eye gaze information.

\noindent\textbf{Gaze in computer vision} has started to gain popularity, where gaze has many applicable uses. One popular route is to perform gaze prediction \textit{i.e.} predicting where the user is looking at \cite{ozdel2024transformer, fang2024oat, mondal2024look,strohm2024learning} or how the user interacts with objects he/she observes \cite{jin2024boosting, tian2024gaze, lin2024gazehta}. In medical applications, eye gaze can be used as a saliency test to ensure integrity in medial image analyses \cite{wu2024gaze, kong2024gaze, kim2024enhancing, liu2024gem}, as well as predicting learning disorders \cite{islam2024involution}.
Recently, gaze has also been used to complement vision, such as in action recognition \cite{zhang2022can}, narration \cite{chen2024gazexplain}, and vision-language models \cite{konrad2024gazegpt}. Our paper further explores whether gaze can \textit{reduce} the input requirements for computer vision models by only using gaze and/or parts of vision that are associated with gaze instead of using the whole image sequence.

\begin{table*}[]
\setlength{\tabcolsep}{3pt}
\centering \tiny
\begin{tabular}{l|l|ll|lllll|ll} \toprule
Subset &
  Size &
  Indoor & Outdoor &
  Medium &
  Text type &
  Multi-task &
  Mode &
  Language &
  Not reading &
  Mixed \\ \midrule
\textbf{Seattle} &
  80 hours &
  Offices & Balconies &
  Print &
  Paragraphs &
  None &
  Engaged &
  English ($\rightarrow$) &
  Daily activities &
  Alternating  \\
  (train/val/test)
 &
  81 people &
  Libraries & Patios &
  Digital &
  Short texts &
  Walking &
  Skimming &
   &
  Hard negatives &
  sequences \\
  \textbf{Focus}: diversity
 &
  1061 videos &
  Homes & Roads/trails &
  Objects &
  Non-texts &
  Writing &
  Scanning &
   & (71\%/29\%)
   &
   (reading /\\
 &
   &
 Stores & In the woods &
   & Dynamic texts 
   & Typing
   &
  Out loud &
   &
   &not reading)
   \\ \midrule
\textbf{Columbus}  &
  20 hours &
  Offices &  &
  Print &
  Paragraphs &
   None &
  Engaged &
  English ($\rightarrow$) &
  Hard negatives &
  Mirror setups  \\
(test)&
  31 people &
   Libraries & &
  Digital &
  Short texts &
   &
  Scanning  &
  Bengali ($\rightarrow$) &
  Daily activities & (same settings, 
   \\
\textbf{Focus}: edge cases,  &
  655 videos & Lounges & 
   &
  Objects &
  Non-texts &
   &
   &
  Chinese ($\downarrow$) & (58\%/42\%)
   &
   one reading,\\
generalization &
   & Corridors
   & 
   &
   & &
   &
   &
  Arabic ($\leftarrow$) &
   & another not)\\ \bottomrule
  
\end{tabular}
\vspace{-5pt}
\caption{\textbf{Dataset overview.} We separately collect two subsets for the dataset. Seattle subset focuses on diversity, while Columbus subset looks at the model's generalization towards unseen settings, as well as edge cases where the model fails. See Appendix A for more details.}
\vspace{-0.6cm}
\label{tab:subset}
\end{table*}
\section{Reading in the Wild Dataset}

\vspace{-0.1cm}

\subsection{Overview}

The dataset contains about $100$ hours of recordings of reading and non-reading activities collected from one RGB ($30$Hz, $1408$p, $110$° FoV) and two SLAM ($150$° FoV) cameras, two eye tracking cameras ($60$Hz, calibrated), two IMUs (with odometry outputs from visual SLAM), and audio transcribed using WhisperX \cite{bain2023whisperx}. We independently collect two subsets of this dataset, as in Table \ref{tab:subset}.

\noindent\textbf{Seattle} is collected for training, validation, and testing. We mainly focus on collecting reading and non-reading activities in diverse scenarios, in terms of participants' identities, reading scenarios, reading modes, and reading materials. It contains a mix of normal and hard examples, as well as mixed sequences alternating between reading and non-reading activities. The dataset is collected in homes, office spaces, libraries, and outdoors. 

\noindent\textbf{Columbus} is collected to find out where the model breaks in zero-shot experiments. It contains examples of hard negatives (where text is present but is not being read), searching/browsing (which gives confusing gaze patterns), and reading non-English texts (where reading direction differs). 

\begin{figure*}[t]
    \centering
    \begin{minipage}{0.58\textwidth}
        \centering
        \footnotesize
        \begin{tabular}{l|cc|ccc} \toprule
            Dataset & Gaze  & RGB & Reading & Real & HN \\ \midrule
            Ego4D  & \xmark & \checkmark & Limited & \checkmark & \xmark \\
            Ego-Exo4D & 10Hz & \checkmark & Limited & \checkmark & \xmark \\
            EGTEA & 30Hz & \checkmark & Limited & \checkmark & \xmark \\\midrule
            ZuCo  & 500Hz & \xmark & \checkmark & \xmark & \xmark \\
            InteRead  & 1.2kHz & \xmark & \checkmark & \xmark & \xmark \\\midrule
            Ours & 60Hz & \checkmark & \checkmark & \checkmark & \checkmark \\\bottomrule
        \end{tabular}
        \captionof{table}{\textbf{Comparison to existing datasets}. Our dataset is the first reading dataset that contains high-frequency eye-gaze, diverse and realistic egocentric videos, and hard negative (HN) samples.}
        \label{tab:data-compare}
        \vspace{-5pt}
    \end{minipage}
    \hfill
    \begin{minipage}{0.4\textwidth}
        \centering
        \includegraphics[width=\textwidth]{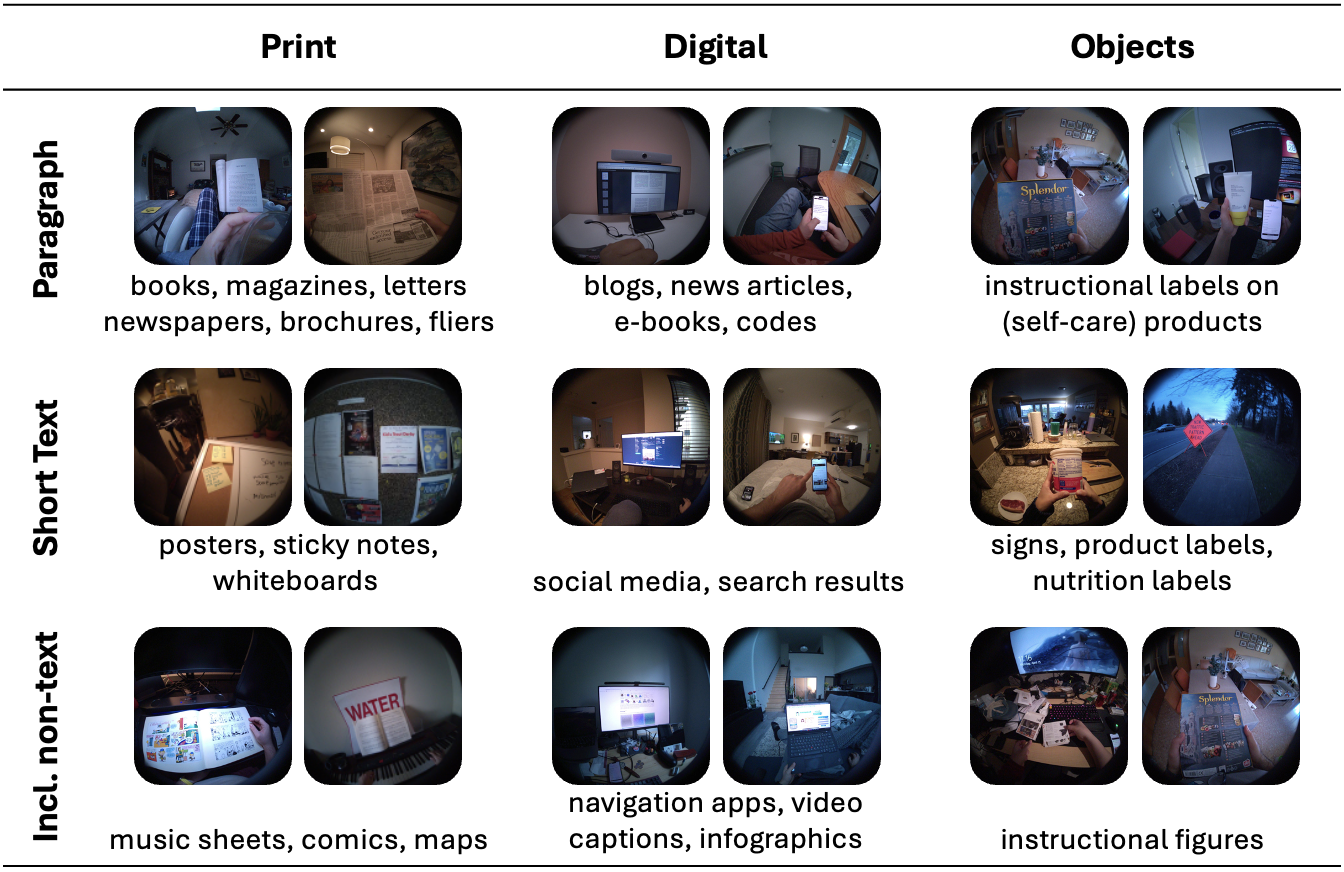}
        \vspace{-12pt}
        \caption{\textbf{Diversity in reading materials.} Reading examples across different materials, both text type (rows) and medium (column).}
        \label{fig:variety}
    \end{minipage}
    \vspace{-0.1cm}

\end{figure*}

\subsection{Comparison to existing datasets}

The closest kins to our dataset come in two categories, as shown in Table \ref{tab:data-compare}. First, in egocentric video datasets \cite{grauman2024ego,Grauman_2022_CVPR,li2018eye}, there are very limited reading sequences and they lack diversity as each dataset only reads from 1-2 examples (COVID test kits for Ego-Exo4D, recipes for EGTEA). Moreover, their eye tracking frequencies are also limited. 
Second, there are cognitive studies that focus on human gaze behavior during reading \cite{hollenstein2018zuco, zermiani2024interead} with high-frequency eye tracking. However, these studies are conducted in very constrained scenarios such as reading a text in front of a screen.
Moreover, these studies only collect gaze data without RGB stream.

\subsection{Contents}
\label{sec:contents}
\noindent\textbf{Reading.} Our dataset presents a large diversity in reading activities, including:

\begin{itemize}[left=0pt, nosep]
    \item \textbf{Reading mode}: Our dataset contains different reading modes, including deep reading (careful, engaged reading), skimming (quickly glancing through for general ideas), scanning (searching for specific information), and reading aloud (verbalizing the text).
    \item \textbf{Single/Multi-task reading}: Our dataset not only covers single-task reading, where the focus is solely on the reading material, but also reading while multitasking, such as reading while writing, typing, or walking.
    \item \textbf{Medium and text type}: We collect data across mediums: print (books, newspapers, flyers), digital (phones, monitors), everyday objects (product labels, whiteboards); and text types: paragraphs, short texts, non-texts, and dynamic texts (video captions and subtitles) as illustrated in Figure \ref{fig:variety}.
    \item \textbf{Demographics}: We collect data among $111$ participants and include their age range and gender.
    \item \textbf{Location}: For diversity, we collect scenes across indoor (e.g., meeting rooms, bedrooms, living rooms), balconies, outdoors, and in the woods.
\end{itemize}

\noindent\textbf{Non-reading.} 
We also collect negative examples. This includes \textit{Everyday activities} that do not involve reading such as physical exercise, outdoor activities, creative arts, culinary activities, and household chores, as well as \textit{Hard negatives}, where text is present in the scene but is not being read, which would confuse RGB-only models. 

\noindent \textbf{Mixed}. We also collect \textit{Alternating sequences}, where the participants alternating between reading and non-reading with annotated timestamps,
and \textit{Mirror setups} where we have the same participant perform reading and non-reading activity in the same environment and the same material.

\subsection{Data collection process}


\noindent\textbf{Logistics.}
We recruited a total of $111$ participants, targeting a uniform distribution for gender and age. We gave each participant a list of tasks to record, with moderators monitoring to ensure that the recordings are correct as desired.

\noindent\textbf{Instructions.} We divided the collections into tasks, each with specific instructions, as elaborated in the Appendix. We also asked the participants to perform eye gaze calibration within each recording.

\noindent\textbf{Privacy.}
We strictly followed Project Aria Research guidelines.
All data has been de-identified, and faces and license plates were anonymized with EgoBlur \cite{raina2023egoblurresponsibleinnovationaria}. We source the venues ourselves do not use the participants' private spaces to prevent exposure of sensitive or identifiable information. 


\noindent\textbf{Scalable Protocol through Automatic labeling.} In addition to the dataset itself, we also present a protocol for scalable, high-quality data collection. Instead of manually labeling the timestamps, we instruct the participants to say ``start reading!" whenever they start reading, and ``finished reading!" whenever they finish. In doing so, we can simply use WhisperX \cite{bain2023whisperx} to obtain accurate timestamps without requiring manual annotations.

\noindent\textbf{Quality assurance.} We have several protocols to ensure that participants have read the text. This involves before (pre-reading questions) and after (post-reading questions and summarization). For the subset where the user reads out loud, the audio transcription can also be used to for quality assurance.

\section{Method}
\vspace{-0.1cm}
\subsection{Task definition}

Formally, at time $t$, we want to predict the confidence score $s_t \in [0, 1]$ whether the user is reading or not, given several input modalities: eye gaze patterns $g_{t-T \leq \tau \leq t} \in \mathbb{R}^{f \times T \times d}$, instantaneous RGB $I_t \in \mathbb{R}^{H \times W \times C}$ and head pose (IMU) sensor readings $z_{t-T \leq \tau \leq t} \in \mathbb{R}^{f \times T \times d}$, where $f$ is the sampling frequency and $T$ is the input duration \textit{i.e.} 
$
s_t = \Phi(g_{t-T \leq \tau \leq t}, I_t, z_{t-T \leq \tau \leq t}).
$
Each modality has different advantages and drawbacks. To harness the strength of all modalities, we propose a multimodal model that takes into account all three modalities as input. In the following sections, we first discuss individual modalities, followed by the model architecture.

\subsection{Input modalities}

\noindent\textbf{Gaze. 
}
There exists a vast literature suggesting that gaze can be used to detect reading activity without visual information \cite{kelton2019reading, campbell2001robust, ahn2020towards, landsmann2019classification}. However, their experiments are limited to constrained environments (reading long paragraphs in front of a screen), and they rely on feature engineering methods such as fixation detection to circumvent small-scale data. As we demonstrate in the experiments section, training on diverse data translates well to open-world settings, and feature engineering is unnecessary at scale, which makes it robust to low frequency eye tracking inputs.


\begin{figure*}[t]
    \begin{minipage}{0.41\textwidth}
        \centering
        \includegraphics[width=\textwidth]{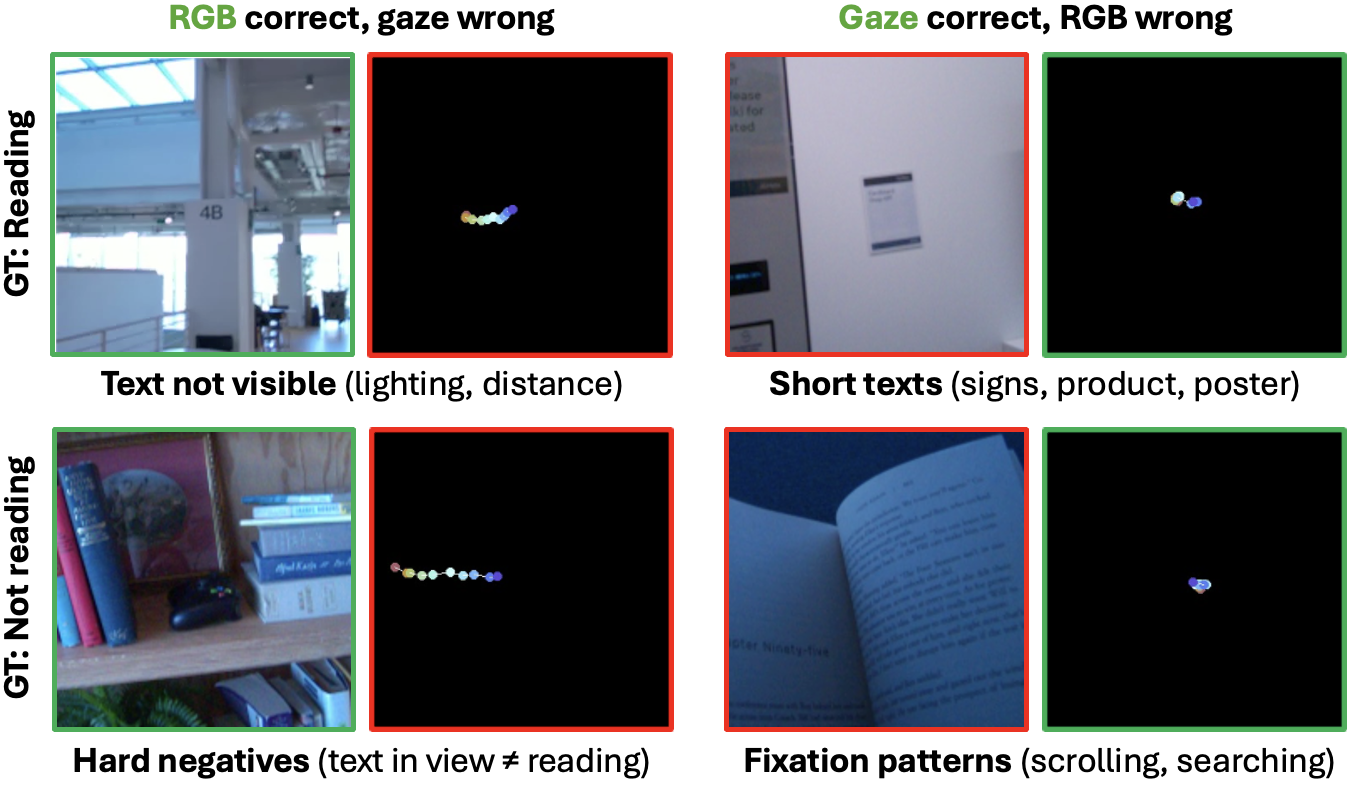}
        \caption{\textbf{Complementary modalities}. Example success and failure cases for gaze and RGB, suggesting the benefit of multimodality.}
        \label{fig:complementary}
    \end{minipage}
    \hfill
\begin{minipage}{0.56\textwidth}
  \centering
  \vspace{-3pt}
  \includegraphics[width=\textwidth]{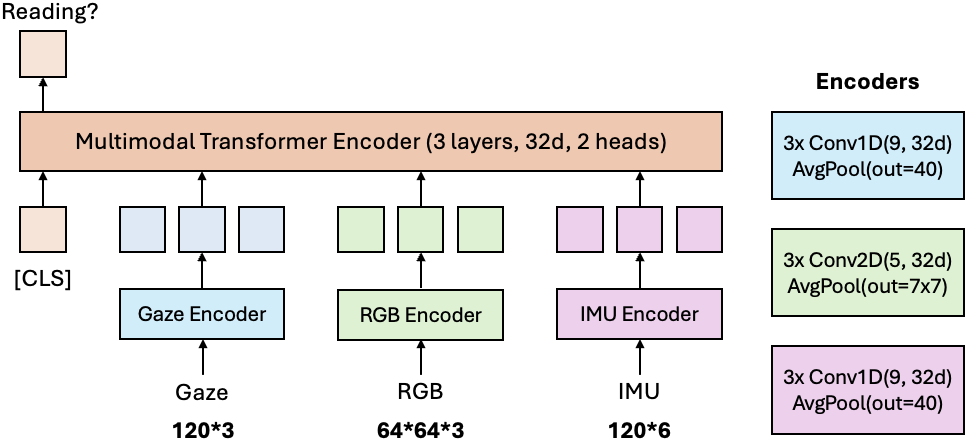}
  \caption{\textbf{Model architecture}. Our model is a simple transformer encoder with any combination of gaze, RGB, and IMU as input.}
  \label{fig:transformer_arch}
    \end{minipage}
    \vspace{-1cm}
\end{figure*}

\noindent\textbf{RGB. 
}
As with action recognition methods, visual information has been an effective cue in the computer vision community. However, processing video models on a wearable device is expensive.
Meanwhile, there has been an interest in using gaze to guide model attention in action recognition \cite{li2018eye, grauman2024ego,zhang2022can}. For reading, we argue that region outside the gaze point is likely to be irrelevant, as the high-resolution human fovea capable of reading only covers a small region (2°) around the gaze \cite{thumb}. Therefore, we only crop the image around the gaze region. 
This also allows for large efficiency gains as capture and processing only needs to be done on a small patch.
We find that cropping using only 1/484 of an image (64px, 5° from 110° FoV) can result in good accuracy, with the remainder for context and gaze uncertainties.

\noindent\textbf{Head pose (IMU). 
}
We also explore using odometry measurements.
While not a good indicator on its own, we find that it helps as a secondary sensor. The intuition here is that some inertial motions can be used to address ambiguities, such as distinguishing between reading and horizontal head motion. 

\noindent\textbf{Complementary modalities.} The main reason for using multiple modalities is that they are complementary: they excel and fail in different places. For example, eye gaze can perform well even if the text is not visible due to lighting or distance that images sometimes miss out, while RGB works in cases where gaze patterns are not obvious, such as when reading short texts like signs, as shown in Figure \ref{fig:complementary}. While IMU is not strong on its own, we show later that it further provides cues to disambiguate some cases (e.g. turning heads vs reading).

\subsection{Model}

In order for this to be practical towards always-on wearable devices, we propose a simple and efficient model that achieves a strong practical baseline for the task. Particularly, we propose a flexible multimodal transformer model that takes in different modalities as input, as shown in Figure \ref{fig:transformer_arch}. By keeping the model simple, we can investigate different combinations and forms of modalities.

\noindent\textbf{Input.} Unless otherwise stated (such as in ablation studies), we use $T=2$, $f=60$ for 3D eye gaze and 6DoF IMU, and a 5° FoV ($H,W=64$) crop for RGB as default.

\noindent\textbf{Modality encoder.} 
The model consists of different encoders $\Phi_{\{g, r, i\}}$ (where g,r,i represent gaze, RGB, and IMU respectively) to tokenize individual modality into feature tokens $f_{\{g,r,i\}}\in \mathbb{R}^{N\times D}$. We use three layers each of 1D (gaze and IMU) and 2D (RGB) convolutions.

\noindent\textbf{Multimodal transformer.} We then combine these feature tokens using a simple transformer encoder $\Phi_t$ and a linear head over the [CLS] token \textit{i.e.} 
$
s_t = \Phi_t(f_g, f_r, f_i).
$

\noindent\textbf{Modality dropout.} During training, we dropout entire modalities at random, which serves two purposes: (i) it helps with training less-used modalities; (ii) during inference, the model can perform well even without all modalities being present.

\subsection{Generalization}

While we train on English texts, we find that our model generalizes well to other left-to-right languages across different writing systems, but struggles with vertical and right-to-left texts, as the gaze pattern is in a different direction. To address this, we find that simply augmenting the gaze at inference time (90° rotation for vertical texts and horizontal flip for right-to-left texts) allows the model to generalize well. In practical scenarios, this can be done depending on geo-location. During training, we also add a small fraction of rotated gaze to help with reading vertical texts.

\section{Experimental Setup}
\vspace{-0.1cm}
\subsection{Dataset split}

We split the Seattle subset into training, validation, and test sets, and train the model on the training set. We evaluate on (i) the test set of the Seattle subset, and (ii) the entire Columbus subset. We also evaluate on specific subsets to study latency and generalization.

\subsection{Implementation details}

\noindent\textbf{Model.} For the encoders, we use three layers of 1D convolution (kernel size 9, 32 dims) for gaze and IMU, and three layers of 2D convolution (kernel size 5, 32 dims) for RGB. We then feed the tokens as input to three layers of transformer encoder (32 dims, 2 heads) before linearly projecting the [CLS] token to two classes. The combined model is lightweight, with 137k parameters.

\noindent\textbf{Training.} We impose modality dropout such that there is an equal probability of using one, two, or three modalities at the same time, as well as perform rotation augmentation. We use Adam optimizer with learning rate $1e^{-3}$ for ten epochs. All models are trained using a single GPU. The code and models will be released alongside the dataset.

\subsection{Evaluation metrics}

\noindent\textbf{Classification metrics.} We calculate the accuracy and F1 scores for each task at 0.5 confidence threshold. We also vary this threshold, and report the precision at 0.9 recall (denoted as $P_{R=.9}$).

\noindent\textbf{Latency.} We consider latency to be the time between a state change and model detecting it, and is unrelated to the computational time, which we assume to be negligible given the small model size.
\begin{figure*}
\setlength{\tabcolsep}{4pt}
\subcaptionbox{Main results}{
\scriptsize
\begin{tabular}{lll|ccc}
\toprule
Gaze & RGB & IMU & Acc & F1 & P$_{R=.9}$ \\ \midrule
\checkmark &  &  & 82.3 & 84.5 & 79.8 \\
 & \checkmark &  & 82.2 & 83.7 & 76.5 \\
 &  & \checkmark & 74.7 & 80.0 & 71.9 \\ \midrule
\checkmark &  & \checkmark & 84.9 & 86.5 & 83.6 \\
 & \checkmark & \checkmark & 83.5 & 85.2 & 82.3 \\
\checkmark & \checkmark &  & 86.0 & 87.8 & 87.3 \\ \midrule
\checkmark & \checkmark & \checkmark & \textbf{86.9} & \textbf{88.1} & \textbf{88.0} \\ \bottomrule
\end{tabular}
}
\hfill
\subcaptionbox{Visualization (\textbf{G}/\textbf{R} are for gaze/RGB, with wrong ones in red)}{
\includegraphics[width=0.64\textwidth]{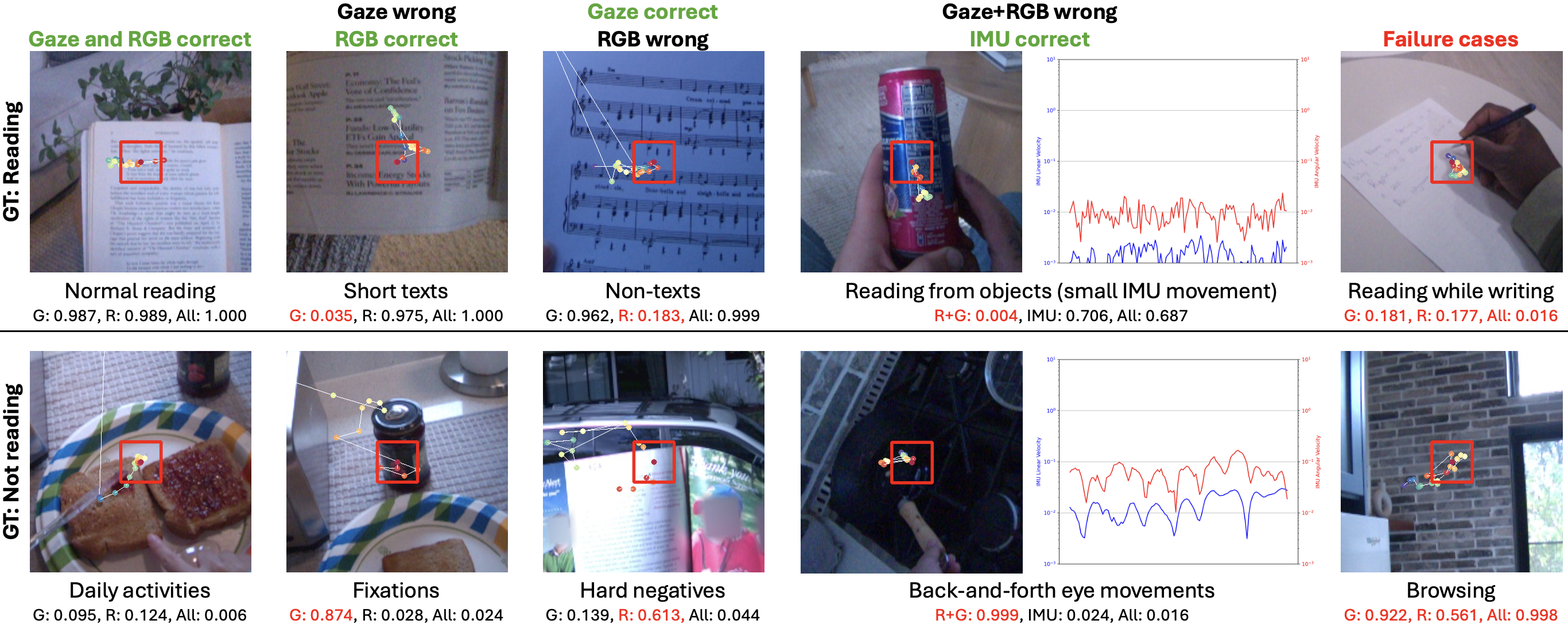}
}
\hfill
\vspace{-0.2cm}
\caption{\textbf{Main results and visualizations.} We show the results on Seattle (test set). (a) Our method performs the task to good accuracy, and combining all modalities yields the best results. Metrics are accuracy and F1 score at 0.5 threshold, and precision at 0.9 recall. (b) We show: (i) Col. 1, banal success cases distinguishing reading from daily activities; (ii) Col. 2-4, difficult cases where our combined model predicts correctly even if individual modality fails, including reading from objects, short texts, non-texts, fixation patterns, and hard negatives; (iii) Col. 5, failure cases where all modalities fail, including reading while writing and browsing.}
\label{fig:main}
\vspace{-0.4cm}
\end{figure*}

\section{Results}
\vspace{-0.1cm}

\subsection{Main results}

We present the main results and visualizations in Figure \ref{fig:main}.

\noindent\textbf{Single modality.} We find that gaze and RGB are able to achieve reasonable performance individually, and their performances are similar to each other (82.3\% and 82.2\% accuracy respectively). However, as shown in the visualizations, they have different success and failure cases.
IMU alone does not perform very well, which is reasonable, as the problem becomes ill-posed, and the model can only guess the lack of motion as not reading (and vice versa).

\noindent\textbf{Combined modalities.} We find that IMU monotonically improves upon gaze (+2.6\%) or RGB (+1.3\%) as secondary modality, with small extra compute. Qualitatively, we see that IMU helps improve several corner cases, and RGB is particularly strong for short texts. We also find that all modalities combined yields the best performance of 86.9\% in accuracy (+4.6\% from best single-modality model), validating the complementary roles of different modalities.

\begin{figure*}[]
\setlength{\tabcolsep}{4pt}
\subcaptionbox{Precision-recall curve}{
\includegraphics[width=.3\textwidth]{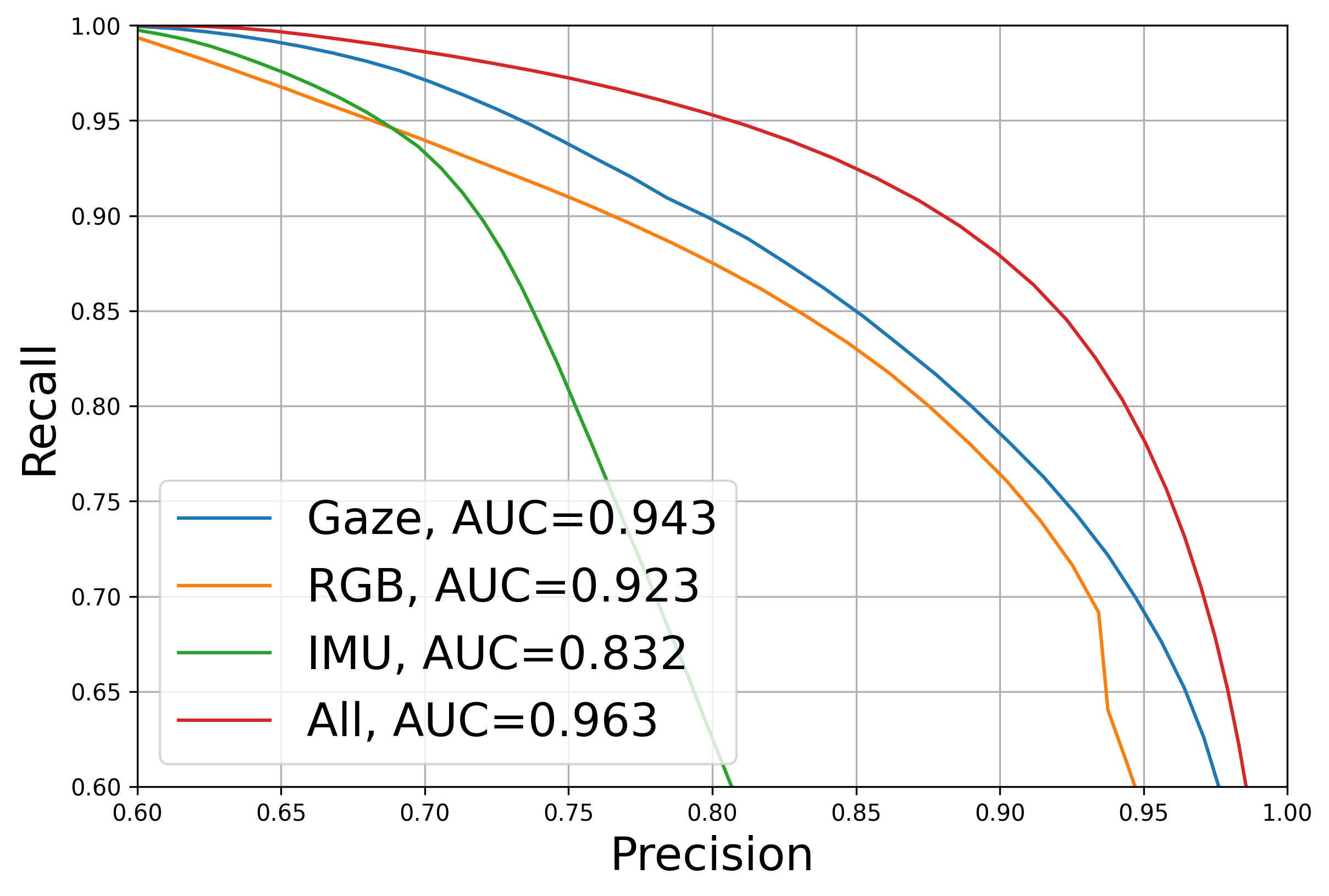}
}
\hfill
\subcaptionbox{Breakdown by scenario}{
\tiny
\begin{tabular}{l|c}
\toprule
{Scenario} & {Acc (\%)} \\ \midrule
Digital media (normal scenarios) & 95.3 \\
Print media (normal scenarios) & 93.8 \\
\textbf{Reading average} & \textbf{88.1} \\
Objects (normal scenarios) & 87.6 \\
Reading while walking & 81.4 \\
Reading from videos & 78.0 \\
Reading non-texts& 65.8 \\
Reading while writing/typing & 55.5 \\ \midrule
Daily activities & 95.2 \\
\textbf{Not reading average} &  \textbf{86.4} \\
Hard negatives & 74.7 \\ \bottomrule
\end{tabular}
}
\hfill
\subcaptionbox{Breakdown by gaze span}{
\includegraphics[width=.3\textwidth]{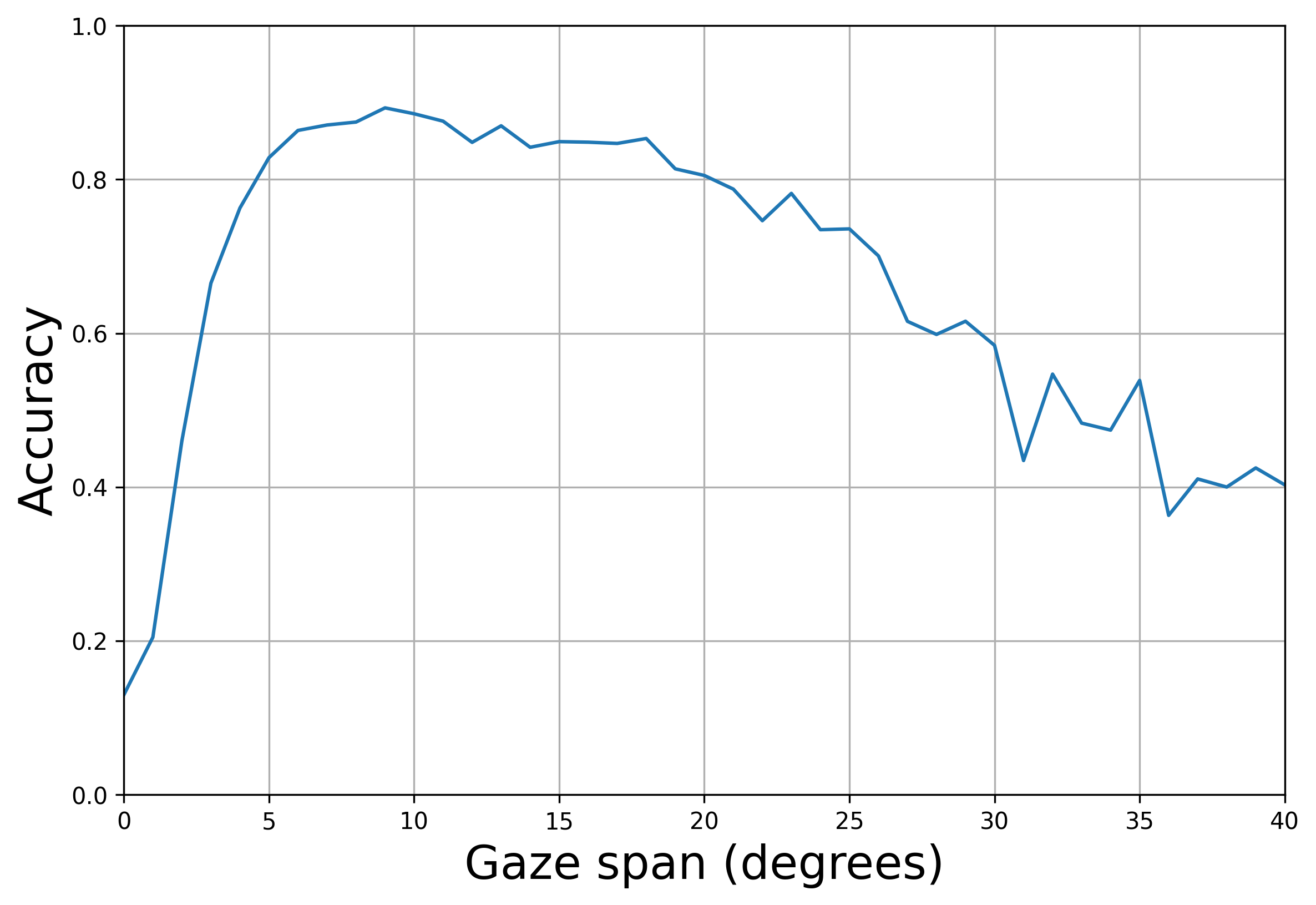}
}
\vspace{-0.2cm}
\caption{\textbf{Results breakdown.} We present the breakdown for the main results, including (a) precision-recall curve for different modalities (b) breakdown by scenario to highlight difficult cases (c) breakdown by gaze span.}
\vspace{-0.4cm}
\label{fig:breakdown}
\end{figure*}
\subsection{Results breakdown}

We show the breakdown of results in Figure \ref{fig:breakdown}.

\noindent\textbf{Scenario breakdown.} We break down the results of the combined model. We find that the model mostly succeeds in normal cases, but fails in cases where reading is atypical, such as reading non-texts (maps, music sheets), or when reading while writing or typing. The model also struggles with hard negative examples introduced in this dataset. 

\noindent\textbf{Gaze span breakdown.} We also break down the results of reading sequences by the horizontal gaze field of view, as it correlates with text size. We find that the accuracy is the highest (86.1\%) for fields of view of 5-20°, corresponding to 64-256 pixels, with accuracy dropping sharply for both below (59.3\%) and above (70.6\%) this range.

\begin{table*}[t]
\setlength{\tabcolsep}{2pt}
\subcaptionbox{Zero-shot on Columbus}{
\scriptsize
\begin{tabular}{lll|ccc}
\toprule
Gaze & RGB & IMU & Acc & F1 & P$_{R=.9}$ \\ \midrule
\checkmark &  &  & 77.1 & 84.0 & 84.1 \\
 & \checkmark &  & 76.7 & 84.5 & 83.4 \\
 \midrule
\checkmark & \checkmark &  & 82.8 & {88.7} & \textbf{88.2} \\ 
\checkmark & \checkmark & \checkmark & \textbf{82.9} & \textbf{88.8} &  \textbf{88.2}\\ \bottomrule
\end{tabular}
}
\hfill
\setlength{\tabcolsep}{2pt}
\subcaptionbox{Cross-language (text direction)}{
\tiny
\begin{tabular}{ll|cc}
\toprule
Language & Aug & Acc & F1 \\ \midrule
English $\rightarrow$ & - & 81.2 & 87.0 \\
Bengali $\rightarrow$ & - & 93.0 & 95.9  \\ 
Chinese $\downarrow$  & - & 35.5 & 51.6 \\ 
 & rotate & 85.1 {\tiny\color{ForestGreen}(+49.6)} & 91.9 {\tiny\color{ForestGreen}(+40.3)}  \\ 
Arabic $\leftarrow$  & - &  21.0 & 23.8  \\ 
 & flip & 51.5 {\tiny\color{ForestGreen}(+30.5)} & 63.8 {\tiny\color{ForestGreen}(+40.0)} \\
\bottomrule
\end{tabular}
}
\hfill
\subcaptionbox{Generalization to EGTEA}{
    \scriptsize
\setlength{\tabcolsep}{2pt}
\begin{tabular}{ll|cc}
\toprule
Test & Train & Acc & F1  \\ \midrule
Seattle & Seattle & 79.3 & 81.2 \\
 & EGTEA & 62.9 {\tiny\color{red}(-16.4)} & 56.9 {\tiny\color{red}(-24.3)}\\ \midrule
EGTEA  & EGTEA & 89.6 & 70.6  \\
& Seattle & 87.7 {\tiny\color{red}(-1.9)} & 63.4 {\tiny\color{red}(-7.2)}  \\
 \bottomrule
\end{tabular}
}
\hfill
\vspace{-0.2cm}
\caption{\textbf{Generalization results.} Using model trained on Seattle subset, we test on (a) separately collected Columbus subset; (b) different languages with different reading patterns and direction (despite only being trained with English), where we explore using rotation and flipping augmentations; (c) cross-generalization with EGTEA. The model generalizes one way (Seattle $\rightarrow$ EGTEA) but not the other.}
\label{tab:generalize}
  \vspace{-0.2cm}
\end{table*}

\subsection{Generalization}

We use the model trained on the Seattle subset to evaluate on unseen scenarios, shown in Table \ref{tab:generalize}.

\noindent\textbf{Zero-shot generalization.} To evaluate zero-shot capabilities, we test on the separately collected Columbus subset. We show that the model performs reasonably zero-shot, and draw similar conclusions in terms of the complementary role between gaze and RGB, but IMU does not help as much given that the dataset does not contain freeform daily activities where IMU helps the most.

Further, we also notice the differences in reading speed across different users, especially across different languages. It is possible to personalize the model by scaling the gaze to the magnitude of the reader, and empirically this solves some of the failure cases.

\noindent\textbf{Cross-language generalization.}
While we only train the model on English, we find that our model generalizes well towards non-English, left-to-right texts, but less well on other languages where the reading direction is different. To circumvent this during inference, we perform 90° rotation to tackle vertical texts, and horizontally flip the gaze for right-to-left texts. We show that using gaze-only model solves the problem to a reasonable extent.

\noindent\textbf{Cross-dataset generalization.}
To demonstrate the importance of collecting reading examples in freeform settings, we conduct experiments to test for generalizability across datasets. For this, we utilize EGTEA Gaze+ dataset \cite{li2018eye}, where we only use their `reading' action labels, and treat other labels as not reading. To match the data available in EGTEA, we use 2D gaze projection at $30$Hz. 
We conduct cross-generalization experiments where we train on one training set and evaluate on the other test set. We show that training on EGTEA with limited training samples does not generalize to in-the-wild scenarios, whereas the generalization gap for our dataset is much smaller.

\begin{figure*}
\setlength{\tabcolsep}{3pt}
\subcaptionbox{Latency}{
\tiny
\begin{tabular}{lll|ccc}
\toprule
Gaze & RGB & IMU & Acc & F1 & Latency (s) \\ \midrule
\checkmark(1s) &  &  & 77.1 & 75.6 & 0.526 \\
\checkmark(2s) & & & 79.0 & 78.9  & 0.831 \\
\checkmark(3s) &  &  & 79.3 & 77.8 & 1.013 \\  \midrule
 & \checkmark &  & 73.8 & 68.7 & \textbf{0.321} \\
\checkmark (2s) & \checkmark &  & 81.7 & 79.5  & 0.642 \\
\checkmark (2s) & \checkmark & \checkmark & \textbf{82.7} & \textbf{81.0} & 0.720 \\ \bottomrule
\end{tabular}
}
\hfill
\subcaptionbox{Visualization using Gaze+RGB+IMU model. [\textcolor{blue}{Prediction}/\textcolor{red}{GT}]}{
\includegraphics[width=0.62\textwidth]{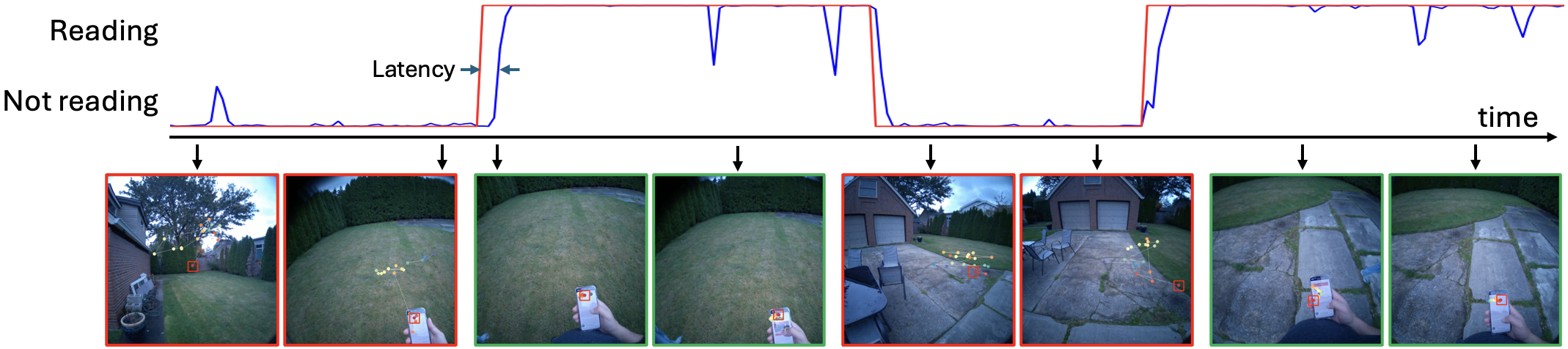}
}
\hfill
\vspace{-0.2cm}
\caption{\textbf{Real-time detection.} We evaluate our model on alternating sequences for real-time detection. In (a), we show that (i)  longer gaze sequences result in higher latency, (ii) RGB has lower latency than temporal signals (iii) adding RGB to gaze reduces the latency compared to gaze alone. We illustrate the results in (b).} 
\vspace{-0.4cm}
\label{tab:latency}
\end{figure*}
\begin{table*}[t]
\setlength{\tabcolsep}{4pt}
\subcaptionbox{Input representation}{
\tiny
\begin{tabular}{l|ccc}
\toprule
Input & Acc & F1 & P$_{R=.9}$ \\ \midrule
Retina images & 79.2 & 83.0 & 76.2 \\
3D ray (d/dt) & 82.1 & 84.2 & 78.4 \\
3D point & 80.8 & 83.3 & 77.9 \\
3D point (d/dt) & \textbf{82.3} & \textbf{84.5} & \textbf{79.8} \\
2D projection & 79.8 & 81.3 & 74.6 \\ 
Gaze + IMU  & 83.9 & 85.7 & 80.0 \\
Gaze + VIO  & \textbf{84.9} & \textbf{86.5} & \textbf{83.6} \\\bottomrule
\end{tabular}
}
\hfill
\subcaptionbox{Gaze frequency and duration}{
\scriptsize
    \begin{tabular}{l|cc||l|cc} \toprule
    Freq & Acc & F1 & Dur & Acc & F1  \\ \midrule
    \underline{60}& \textbf{82.3} & \textbf{84.5} & 5 & \textbf{85.8} & \textbf{87.5} \\
    30 & 81.7 & 84.3 & 4 & 85.4 & 87.1 \\
    20 & 81.3 & 83.6 & 3 & 83.6 & 85.7 \\
    10 & 80.4 & 82.9 & \underline{2}& 82.3 & 84.5 \\
    6 & 79.2 & 82.0 & 1 & 79.6 & 82.2 \\ \bottomrule
    \end{tabular}
}
\hfill
\subcaptionbox{RGB crop size}{
\scriptsize
\begin{tabular}{l|cc}
\toprule
FoV & Acc & F1  \\ \midrule
14  & \textbf{83.5} & \textbf{85.1}   \\
10  & 82.9 &  84.6 \\
7  & 82.9 &  84.3 \\
\underline{5}& 82.2 &  83.7 \\
3.5 & 79.5 & 80.6 \\ \bottomrule
\end{tabular}
}
\hfill
\subcaptionbox{Model size}{
\scriptsize
    \begin{tabular}{l|cc}
    \toprule
    Model & Acc & F1  \\ \midrule
    XS (6k) &  82.0 & 83.6  \\
    S (34k) & 86.3 & 87.7  \\
    \underline{M (137k)}&  86.9 & 88.1   \\
    L (600k) &  87.1 & 88.8   \\
    XL (1M) & \textbf{88.5} & \textbf{90.1}   \\ 
    \bottomrule
    \end{tabular}
}
\hfill
\vspace{-0.2cm}
\caption{\textbf{Ablation studies.} We show ablation studies for (a) the representations for gaze and IMU, (b) the gaze frequency and duration, (c) RGB crop size, and (d) model size. 
We fix other experiments to 60Hz, 2s, and 5° FoV using the M (137k) model, as \underline{underlined}.}
\label{tab:ablation}
\end{table*}
\begin{figure*}[t]
\setlength{\tabcolsep}{2pt}
    \centering
    \begin{minipage}{0.32\textwidth}
        \centering
        \includegraphics[width=\linewidth]{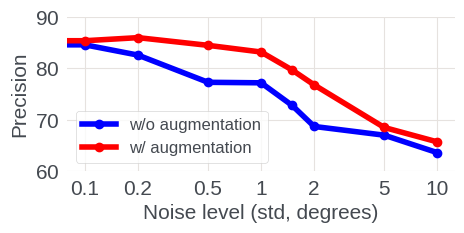}
        \caption{\textbf{Noise robustness.} Augmentation (red) lowers degradation.}
        \label{fig:noise}
    \end{minipage}\hfill
    \begin{minipage}{0.32\textwidth}
        \centering
        \tiny
        \begin{tabular}{lccccccc}
         GT \textbackslash Pred &
          1 &
          2 &
          3 &
          4 &
          5 &
          6 &
          7 \\ 
        1 No read &
          \cellcolor[HTML]{6CC499}0.88 &
          \cellcolor[HTML]{F9FDFB}0.04 &
          \cellcolor[HTML]{FCFEFD}0.02 &
          \cellcolor[HTML]{FCFEFD}0.02 &
          \cellcolor[HTML]{FEFFFE}0.01 &
          \cellcolor[HTML]{FAFDFC}0.03 &
          \cellcolor[HTML]{FFFFFF}0.00 \\
        2 Walk &
          \cellcolor[HTML]{F0F9F5}0.09 &
          \cellcolor[HTML]{71C69C}0.85 &
          \cellcolor[HTML]{F9FDFB}0.04 &
          \cellcolor[HTML]{FEFFFE}0.01 &
          \cellcolor[HTML]{FFFFFF}0.00 &
          \cellcolor[HTML]{FFFFFF}0.00 &
          \cellcolor[HTML]{FEFFFE}0.01 \\
        3 Out loud &
          \cellcolor[HTML]{EAF7F0}0.13 &
          \cellcolor[HTML]{FCFEFD}0.02 &
          \cellcolor[HTML]{94D4B5}0.64 &
          \cellcolor[HTML]{E3F4EC}0.17 &
          \cellcolor[HTML]{FCFEFD}0.02 &
          \cellcolor[HTML]{FEFFFE}0.01 &
          \cellcolor[HTML]{FEFFFE}0.01 \\
        4 Engaged &
          \cellcolor[HTML]{E8F6EF}0.14 &
          \cellcolor[HTML]{FCFEFD}0.02 &
          \cellcolor[HTML]{F5FBF8}0.06 &
          \cellcolor[HTML]{A5DBC0}0.54 &
          \cellcolor[HTML]{EBF7F1}0.12 &
          \cellcolor[HTML]{FEFFFE}0.01 &
          \cellcolor[HTML]{EDF8F3}0.11 \\
        5 Scan &
          \cellcolor[HTML]{F2FAF6}0.08 &
          \cellcolor[HTML]{FEFFFE}0.01 &
          \cellcolor[HTML]{FAFDFC}0.03 &
          \cellcolor[HTML]{BEE5D2}0.39 &
          \cellcolor[HTML]{BBE4D0}0.41 &
          \cellcolor[HTML]{FFFFFF}0.00 &
          \cellcolor[HTML]{F2FAF6}0.08 \\
        6 Write/type &
          \cellcolor[HTML]{ADDEC6}0.49 &
          \cellcolor[HTML]{FEFFFE}0.01 &
          \cellcolor[HTML]{FAFDFC}0.03 &
          \cellcolor[HTML]{FCFEFD}0.02 &
          \cellcolor[HTML]{F7FCFA}0.05 &
          \cellcolor[HTML]{BEE5D2}0.39 &
          \cellcolor[HTML]{FEFFFE}0.01 \\
        7 Skim &
          \cellcolor[HTML]{EAF7F0}0.13 &
          \cellcolor[HTML]{F9FDFB}0.04 &
          \cellcolor[HTML]{F7FCFA}0.05 &
          \cellcolor[HTML]{B1E0C9}0.47 &
          \cellcolor[HTML]{E6F5EE}0.15 &
          \cellcolor[HTML]{FFFFFF}0.00 &
          \cellcolor[HTML]{E5F5ED}0.16 \\
        \end{tabular}
        \captionof{table}{\textbf{Reading mode classification} using Gaze, RGB and IMU.}
        \label{tab:mode}
    \end{minipage}\hfill
    \begin{minipage}{0.32\textwidth}
        \centering
        \tiny
        \begin{tabular}{lcccc||cccc}
          GT \textbackslash Pred  &
          1 &
          2 &
          3 &
          4 &
          1 &
          2 &
          3 &
          4 \\
        1 No read &
          \cellcolor[HTML]{7ECBA5}0.77 &
          \cellcolor[HTML]{F4FBF7}0.07 &
          \cellcolor[HTML]{F9FDFB}0.04 &
          \cellcolor[HTML]{EBF7F1}0.12 &
          \cellcolor[HTML]{74C79E}0.83 &
          \cellcolor[HTML]{F9FDFB}0.04 &
          \cellcolor[HTML]{FAFDFC}0.03 &
          \cellcolor[HTML]{EFF9F4}0.10 \\
        2 Print &
          \cellcolor[HTML]{F4FBF7}0.07 &
          \cellcolor[HTML]{A3DABF}0.55 &
          \cellcolor[HTML]{CFECDE}0.29 &
          \cellcolor[HTML]{F0F9F5}0.09 &
          \cellcolor[HTML]{F2FAF6}0.08 &
          \cellcolor[HTML]{A6DBC1}0.53 &
          \cellcolor[HTML]{D5EEE2}0.25 &
          \cellcolor[HTML]{E8F6EF}0.14 \\
        3 Digital &
          \cellcolor[HTML]{F2FAF6}0.08 &
          \cellcolor[HTML]{CAEADA}0.32 &
          \cellcolor[HTML]{ADDEC6}0.49 &
          \cellcolor[HTML]{EDF8F3}0.11 &
          \cellcolor[HTML]{F4FBF7}0.07 &
          \cellcolor[HTML]{D2EDE0}0.27 &
          \cellcolor[HTML]{A6DBC1}0.53 &
          \cellcolor[HTML]{EAF7F0}0.13 \\
        4 Objects &
          \cellcolor[HTML]{EAF7F0}0.13 &
          \cellcolor[HTML]{D0ECDF}0.28 &
          \cellcolor[HTML]{CDEBDC}0.30 &
          \cellcolor[HTML]{CFECDE}0.29 &
          \cellcolor[HTML]{EAF7F0}0.13 &
          \cellcolor[HTML]{E1F3EA}0.18 &
          \cellcolor[HTML]{DBF1E6}0.22 &
          \cellcolor[HTML]{B1E0C9}0.47 \\
           &
          \multicolumn{4}{c}{(i) Gaze-only} &
          \multicolumn{4}{c}{(ii) Gaze+IMU} \\
        \end{tabular}
        \vspace{7pt}
        \captionof{table}{\textbf{Reading medium classification} using (i) gaze only (ii) gaze and IMU.}
        \label{tab:medium}
    \end{minipage}
    \vspace{-0.4cm}
\end{figure*}
\subsection{Application: real-time reading detection}

So far, we only consider atomic predictions to answer \textit{whether} someone is reading. To extend to \textit{when}, we simply perform predictions over time. To evaluate this task, we use the alternating sequences between reading and not reading with labeled timestamps, as shown in Figure \ref{tab:latency}. On top of the evaluation metrics, we also evaluate the latency (i.e. the duration required for a state change to be detected).
Our results show that (i) there is a trade-off between gaze duration and latency; (ii) RGB has lower latency as the predictions are instantaneous, and does not rely on past detections; and (iii) combining gaze and RGB reduces the latency compared to gaze-only model.

\noindent\textbf{Localization.} To extend to \textit{where} the user is reading, we can use the gaze point to locate the texts. 
As such, OCR only needs to be performed around the gaze, which results in additional compute savings.
Also, the gaze scanpath can be used to estimate how much to crop the image for OCR.

\noindent\textbf{Efficient interface for OCR.} OCR comes in two phases: text detection and recognition. Using reading recognition as a low-compute interface allows OCR to run not as often, and on a smaller image each time.
Furthermore, the reading detection model is designed to be small enough for on-device compute, so that images need to be transferred off-device only when reading detected, significantly reducing bandwidth requirements.

\noindent\textbf{Practical deployment.} We also investigate whether model of such size can be run practically. From parallel comparisons, the model can indeed comfortably run real-time on Aria Gen 2 glasses on-device, without the need to off-load the model to online computation. Given the estimated power consumption, the glasses can run for at least 4 hours continuously (inclusive of the base power consumption for basic computation, power delivery and sensor suites, and the thermal constraints). As the model runs on-device without having to send the model input and output back and forth to the server (as would have been done with, say, VLMs), the latency is negligible.

%

\subsection{Ablation studies}

Table \ref{tab:ablation} summarizes our results for ablation studies.

\noindent\textbf{Gaze representation.} The gaze processing pipeline involves transforming the retina images into ray angles for each eye, the intersection of which is the 3D gaze point in space, then projecting it onto the 2D image plane. We experiment using all these representations, and find that 3D gaze yields superior results, and pre-differentiating the input with respect to time leads to better generalization.

\noindent\textbf{Head pose representation.} 
With SLAM camaras, we can calculate the visual-intertial odometry (VIO) outputs using visual SLAM, which yields slightly better results compared to raw IMU sensors.

\noindent\textbf{Input frequency and duration.}
We experiment with varying frequency and duration for eye gaze. We find that higher frequency results in better performance, but also comes with compute tradeoffs. We notice similar trends for IMU.

\noindent\textbf{RGB crop size.} While we know that human fovea only covers 2°, we find that a larger crop provides context and covers for errors in gaze estimation. However, the compute also grows quadratically.

\noindent\textbf{Model size.}
We experimented other model sizes, with XS, S, M, L having 8, 16, 32, and 64 latent dimensions respectively. We also experimented with a pretrained image encoder (MobileNetV3-S) in the XL variant. We find that stronger model results in better results, and notably the S model performs surprisingly well with only $6$k parameters.


\noindent\textbf{Robustness to eye tracking precision.}
While our model is robust to fixed gaze offsets as we only use relative positions, noisy gaze predictions can ruin the gaze pattern. We test for the robustness to noise using our gaze-only model by adding Gaussian noise to the gaze inputs in two settings (i) only at test time and (ii) both during training (as augmentation) and testing. Our results in Figure \ref{fig:noise} show the performance degrades with noise, and training with noise helps with robustness.


\subsection{Extension: understanding types of reading}
Many existing cognitive studies try to understand how humans read, as it is related to understanding human behavior, comprehension, and health. As mentioned, current experiments and datasets are unrepresentative of how we read. In contrast, our dataset extends to ``in the wild'' settings, and we hope that our dataset will be useful in advancing the understanding of reading in the real world. Note that we use the same settings as previous experiments (2s time window), which may be limited in such fine-grained classification tasks.

\noindent\textbf{Reading mode classification.}
Many studies are interested in how people read \cite{biedert2012robust, yu2018fast, liao2017classification}. We conduct similar studies using our dataset. Specifically, we treat this as a 7-way classification problem (not reading, reading while walking, reading out loud, engaged reading, scanning, reading while writing/typing, skimming), and train the model for this task on our dataset. As shown in Table \ref{tab:mode}, we find that walking is an obvious category to detect (perhaps due to IMU), followed by reading out loud. Distinguishing between skimming, scanning, and engaged reading proved to be difficult.

\noindent\textbf{Reading medium classification.}
Inspired by \cite{kunze2013know} that tries to answer ``what" someone is reading, we also conduct similar experiments. In this case, we do not use RGB as the solution would have been trivial, and use the model to classify between four classes (not reading, print media, digital media, objects). We find that the task is difficult, and IMU helps in this case, as shown in Table \ref{tab:medium}.
\section{Broader Impact}
Always-on smart glasses raise important questions about social acceptability, both for the wearer and for the public, especially when such technologies are deployed at scale. We hope that the ideas presented in this paper can also help mitigate such concerns.

\textbf{Safety.} Sensitive personal data, such as eye gaze, introduces unique risks. Our algorithm runs fully on-device, which ensures that sensitive information does not need to leave the user’s device. This is a step toward stronger privacy protections for wearers. At the same time, we acknowledge that using eye gaze as a signal creates new challenges. Eye movement can reveal intentions, interests, and even emotional states, which raises a distinct category of privacy concerns.

\textbf{Surveillance.} Our work aims to reduce reliance on invasive sensing. First, by leveraging eye gaze, we minimize the required front-camera capture to a very small patch (0.2\% of the full image) rather than recording the entire scene. Second, our approach can operate solely on eye gaze data without requiring any camera input. More broadly, eye gaze offers a powerful cue about where the user is looking, which enables RGB capture to be more targeted. This reduces the risk of collecting unintended or intrusive information about bystanders. We hope that future algorithms continue in this direction.

\textbf{Data Governance.} We follow the Project Aria Research Guidelines and will release our system with a Responsible Use Policy to promote ethical research practices and to support safe deployment.

\section{Conclusion}
Motivated by use cases in contextual AI and other applications, we explore the problem of reading recognition in real-world scenarios, and present a dataset that reflects this nature. We then present a method to solve the task using three modalities, and extend the studies towards reading understanding tasks. 
There are vast opportunities for future work. Our dataset can be used to study the reading behavior of people in realistic settings in greater detail which links to cognitive understanding. Our proposed protocol allows for scalable future data collection using smart glasses.
Additionally, model personalization to address variations in reading speed and style, along with predicting optimal modality activation for enhanced efficiency, represents another promising area for future work.


\clearpage
\section*{Acknowledgements}
We thank Rowan Postyeni for assistance with data collection, and Michael Goesele, Laurynas Karazija, and Lingni Ma for helpful feedback.

{
    \small
    \bibliographystyle{ieeenat_fullname}
    \bibliography{main}
}


\clearpage
\appendix

\begin{center}
    {\LARGE \bfseries Reading Recognition in the Wild}\\[1ex]
    {\large --- \textit{Supplementary Material} ---}
\end{center}

\vspace{2ex}

\section{Introduction}\label{sec:appendix_intro}

\textbf{Additional dataset details.} Our dataset is the first instance of reading activity recognition dataset in unconstrained environments and is also the first egocentric dataset with high-frequency eye-tracking (60 Hz) collected through Project Aria. By open-sourcing this dataset, we aim to foster greater involvement from the research community in leveraging the eye gaze data from egocentric devices. Table \ref{tab:supp_1} describes the dataset content in greater detail. In Sections \ref{sec:seattle} and \ref{sec:columbus}, we provide detailed information on the Seattle and Columbus subsets of the datasets, respectively. 

\textbf{Additional experimental results.} In Section \ref{sec:columbus_supplementary_results}, we provide additional zero-shot experiment results on Columbus subset, which includes ablations and qualitative results. In Section \ref{sec:compare}, we compare our method with other existing solutions, such as VLMs, action recognition models, and alternative baselines. Lastly, we include a discussion in Section \ref{sec:discussion} to expand on some experiments, limitations, and future work.



\begin{table}[h]
\centering
\scriptsize
\begin{tabular}{l|l|l}
\toprule
Category  & Sub-category           & Description/examples                                                                     \\ \toprule
Medium    & Print                  & Books, magazines, newspapers, fliers/brochures                                           \\
            & Digital          & Content: news, emails, wikipedia articles, blog posts/forum, e-books, social media, research paper           \\
          &                        & Devices: phone, laptop screen, monitor screen, tablet                                    \\
            & Objects          & Nutrition labels, product labels, posters/bulletin board, signs, sticky notes, text on whiteboard             \\ \midrule
Text type & Paragraphs/Short texts & See Fig. 3 (Main Paper)                                                                  \\
          & Short texts            &                                                                                          \\
          & Non-texts              & Children's books, comic books, instruction manuals, maps, music sheets                   \\ 
          & Dynamic texts        & Participants are asked to read the subtitles as a video is playing                       \\ \midrule
Multi-task  & None                &   N/A                                                                                       \\
            & While walking    & Participants are asked to hold the reading material in their hand(s) while reading and walking               \\
          & While writing/typing   & Participants are asked to read as they write or type                                     \\ \midrule
Mode      & Engaged                &                                                                                          \\
          & Skimming               & Participant is asked to skim the text quickly to get the understand the gist of the text \\
            & Scanning         & Participant is given a pre-reading question, and is asked to look at the reading material to find the answer \\
          & Reading out loud       & Participant is asked to read the text out loud                                           \\ \midrule
Not reading & Daily activities & Five categories: physical exercise, outdoor activity, creative arts, culinary activity, household chores     \\
          & Hard negatives         & Text is present in scene but user is not reading                                         \\ \midrule
Mixed     & Alternating sequences  & Alternating between reading and not reading, with start/end timestamps                   \\
          & Mirror setups          & A pair of sequences with the same settings, one is reading and another is not reading   \\ \bottomrule
\end{tabular}
\vspace{2mm}
\caption{\textbf{Dataset glossary.} This table describes the samples in the dataset in greater detail.}
\label{tab:supp_1}
\end{table}

\clearpage
\section{Reading in the Wild - Seattle Subset}
\label{sec:seattle}
 The Seattle subset contains about 80 hours of Aria recordings (total of 1,061 videos)  with eye gaze calibration, with 81 participants, taken both indoors and outdoors.  

\subsection{Type of reading materials covered}
 The dataset encompasses a wide range of reading materials and content on various mediums. These include:

\begin{itemize}
    \item \textbf{Print:} Books, magazines, newspapers, flyers, and brochures.
    \item \textbf{Digital:} News articles, emails, Wikis, blog posts, bulletin boards, e-books, social networks, and research articles, accessed through four types of digital devices: phones, laptops, monitors, and tablets.
    \item \textbf{Object:} Posters, nutrition/product labels, whiteboards, and sticky notes.
\end{itemize}

\begin{figure}[h]
    \centering
    \includegraphics[width=\linewidth]{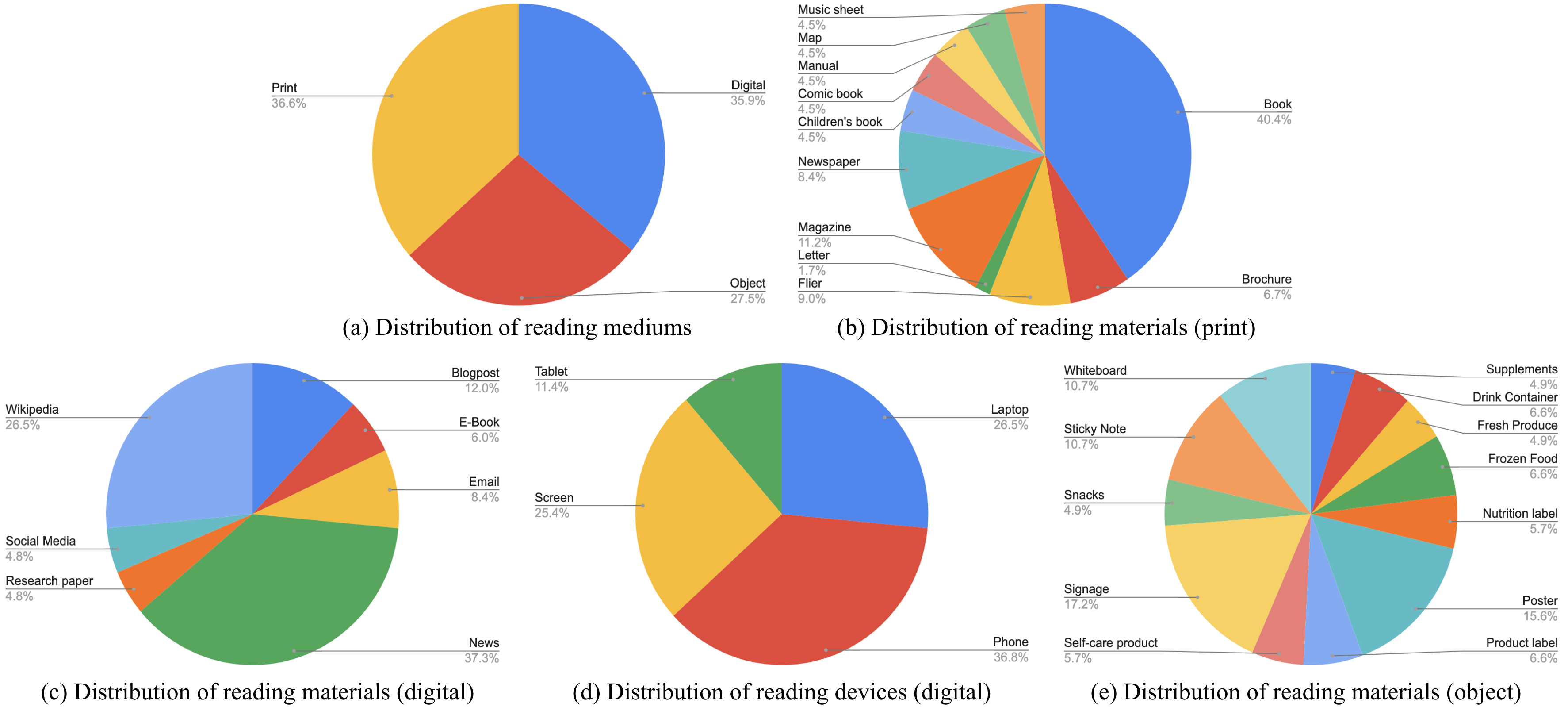}
    \caption{\textbf{Distribution of reading materials in Seattle subset}. In (a), we show the distribution of reading mediums. Within each medium (Print, Digital and Object), we then break down the reading materials in (b), (c), and (e). Additionally, for Digital media, we break down by the device involved in the recording. Refer to Fig. 3 (Main Paper) for illustrations.}
    \label{fig:material}
\end{figure}

In addition, the dataset includes comic books, maps, video captions, and music sheets, as described in Table \ref{tab:supp_1}. We provide the statistics in Figure \ref{fig:material}.

\subsection{Type of reading modes covered}
\begin{figure}[h]
  \centering
  \includegraphics[width=\linewidth]{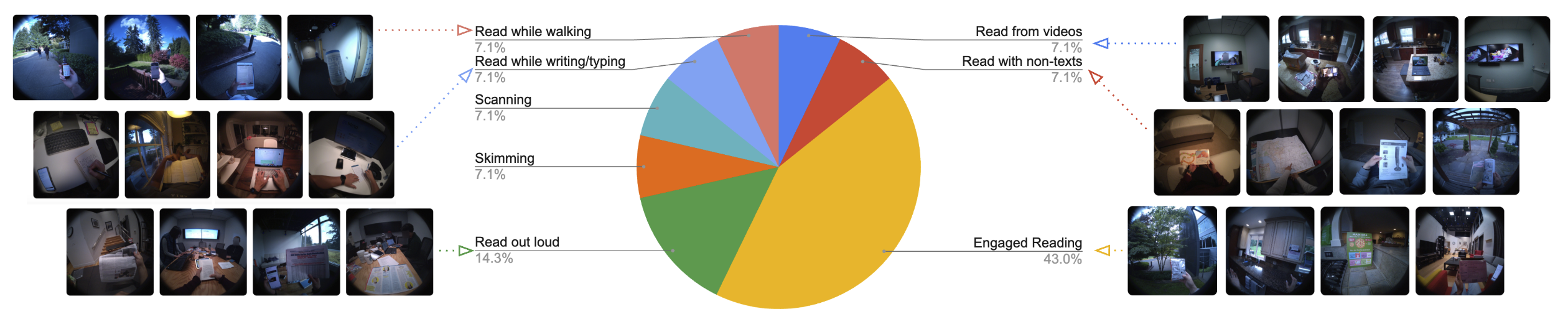}
  \vspace{-5pt}
  \caption{\textbf{Distribution of reading modes in Seattle subset} together with illustrations. Almost half of our reading samples is engaged reading, while other scenarios diversely reflect how people read in different scenarios.}
  \label{fig:stats_seattle_reading_mode}
  \vspace{-0.4cm}
\end{figure}

We collected data for different types of reading modes: read out loud, read as you write or type, read as you walk, scanning, and skimming. This is shown in Table \ref{fig:stats_seattle_reading_mode}.

\subsection{Negative data}
\begin{figure}[h]
\centering
\setlength{\tabcolsep}{4pt}
\subcaptionbox{Distribution of daily activities in normal negative data}{\includegraphics[width=0.4\linewidth]{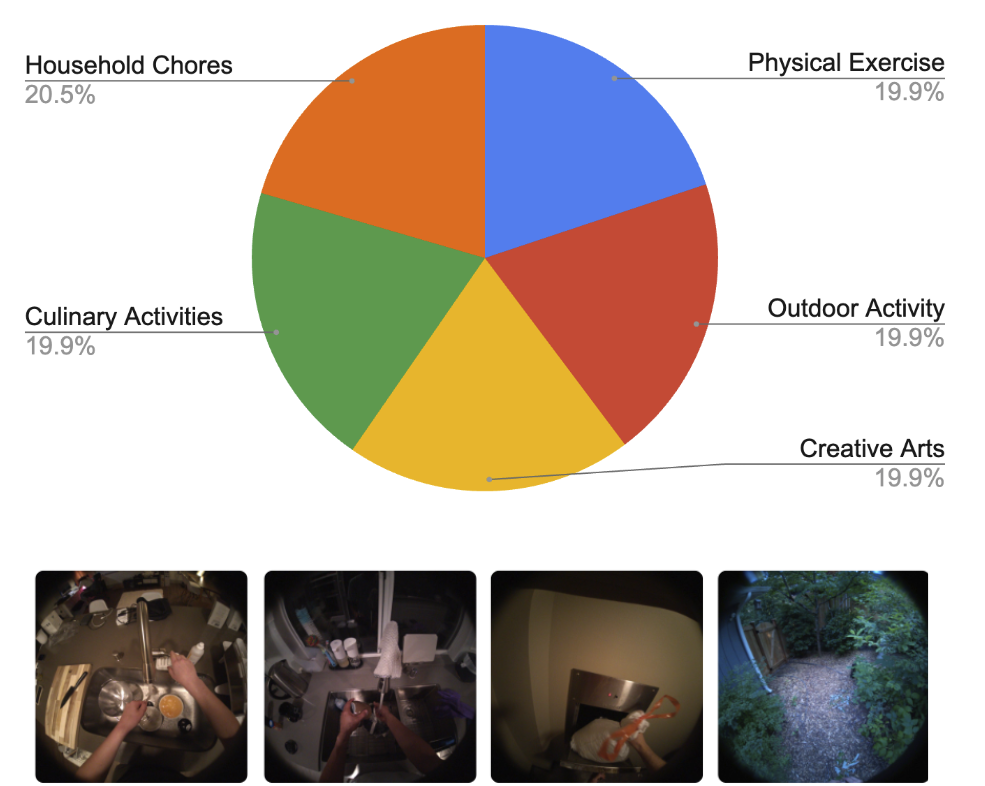}
}
\subcaptionbox{Distribution of reading materials in hard negative data}{
\includegraphics[width=0.4\linewidth]{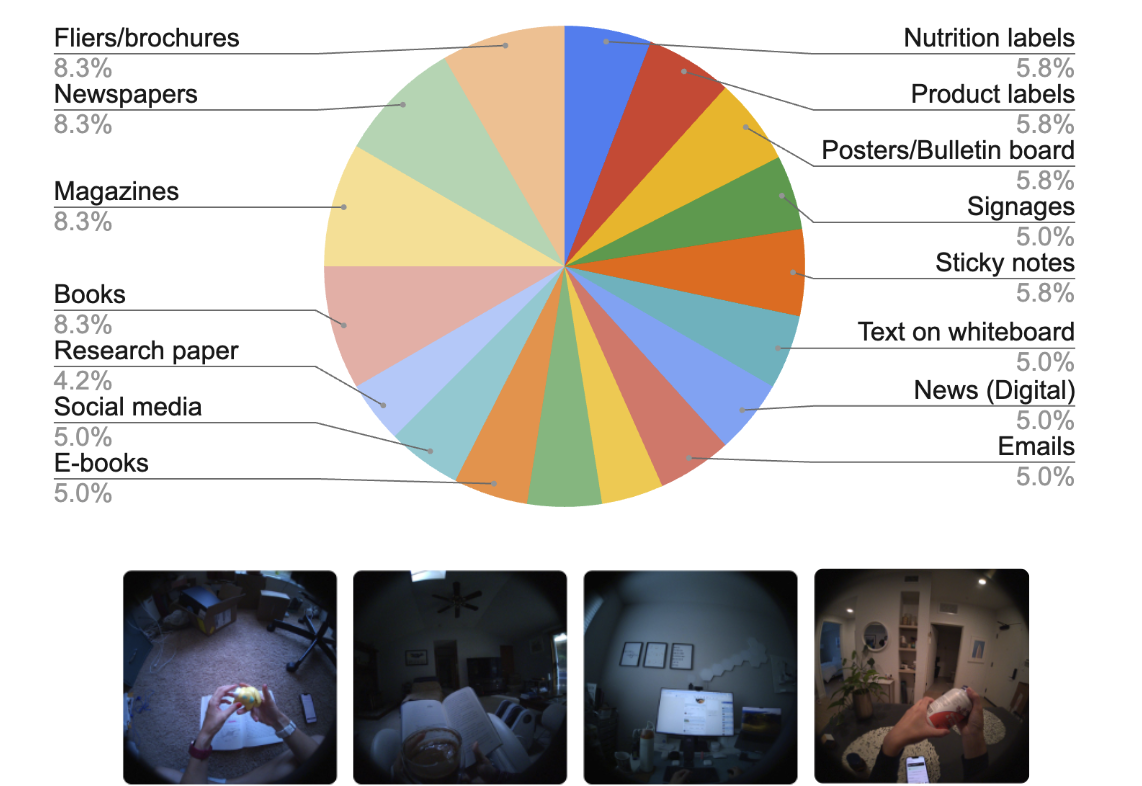}
}
\vspace{-0.2cm}
\caption{\textbf{Distribution of negative data in Seattle subset.} In addition to daily activities, our dataset also includes hard negatives, where the user has a text in scene but is not reading, making it indistinguishable using the RGB stream alone.}
\vspace{-0.2cm}
\label{fig:negative}
\end{figure}

We collected two types of negative data: (1) everyday activities that do not involve reading and (2) hard negatives where text is visible but not being read by the participant. Examples include moving around an object with text (e.g., a vitamin container) without reading it, and spinning a pen while reading material is present but focusing on the pen instead. Metadata is also included to indicate whether the user accidentally read the text. The examples are shown in Figure \ref{fig:negative}.

\subsection{Alternating sequences}
We also collected test sequences that alternate between reading and non-reading activities, allowing for the evaluation of temporal localization and latency. We provide the scenario list in Table \ref{tab:alternating}.
\begin{table}[h]
\scriptsize
\centering
\begin{tabular}{p{0.01\linewidth} | p{0.2\linewidth} | p{0.7\linewidth
}}
\toprule
\multicolumn{1}{l}{\#} & Summary & Scenario \\ 
\midrule
1 & reading + other acitvities (getting up) & You are reading {[}Material{]}, then you stop and get up and (go for a walk/prepare some coffee/adjust the lights/ open or close the door or window/ wash your hands/ do any random activity), then you come back to continue reading \\
2 & reading + mind wondering (while still sitting) & You are reading {[}Material{]}, then while reading your mind wonders, so you stop reading while still holding the material/keeping the same posture (mind wondering), maybe looking around to rest your eyes, maybe staring at nothing, then you get back to reading again. \\
3 & reading text + looking at images & You are reading {[}Material{]} containing texts and images, then while reading you inspect the accompanying graphics instead for a while, then you get back to reading again \\
4 & reading + small physical activity & You are reading {[}Material{]}, then you realize you're sitting in an uncomfortable position, so you stop to adjust your seating, and/or stretch your legs a bit, or move to a different position, chair, or posture, then you come back to continue reading. \\
5 & reading + eating/drinking & You are reading {[}Material{]} while enjoying a cup of tea/coffee/drink/food, everytime you take a sip/bite you stop reading, then come back to reading again \\
6 & walking + stopping at a text & You are walking, then you see a {[}Material -- something fixed, like a sign, or a screen{]}, so you stop to read it, then continue the activity after you finish, maybe you walk back to reread the text or then read another text \\
7 & walking + grab something to read & You are walking, then you see a {[}Material -- something that you can grab{]}, so you grab it either stand while reading or sit down to read it, then you grab something else to read similarly \\
8 & cleaning + reading & You are cleaning a room or organizing a table, then you see a {[}Material{]}, so you grab and sit down to read it. Maybe you are sorting documents and have to briefly read all of them before knowing where to put each one \\
9 & cooking while reading & You are in the kitchen, while making food, you are also multitasking and reading {[}Material{]} \\
10 & assembly while reading & You are assembling something or doing something that requires following step by step (text) instructions, periodically consulting the instruction manual, then continuing with the hands-on work.
\\ \bottomrule
\end{tabular}
\vspace{2mm}
\caption{\textbf{List of scenarios for alternating sequences.} We collect 12 sequences for each scenario with varied media types, for a total of 120 videos, each with varied reading materials. We also provide annotated timestamps for each start/stop section.}
\label{tab:alternating}
\end{table}

\subsection{Demographics}

\begin{figure}[h]
    \centering
    \resizebox{0.7\linewidth}{!}{
\begin{tabular}{cc}
    \textbf{Age Range} & \textbf{Gender} \\
    \begin{tikzpicture}
        \pie[
            text=inside,
            radius=2.5,
            explode=0.0,
            color={osured, green!70, osublue, orange!70, yellow!70, purple!60},
            after number=\%,
           style={very thin}
        ]{
            10/,
            23.75/,
            22.5/,
            22.5/,
            12.5/,
            8.75/
        }
    \end{tikzpicture}
     \begin{tabular}{@{}l@{}}
        \textcolor{osured}{\rule{10pt}{10pt}} 18-24 \\ 
        \textcolor{green!70}{\rule{10pt}{10pt}} 25-30 \\
        \textcolor{osublue}{\rule{10pt}{10pt}} 31-35 \\ 
        \textcolor{orange!70}{\rule{10pt}{10pt}} 36-40 \\
        \textcolor{yellow!70}{\rule{10pt}{10pt}} 41-45 \\
        \textcolor{purple!60}{\rule{10pt}{10pt}} 46-50 \\
    \end{tabular} 
    &
    \begin{tikzpicture}
        \pie[
            radius=2.5,
            explode=0.0,
            color={osured, osublue},
            text=inside,
            after number=\%,
            style={very thin}
        ]{
            52.5/, 
            47.5/
        }
    \end{tikzpicture} 
    \begin{tabular}{@{}l@{}}
        \textcolor{osured}{\rule{10pt}{10pt}} Male \\ 
        \textcolor{osublue}{\rule{10pt}{10pt}} Female
    \end{tabular} \\
    (i) & (ii)
\end{tabular}}
    \caption{\textbf{Demographic statistics of the Seattle subset} (i) shows the age group, (ii) shows gender distribution}
    \label{fig:demograhics_sea}
\end{figure}
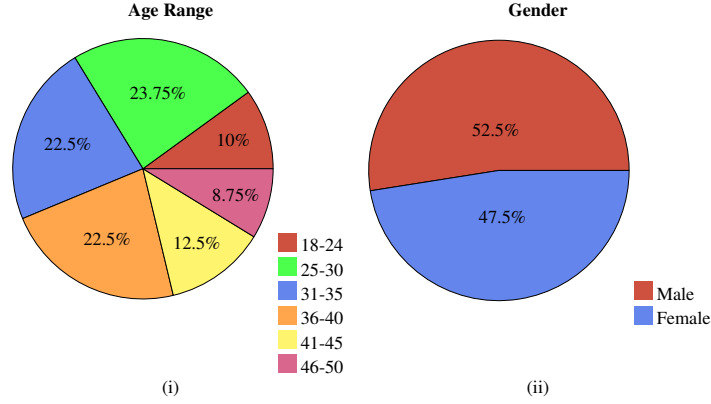

The demographics are presented in Figure \ref{fig:demograhics_sea}. We ensure a diverse range of demographics across gender and age in our collection.

\subsection{Metadata}
The metadata provides valuable information about the reading task and the associated reading material, reading medium, summaries, and multiple-choice question IDs and answers if applicable. This is shown in Figure \ref{fig:metadata_example}. In addition to these, we also collect personal demographics (age range, gender) as part of the metadata.
\begin{figure}[h]
\begin{mdframed}
\footnotesize
\begin{verbatim}
"name": "13. Write or type texts - Read Out Loud 21",
"notes": "I wrote about tattoos, the week, what food I want.",
"recording_profile": "profile28",
"tags": [
    {
        "title": "How long is the text?",
        "tags": [
            "Equal or greater than 4 sentences"
        ]
    },
    {
        "title": "Where did you write/type texts? ",
        "tags": [
            "Phone"
        ]
    },
    {
        "title": "How did you write the text? ",
        "tags": [
            "Typing"
        ]
    }
]
\end{verbatim}
\end{mdframed}
\caption{\textbf{Example of metadata for Seattle subset.} The metadata contains several useful information to facilitate further research, including both multiple-choice questions and short answers.}
\label{fig:metadata_example}
\end{figure}

\subsection{Protocols}
A successful data collection protocol ensures efficient collection while guaranteeing the quality. 
Reading is a complex process that includes word recognition, which encompasses visual processing and language decoding, and comprehension, which involves linking information to memory and cognitive processing. Therefore, using a single protocol often proves challenging. For example, asking the participants to read text out loud guarantees that every single word is read / seen. However, it cannot prevent participants' minds from wandering. On the other hand, giving the participants a set of questions to answer can make sure that the participants actually understood the text. However, it makes the task harder than necessary as it also puts the subjects’ other cognitive capabilities (e.g. memory) in question. 

Instead, different protocols can be tailored to specific reading modes and materials. To address various reading modes such as scanning, skimming, or engaged reading, and to accommodate different types of reading materials, we use the following strategies. Please note that this captures only a portion of the entire problem space:

\begin{itemize}
    \item \textbf{Pre-reading questions:} 
Presenting questions before the reading task can focus participants' attention on specific details, making it effective for scanning tasks where the goal is to quickly locate particular information.
    \item \textbf{Post-reading questions and summaries:} 
Activities such as answering questions, summarizing the content (which can be evaluated using LLMs) can help assess comprehension by indicating whether participants have understood and can apply the information they've read. Specifically, we ask the participant to summarize what they read in a few words for all our tasks. For a set of reading materials (12 digital contents), we prepared multiple-choice questions and had the participant answer them.
    \item \textbf{Verbalization and recall:} 
 For shorter texts, asking participants to read aloud the text or recalling the text is an effective way to ensure that person read the text. Reading out loud is also effective for reading while writing or typing, as the pace of writing is generally slower than speaking. By asking the participant to read out loud can also capture valuable insights into how reading behavior changes between reading silently and reading aloud, especially since the pace of speaking can influence gaze behavior.
    \item \textbf{Embedded tags:} Techniques like embedding AprilTags within the reading material can provide additional signals on whether the participant comprehended the content (For example, if the instruction is to "Go to page 20 if you decide to escape," and the participant turns to page 20, it indicates that they paid attention to content.). This approach also enables automatic segmentation of reading activity (starting and end times). We prepared 10 CYOA Choose Your Own Adventures) books with AprilTags appended to specific pages: April Tag 0 for the first page after the cover, indicating the start of reading; April Tag 1 for the branching page; April Tag 2 on the page where they should land if they correctly read the instructions.
\end{itemize}

In addition, we also tailored protocols for specific reading materials with non-texts:
\begin{itemize}
    \item \textbf{Music sheets:} 
We pre-screen for participants who can read music notes, and they are asked to hum the notes as they read.
    \item \textbf{Maps:} 
Participants are asked to describe how to navigate from place A to place B as they read the map.
    \item \textbf{Instruction manuals:} 
We employ two approaches: participants are asked to summarize the steps and also to follow the instructions to assemble or set up items like board games, Lego-alikes, and origami.
    \item \textbf{Video captions:} 
We turned off the audio, such that the participants are forced to read captions. They are also asked to provide the summary of the video. 
\end{itemize}

\subsubsection{Scalable first-person annotation}
To avoid manual labeling of timestamps, we instruct the participants to say ``start reading!" whenever they start reading, and ``finished reading!" whenever they finish. This approach allows us to use a speech recognition model WhisperX \cite{bain2023whisperx} to obtain accurate timestamps without requiring manual annotations, making the process scalable. It is also important to have the participant  to annotate in this manner, as it would be challenging for a third party to accurately segment the reading sessions.

\subsection{Annotation}
The majority of our data is automatically labeled using WhisperX (speech recognition) and AprilTag detection. However, for certain tasks, such as our test sequences, where participants switch between reading and non-reading activities, manual inspection and re-labeling are necessary to ensure accuracy. For reading out loud tasks, we use the beginning and end of speech to mark the reading portion. For silent reading tasks without AprilTags, we use a simple verbal cue (``start/finished reading") to mark the beginning and end of each reading session.


\newpage

\section{Reading in the Wild - Columbus Subset}
\label{sec:columbus}

The Columbus subset contains data collected from $30$ subjects. The data was collected using the Aria wearable glasses indoors on the university campus. The study was reviewed and approved by the university's Institutional Review Board (IRB). 
Similar to the Seattle subset, the raw data include a single RGB (30Hz 1408p, 110 FoV\textdegree), two SLAM(150\textdegree FoV) and two infrared eye tracking (60Hz, calibrated) video streams, and two IMU data streams. Additionally, we collected spatial audio from the glasses and recorded task completion times of each segment for annotation. The subset contains 640 recordings corresponding to 18 hours of reading/non-reading tasks.

\subsection{Type of reading materials covered}
\begin{figure}[h]
    \centering
    \resizebox{\linewidth}{!}{
    \begin{tabular}{ccc}
    \includegraphics[width=0.3\linewidth,trim={2.5cm 1cm 2cm 0.5cm}, clip]{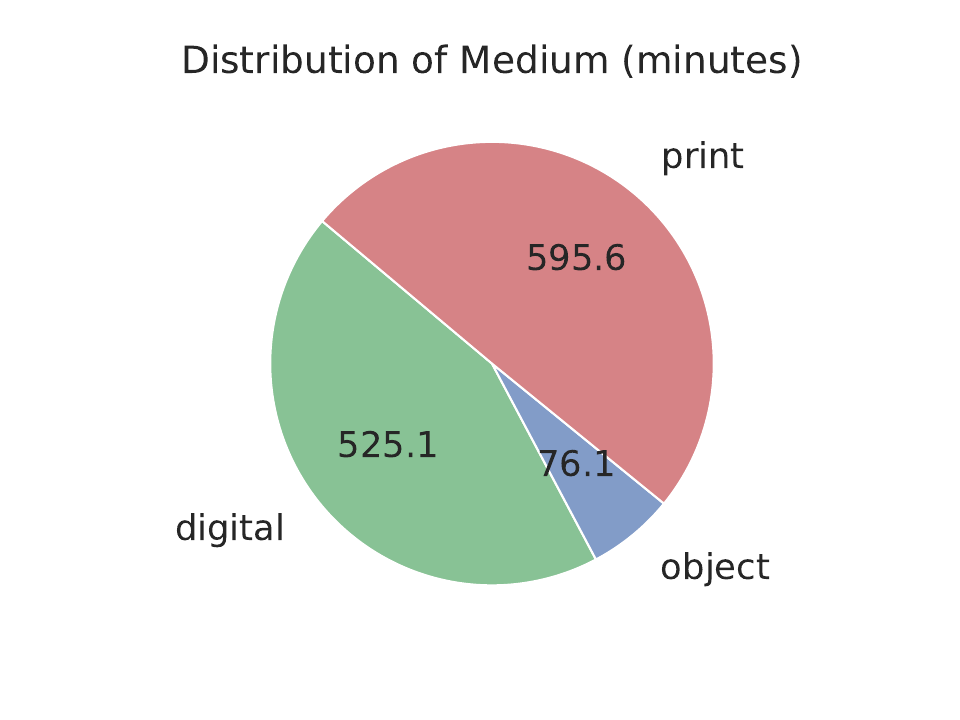}
    & 
        \includegraphics[width=0.33\linewidth,trim={2cm 1cm 1.5cm 0}, clip]{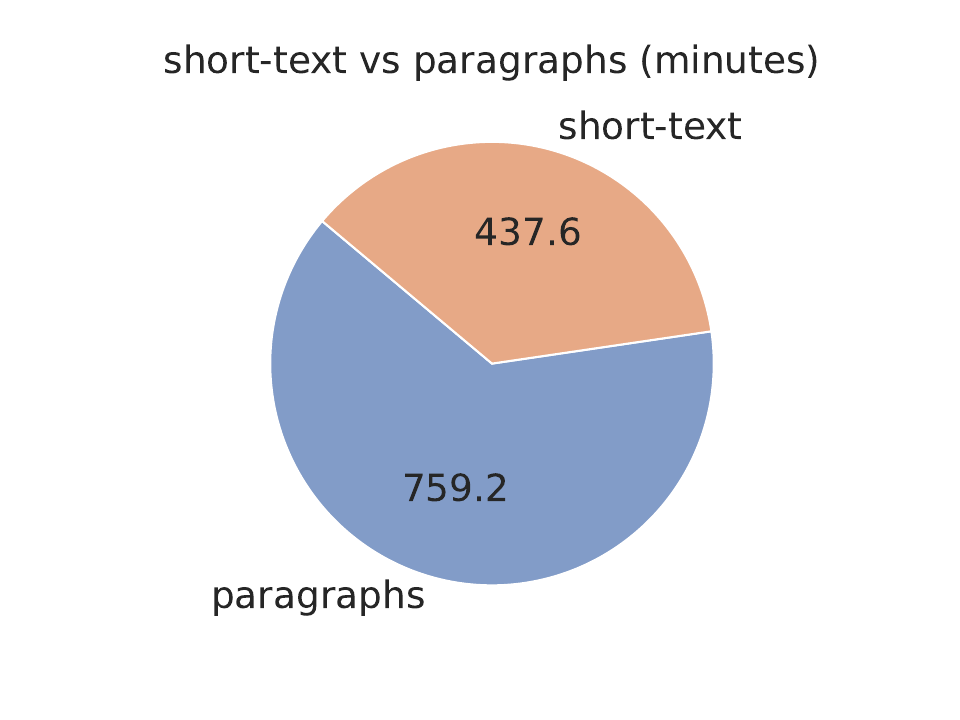}
        &
        \includegraphics[width=0.32\linewidth,trim={1.8cm 3cm 0cm 1cm}, clip]{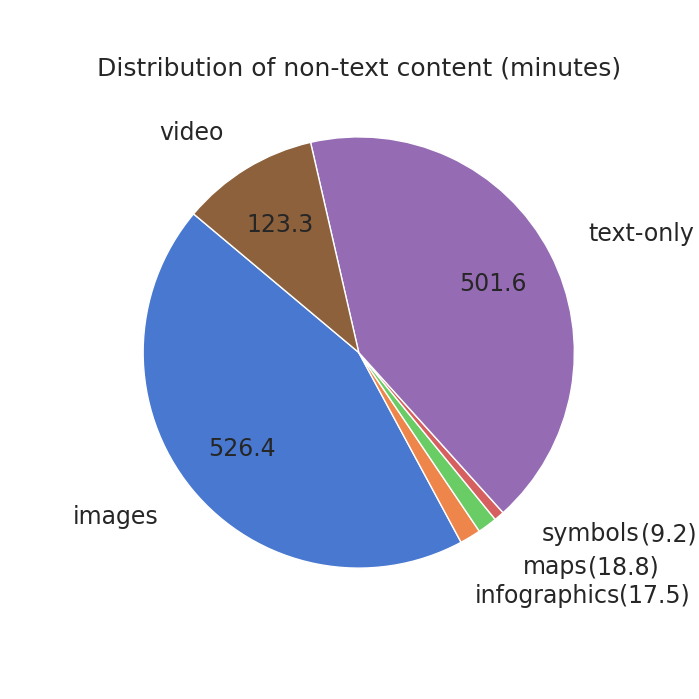}
    \\
      (a)   &  (b) & (c) \\

        \includegraphics[width=0.3\linewidth,trim={2.5cm 1cm 2cm 0.5cm}, clip]{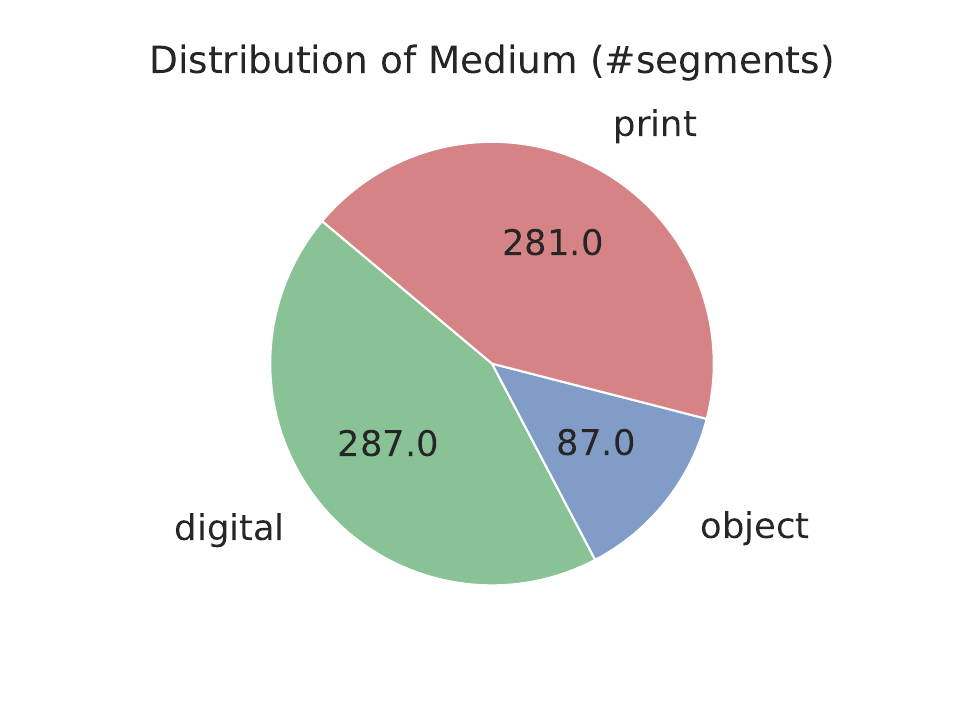}
 & 
        \includegraphics[width=0.33\linewidth,trim={2cm 1cm 1.5cm 0}, clip]{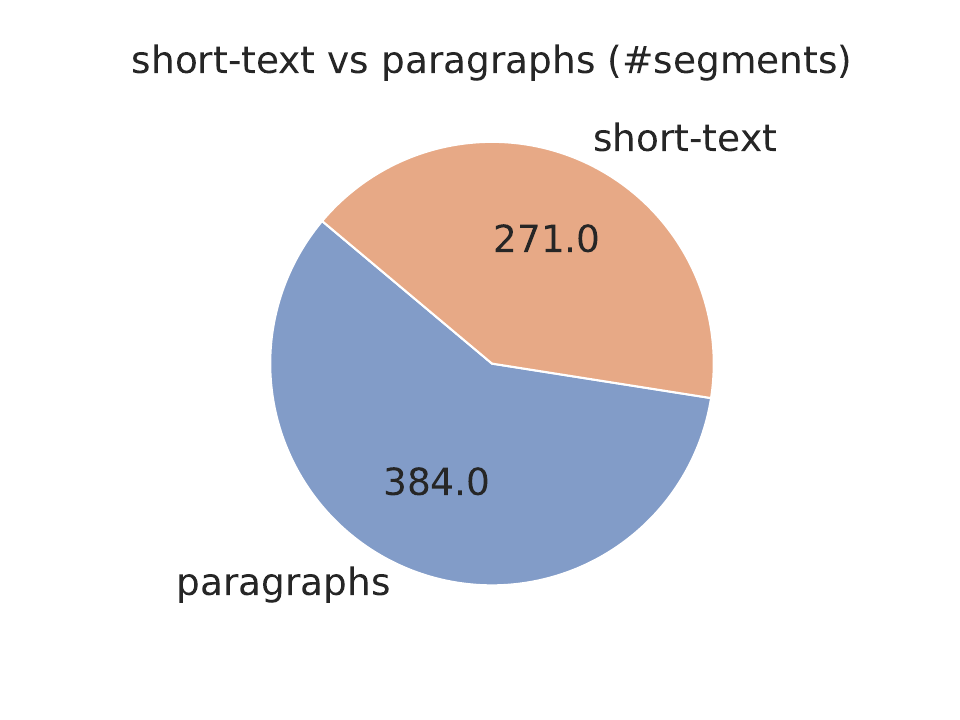}
        &
        \includegraphics[width=0.32\linewidth,trim={1.8cm 3cm 0cm 1cm}, clip]{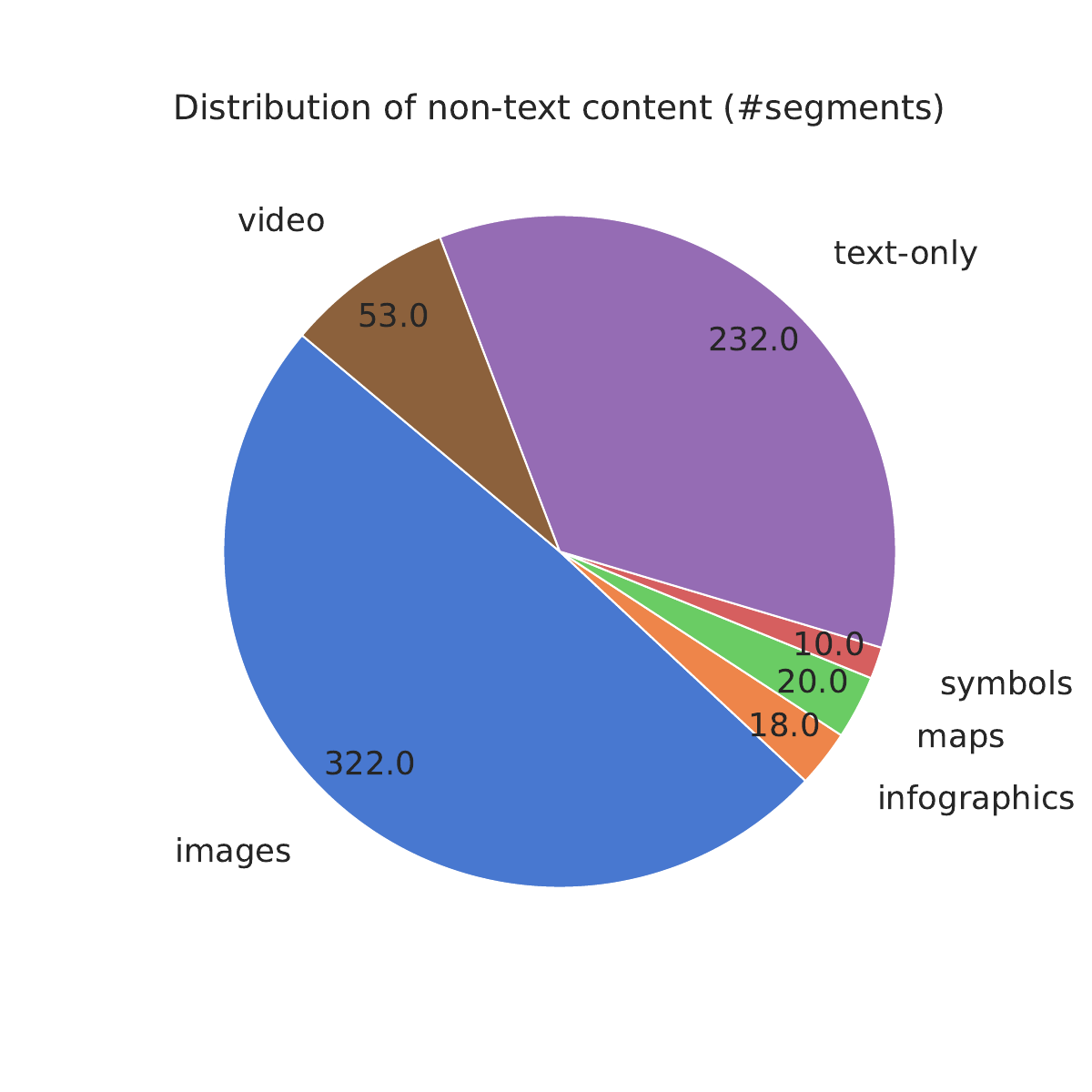}
    \\
      (d)   &  (e) & (f) \\
    \end{tabular}
    }
        \caption{\textbf{Types of reading materials covered.} The top row shows the distributions of (a) medium (b) text types (c) non-text content, in minutes. The bottom row shows the same distributions in terms of number of segments.}
    \label{fig:medium_distribution}
\end{figure}

The Columbus subset includes a diverse range of reading materials designed to evaluate performance across various contexts. To create a diverse dataset, we considered 3 aspects: \textit{medium}, \textit{content length}, and \textit{content type}. Figure~\ref{fig:medium_distribution} illustrates the distribution of different (a) mediums, (b) content lengths and (c) content types.

`Medium' indicates what kind of device or object the subject is reading from. 
The following is a list of reading materials included for different mediums. Similarly to the Seattle subset, we indicate the device (phone, laptop, tablet, screen). For objects, we also indicate the shape of the item which are `cylindrical', `flat', `box', `poster', or `other'.

\begin{itemize}
    \item \textbf{Print:} Books, Booklets, Pamphlets, Spreadsheets, Magazines, Flyers, Instruction Manuals, Handwritten Text, Maps.
    \item \textbf{Digital:} Wikis, Blogs, Forums, Research Articles, News Articles, E-Shopping Sites, Video Sites.
    \item \textbf{Object:} Posters, Nutrition/Product Labels, Flashcards, Signs.
\end{itemize}

`Content length' consists of (1) `paragraphs' where the content requires the subject to read one or more sentences in the text. The scanpath for these usually results in smooth horizontal lines; (2) `short-text' where the text is scattered across the page as short texts. These can include image captions, tabular datasheets, location names on maps, etc. The scan path for short-text is therefore more irregular. 

`Content type' indicates the type of content the user is reading. It can be text-only, where there is only text in the material, or it could be text embedded with other visuals like images, videos, maps, infographics, or symbols.

\begin{figure}[h]
    \centering
    \resizebox{\linewidth}{!}{
    \begin{tabular}{ccc}
        \includegraphics[width=0.32\linewidth,trim={0cm 0cm 0cm 0cm},clip]{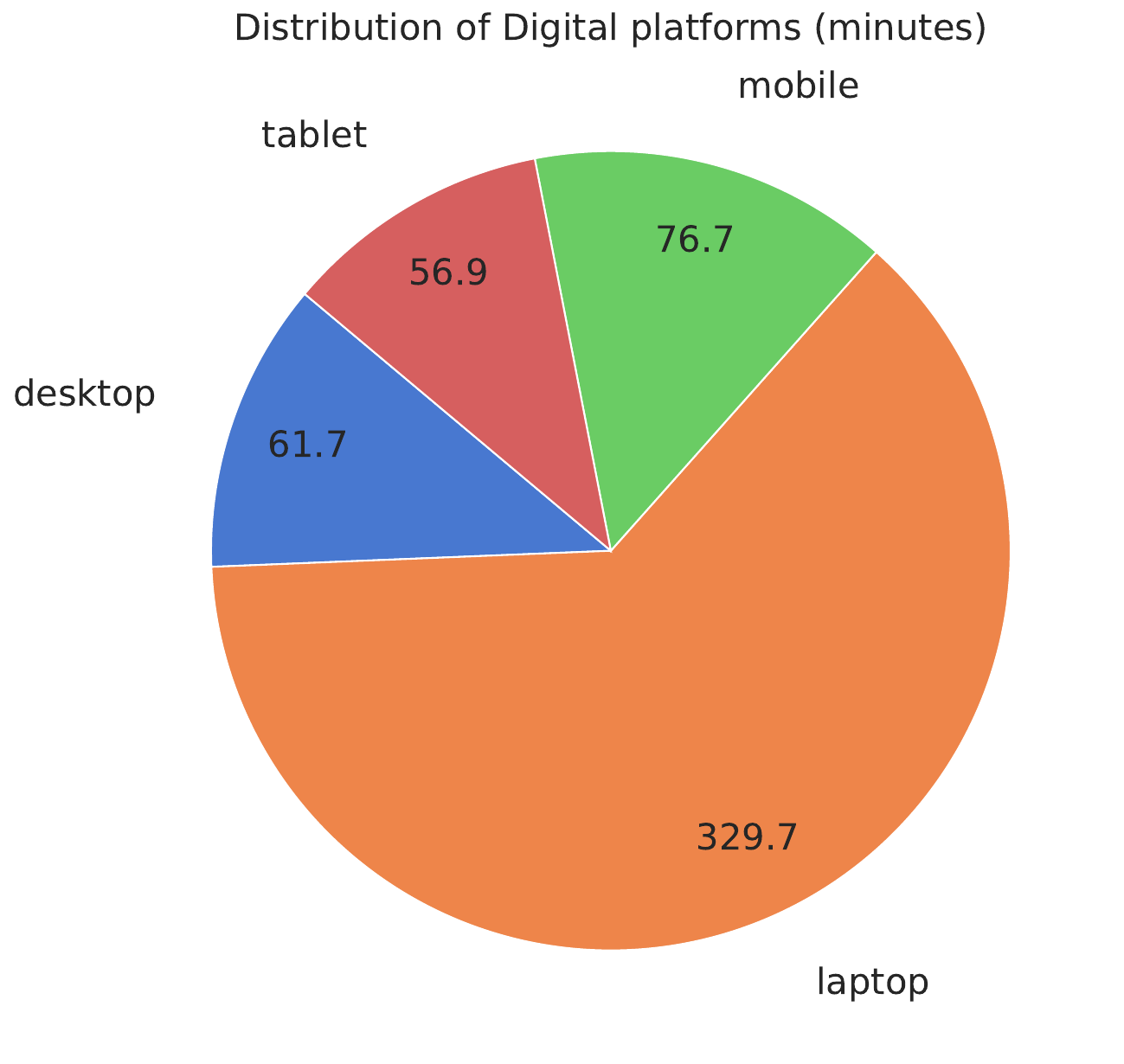} &
        \includegraphics[width=0.33\linewidth,trim={0cm 0cm 0cm 0cm},clip]{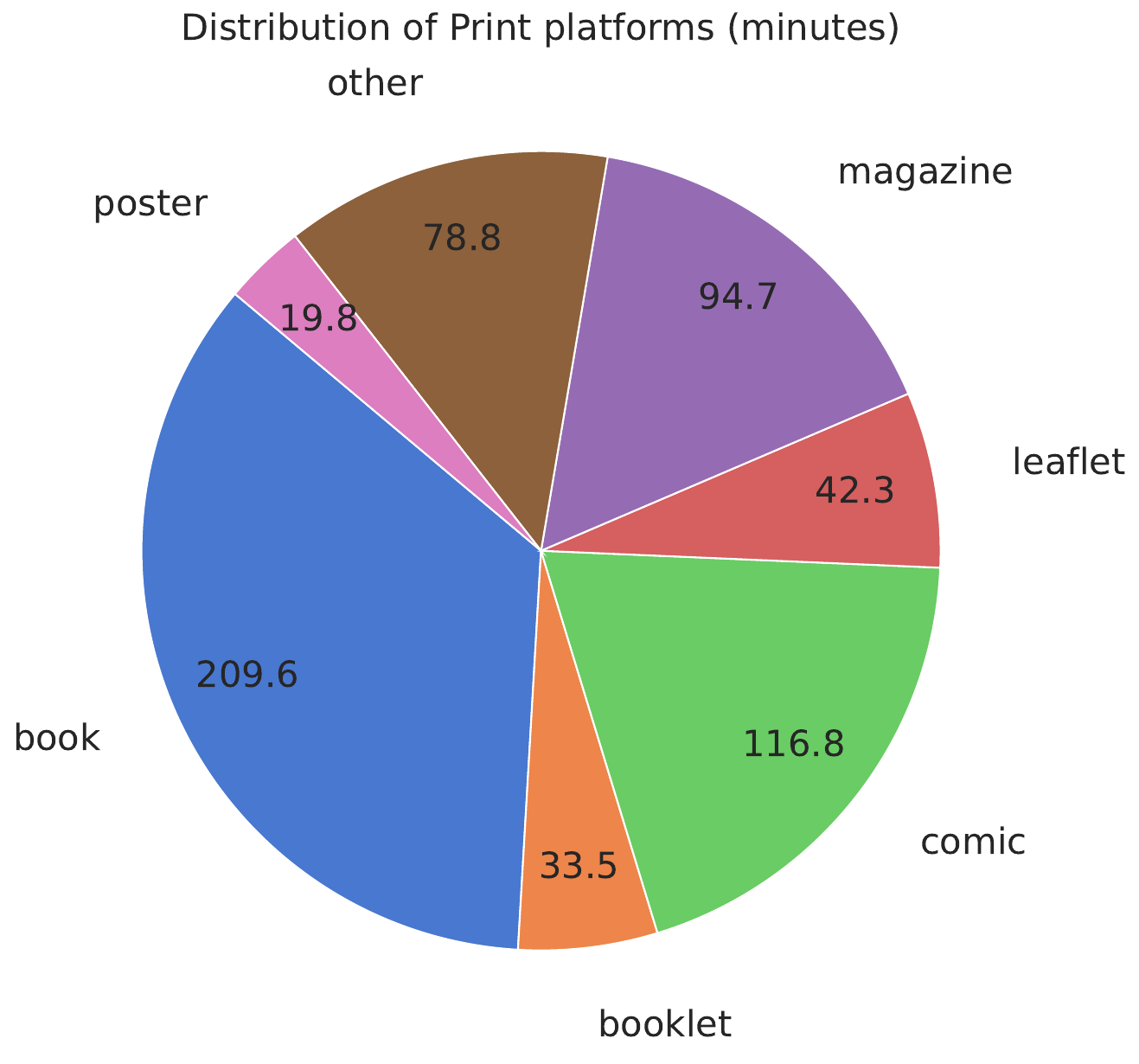} &
        \includegraphics[width=0.31\linewidth,trim={0cm 0cm 0cm 0cm},clip]{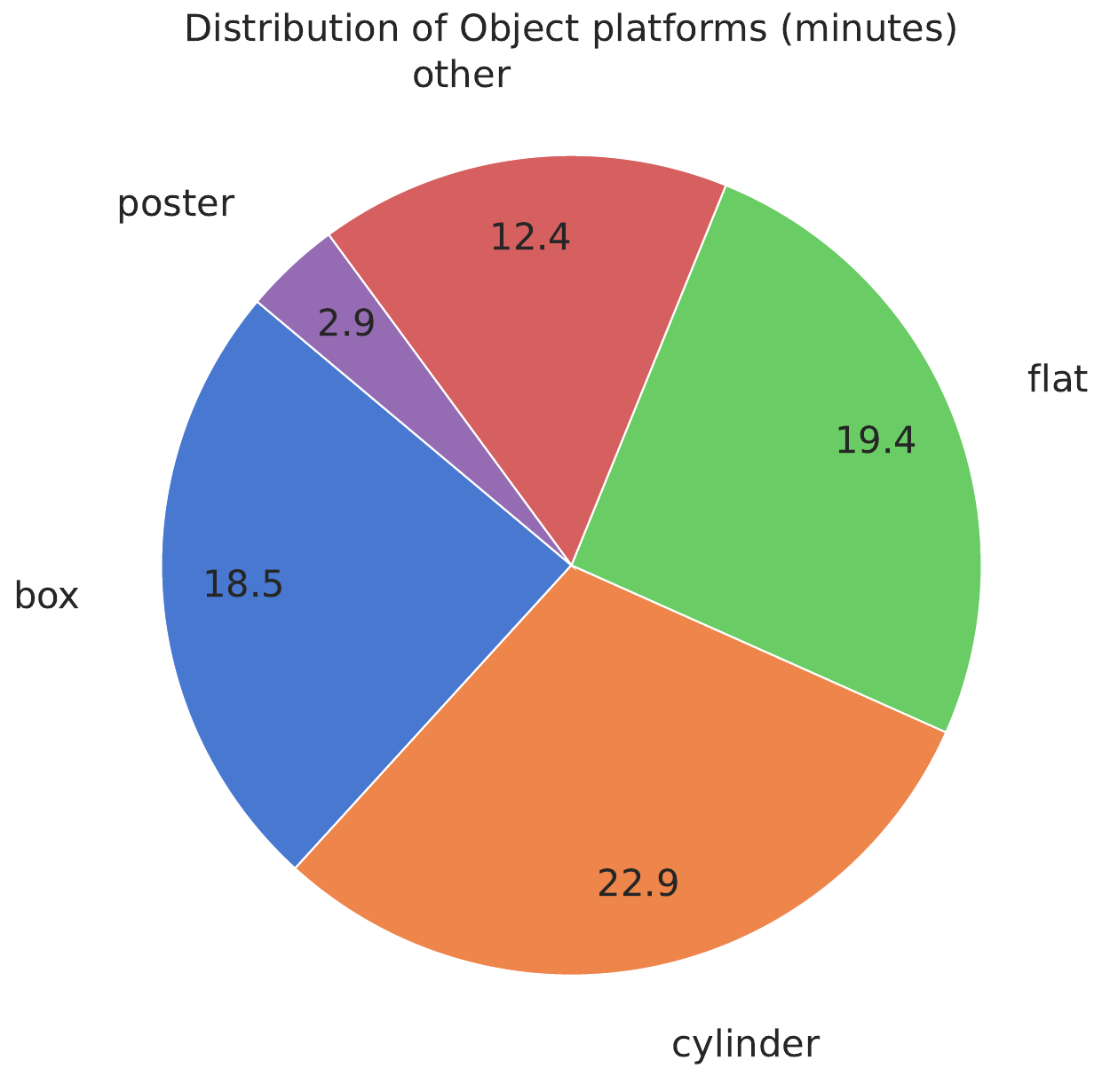} \\
        \footnotesize{(a)} &
        \footnotesize{(b)} &
        \footnotesize{(c)} \\
        \\
        \includegraphics[width=0.31\linewidth,trim={0cm 0cm 0cm 0cm},clip]{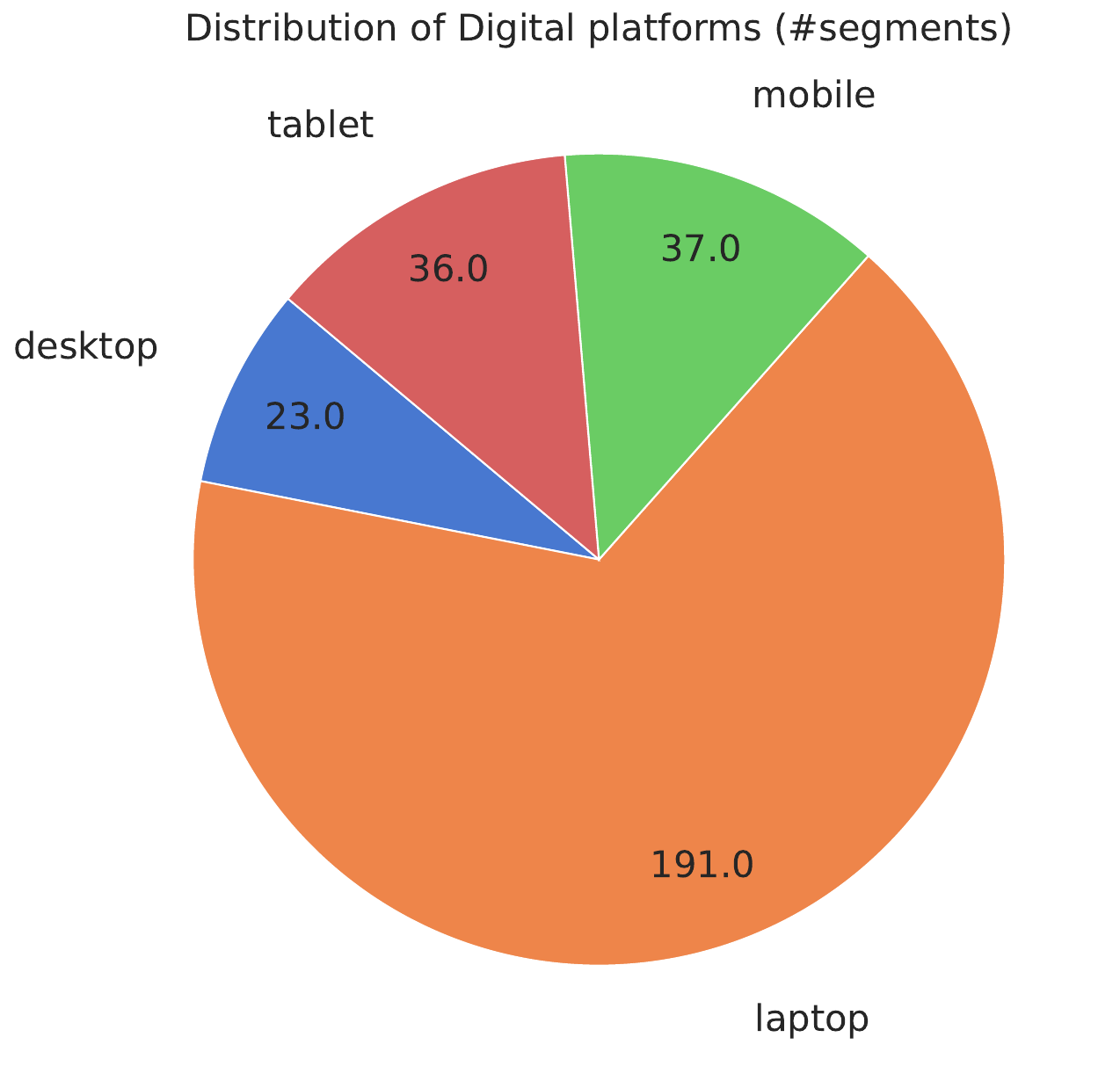} &
        \includegraphics[width=0.33\linewidth,trim={0cm 0cm 0cm 0cm},clip]{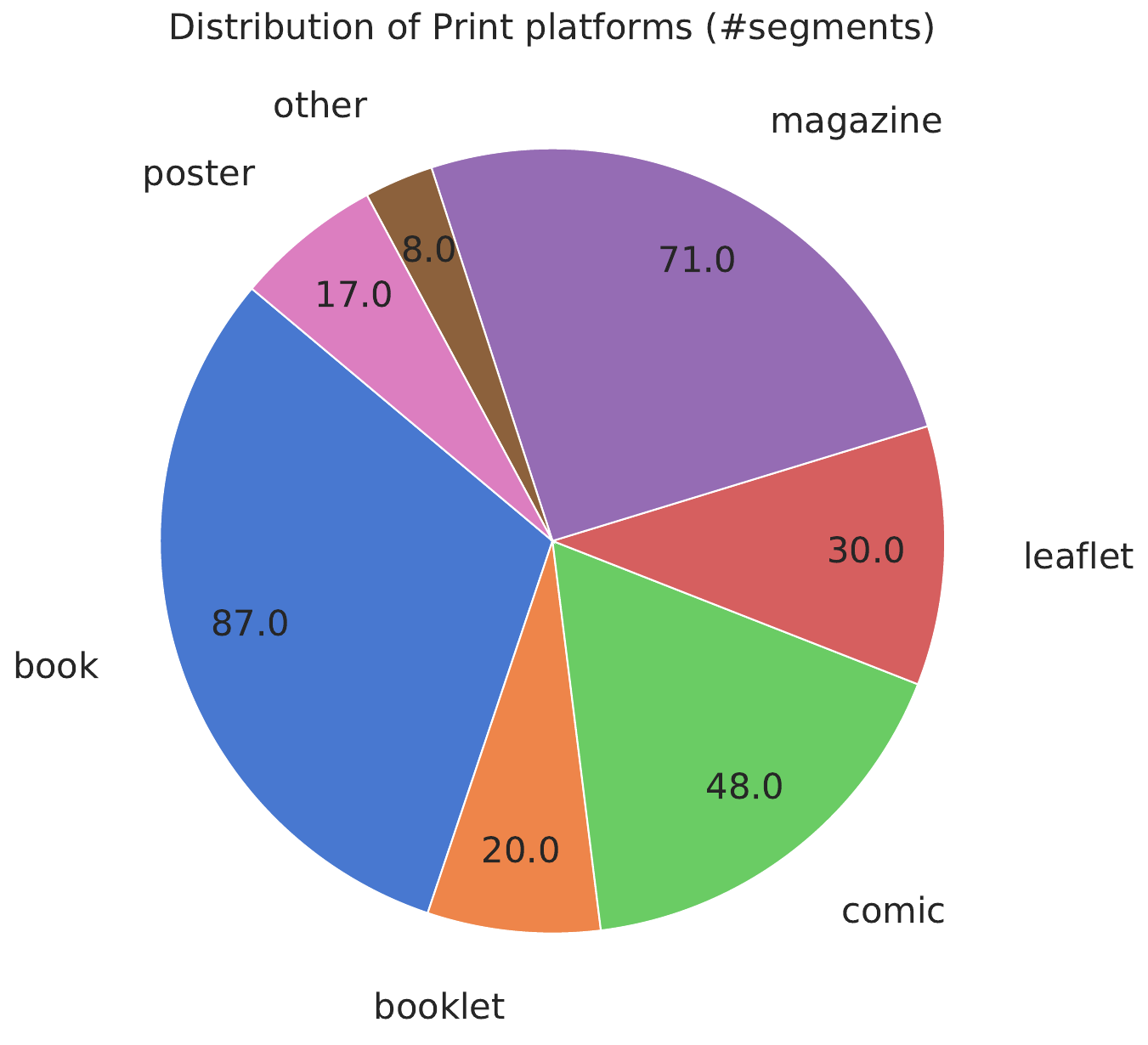} &
        \includegraphics[width=0.30\linewidth,trim={0cm 0cm 0cm 0cm},clip]{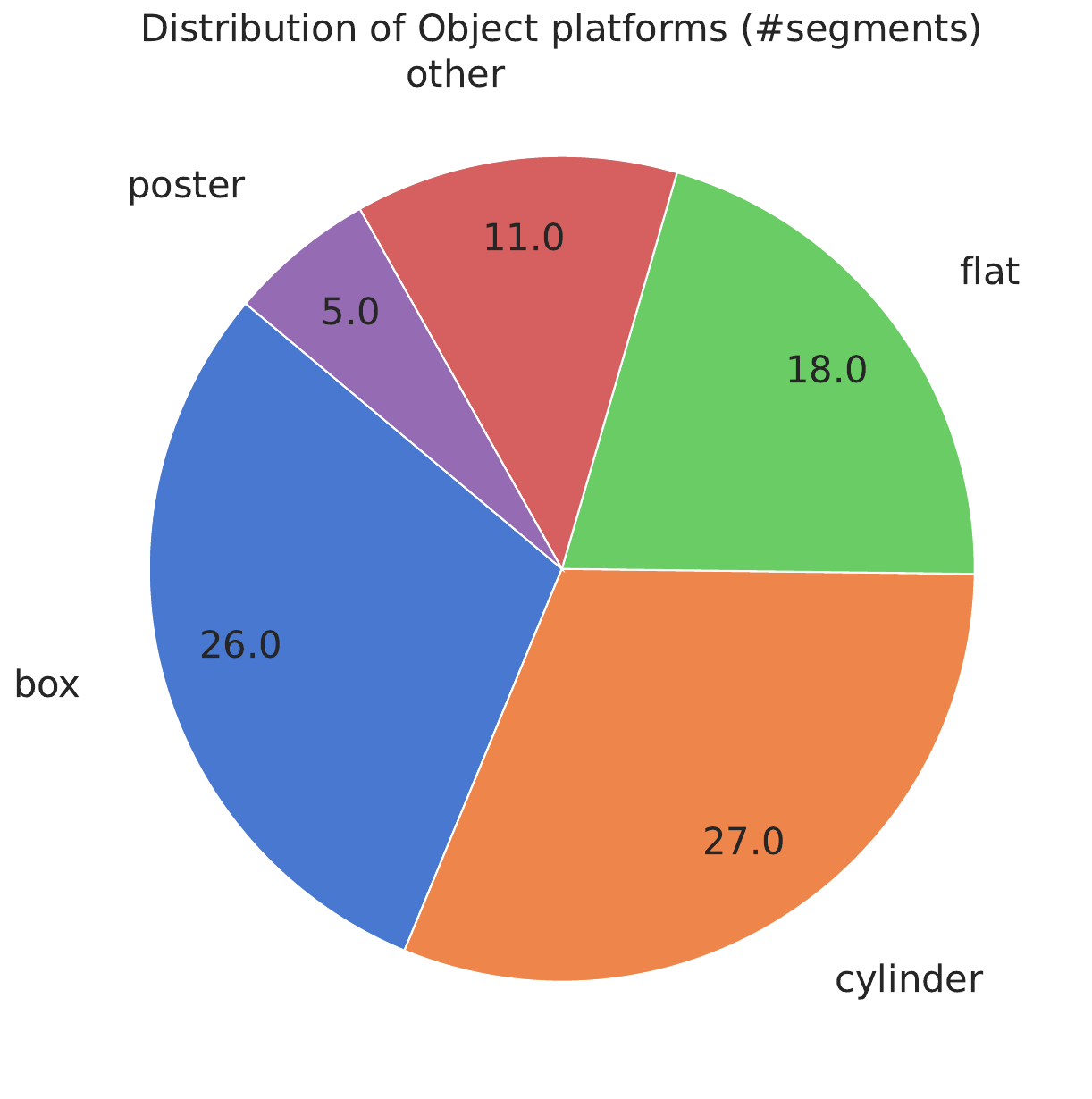} \\
        \footnotesize{(d)} &
        \footnotesize{(e)} &
        \footnotesize{(f)}
    \end{tabular}
    }
    \caption{\textbf{Platform distributions by medium.} The top row shows the total duration in minutes for digital, print, and object platforms, while the bottom row displays the corresponding number of recordings.}
    \label{fig:platform_distribution}
\end{figure}

Additionally, within each medium, the reading material is also varied to create a diverse dataset. The distribution of different platforms for each medium is shown in Figure~\ref{fig:platform_distribution}.

\subsection{Type of reading modes covered}
In the Columbus subset, we have both instances of subject reading or positive cases and not reading or negative cases. However, similar to the Seattle subset, the mode of reading can be different within positive instances. As such we have additional metadata to capture the reading mode. 
There are 2 reading modes in the subset; regular engaged reading and scanning. Engaged reading indicates cases where the subject is asked to read a section of text and is expected to answer a question based on the text. The question is revealed only after completing reading. In contrast, when scanning, the subject is given a question and asked to find the answer in a particular section of text. In the Columbus Dataset, we have 30.6 minutes of scanning data and 786.7 minutes of engaged reading which corresponds to 36 recordings and 435 recordings respectively.

\subsection{Negative data}
Similar to the Seattle Dataset we also collect two types of Negative Data: (1) everyday activities and (2) hard negatives. In everyday activities, subjects are given a general tasks like watching a video or walking in the corridor and asked not to read anything. In this cases there is only minimal or no reading material in the field of view of the subject. For `Hard Negatives', subjects are instructed not to read as well. But in contrast to everyday activity instances, subjects will have significant reading material in their field of view. They are specifically instructed to only focus on non-text content like images. In the data we have 110.4 minutes of everyday activity data and 147 minutes of hard negative data corresponding to 42 and 127 recordings respectively.


\subsection{Mirror setups}
\label{sec:mirror}
The Columbus Dataset contains mirror setups where the reading material is the same. The subject performs two tasks - one engaged reading and the other is either not-reading or scanning. Overall we have 28 such sequence pairs with mirror setups in the subset (25 reading and not-reading pairs and 3 reading and scanning pairs). The reading materials included object and print mediums but no digital mediums. The setup is designed to test the model's discriminative abilities in a symmetrical environment where the primary difference is reading behavior.  

\begin{figure}[h]
    \centering
    \resizebox{\linewidth}{!}{%
\begin{tblr}{
  rowsep=1pt,
  cells = {c},
  row{1} = {m,ht=2cm},
  column{1} = {set1green},
  column{2} = {set1green},
  column{3} = {set1green},
  column{5} = {set1red},
  column{6} = {set1red},
  column{7} = {set1red},
  cell{1}{1} = {c=3}{fg=set1white},
  cell{1}{5} = {c=3}{fg=set1white},
}
\scalebox{5}{\textbf{Reading}} &  &  & & \scalebox{5}{\textbf{Not reading}} \\
\includegraphics{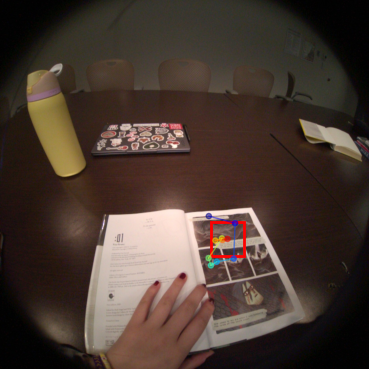} & \includegraphics{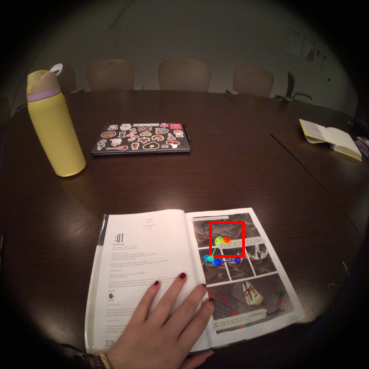} & \includegraphics{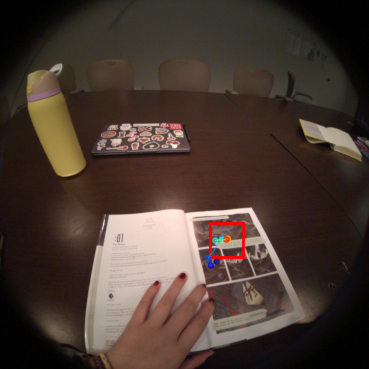} &
&
\includegraphics{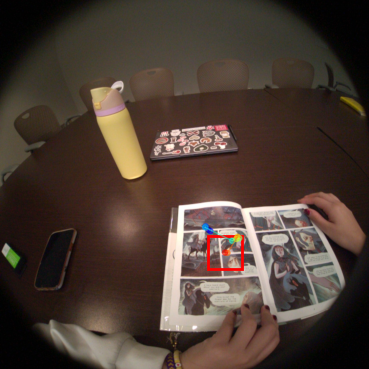} & \includegraphics{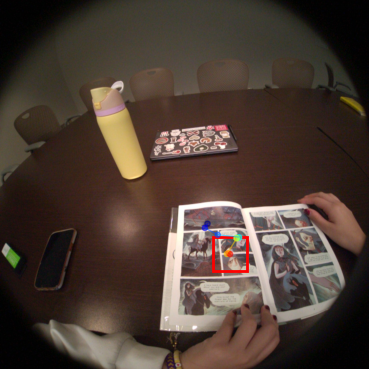} & \includegraphics{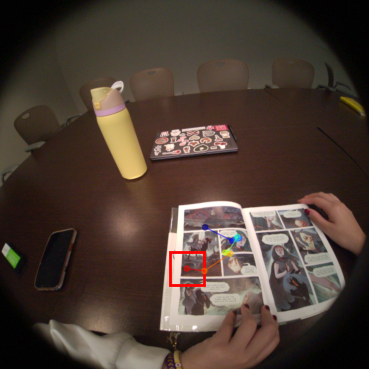}\\
\end{tblr}
}
    \caption{\textbf{Mirror Setup: Print Medium.} Here subject is asked to read a comic vs. when asked to not read anything but instead look at pictures only.}
    \label{fig:mirror-comic}
\end{figure}

\begin{figure}[h]
    \centering
    \resizebox{\linewidth}{!}{%
\begin{tblr}{
  rowsep=1pt,
  cells = {c},
  row{1} = {m,ht=2cm},
  column{1} = {set1green},
  column{2} = {set1green},
  column{3} = {set1green},
  column{5} = {set1red},
  column{6} = {set1red},
  column{7} = {set1red},
  cell{1}{1} = {c=3}{fg=set1white},
  cell{1}{5} = {c=3}{fg=set1white},
}
\scalebox{5}{\textbf{Reading}} &  &  & & \scalebox{5}{\textbf{Not reading}} \\
\includegraphics{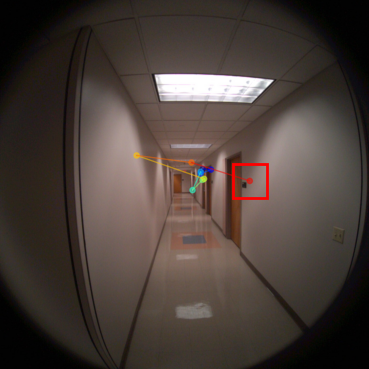} & \includegraphics{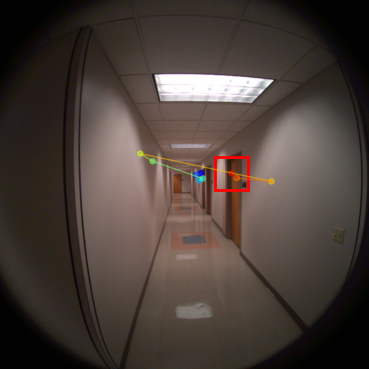} & \includegraphics{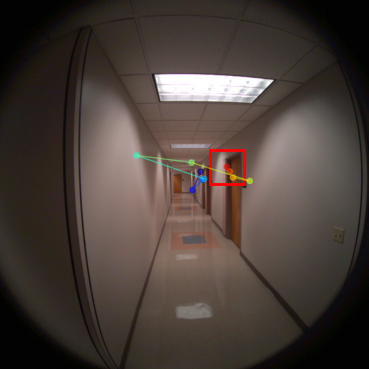} &
&
\includegraphics{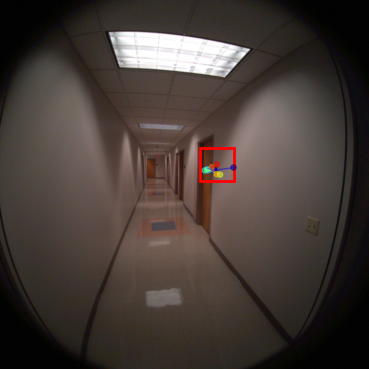} & \includegraphics{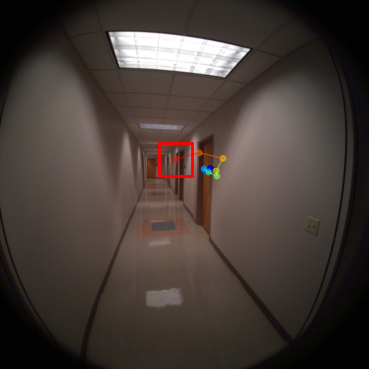} & \includegraphics{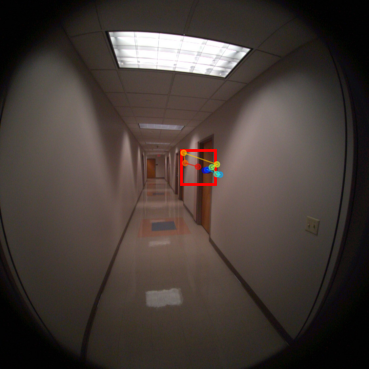} \\

\end{tblr}
}
    \caption{\textbf{Mirror Setup: Walking in Corridor.} Here the subject is asked to read the room numbers and signs in a corridor vs when asked to traverse through normally.}
    \label{fig:mirror-corridor}
\end{figure}

\begin{figure}[h]
    \centering
    \resizebox{\linewidth}{!}{%
\begin{tblr}{
  rowsep=1pt,
  cells = {c},
  row{1} = {m,ht=1.4cm},
  column{1} = {set1green},
  column{2} = {set1green},
  column{3} = {set1green},
  column{5} = {set1blue},
  column{6} = {set1blue},
  column{7} = {set1blue},
  cell{1}{1} = {c=3}{fg=set1white},
  cell{1}{5} = {c=3}{fg=set1white},
}
\scalebox{4}{\textbf{Reading (Engaged)}} &  &  & & \scalebox{4}{\textbf{Reading (Scanning)}} \\
\includegraphics{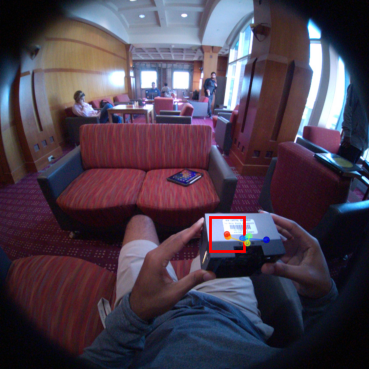} & \includegraphics{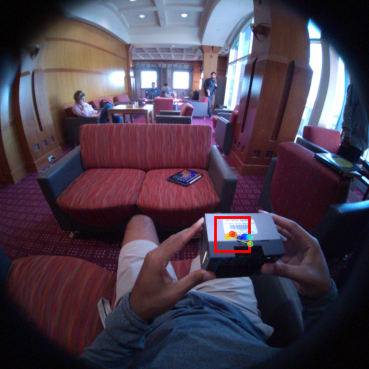} & \includegraphics{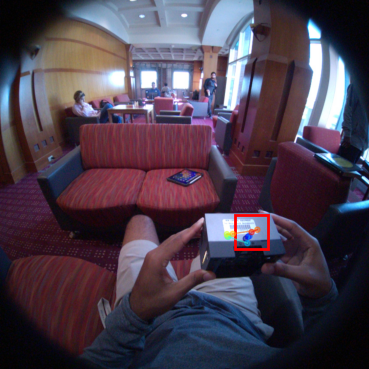} &
&
\includegraphics{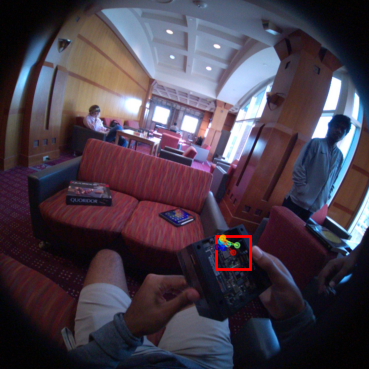} & \includegraphics{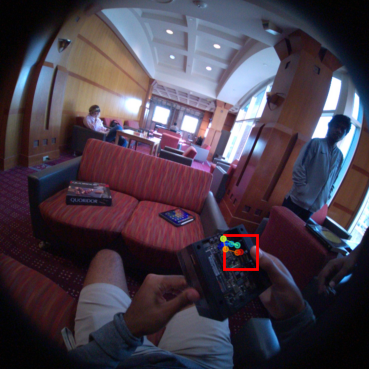} & \includegraphics{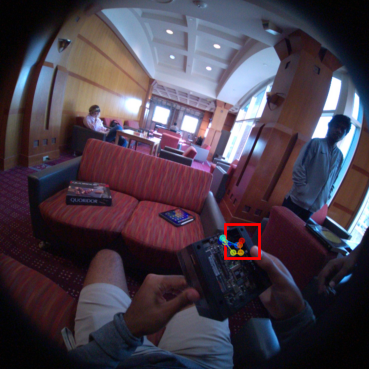}\\
\end{tblr}
}
    \caption{\textbf{Mirror Setup: Reading vs Searching.} Here the subject is asked to read the serial numbers in a circuit board vs when asked to specifically search and count the number of resistors.}
    \label{fig:mirror-search}
\end{figure}

Figures~\ref{fig:mirror-comic}, \ref{fig:mirror-corridor}, and \ref{fig:mirror-search} illustrate three distinct mirror setups from the Columbus Dataset, each designed to evaluate the model's performance in distinguishing between reading and non-reading tasks under controlled conditions. The reading materials remain identical, and variations arise solely from the task being performed. We analyze the performance across 3 frames on Gaze only, RGB Only, Gaze+RGB, and Gaze+RGB+IMU modality combinations. The RGB crop used by the model is shown in the red box on the figures and the gaze pattern is indicated by the connected dots with the red dot being the last and most recent gaze point and blue being the start of the sequence. Note that the gaze pattern fed to the model spans 2s with 120 gaze points in total. However, in the figures only 8 dots are shown which are 250ms apart. 

\subsection{Annotation}
\begin{figure*}[h]
  \centering
  \includegraphics[width=.9\linewidth]{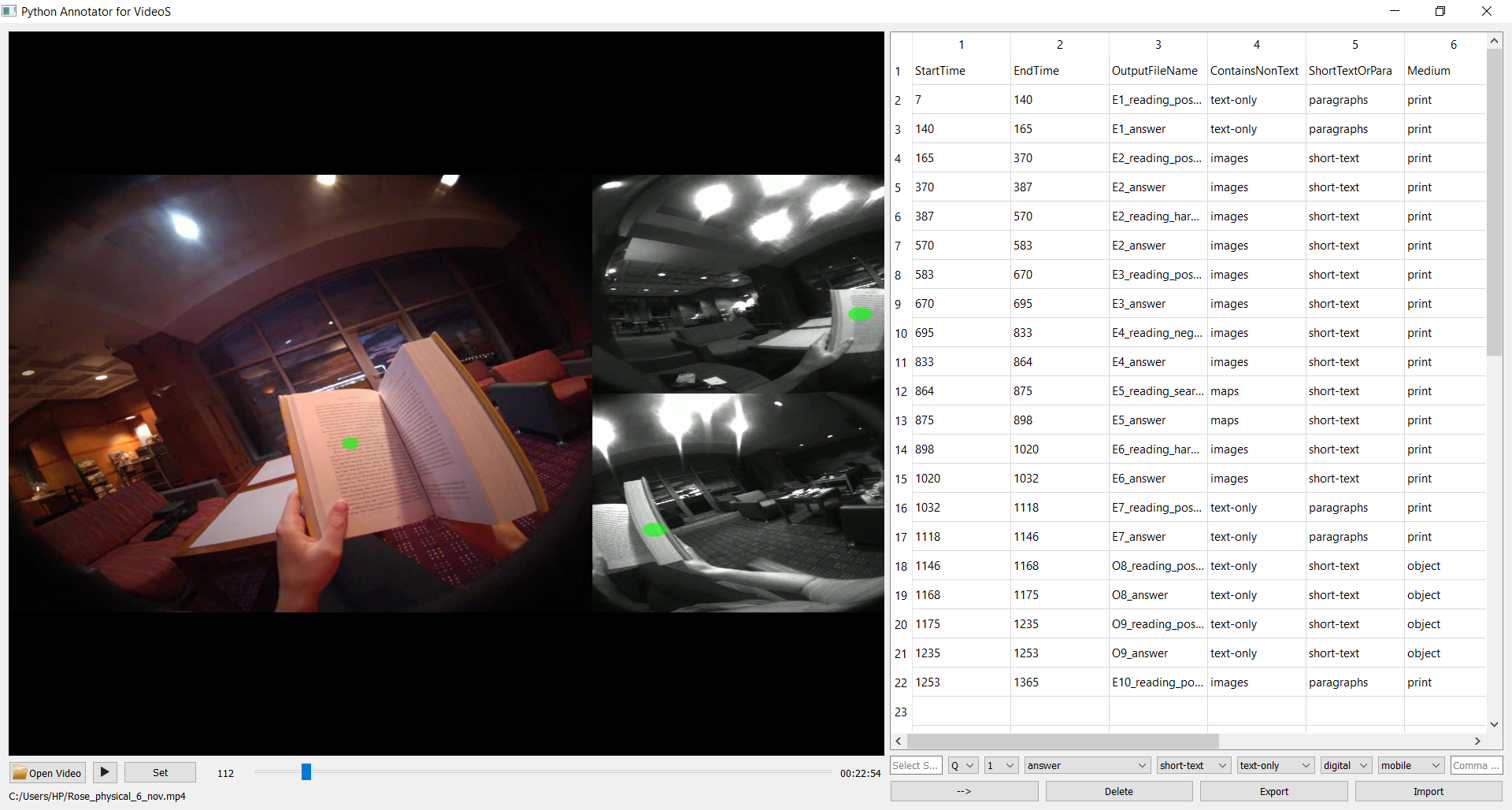}
  \caption{\textbf{Graphical interface of annotation tool.} The user interface consists of the RGB video preview with sound and the gaze point overlaid on top shown in green. The interface allows adjusting the start and end time of each recording as well as annotating metadata like content-type, content-length and medium.}
  \label{fig:annotation_pipeline}
\end{figure*}

To annotate the raw data and extract segments, we utilized a labeling tool developed in PyQt5.
Figure~\ref{fig:annotation_pipeline} outlines the graphical user interface of the annotation tool.
The annotation process begins by importing the raw video file, which is then manually input the timing information using the timeline slider and \texttt{'set'} button. For each segment, we record the start and end times in seconds to mark the exact duration the segment appears in the video. To ensure a clean transition between segments, we trim a small portion from the beginning and end of each segment, avoiding any unintended overlap or transition frames.
Each segment corresponds to a recording in the subset. 
To ensure easy and consistent selection, we use drop-down menus to store the values for `Medium', `Content-Type', `Platform', and `Activity Type'. After annotating each video, segment information is saved in a separate CSV file for further processing.

\subsection{Demographics}

We posted flyers in various locations to inform interested people about the study. Participants reached out to us through email and a session was booked for them based on availability.
Among the 30 participants, 15 were male and 15 were female. The age of the participants ranged from 18 to 34 with 13 participants between 18-24 years old and 17 between 25-34. 12 participants were native English speakers and 18 were non-native speakers. Participants came from a wide range of backgrounds and had at least a high school degree. To be eligible, participants had to be able to read without glasses. Of the participants, 21 had perfect vision and 7 wore contacts. 2 participants had mild myopia but did not require glasses to read. 

\begin{figure}[h]
    \centering
\resizebox{.72\linewidth}{!}{%
\begin{tblr}{
  cells = {c},
  row{2} = {t},
  cell{1}{1} = {t},
  cell{1}{2} = {t},
  cell{1}{3} = {r=8}{},
}
\textbf{Age Range} & \textbf{Gender} & 
\begin{tabular}{c}
\textbf{Education} \\
\begin{tikzpicture}
\pie[
  text=inside,
  radius=2.5,
  explode=0.0,
  color={set1red, set1green, set1blue, set1orange, set1purple},
  after number=\%,
  style={very thin}
]{13.3/, 20.0/, 36.7/, 26.7/, 3.3/}
\end{tikzpicture} \\
\begin{tabular}{@{}l@{}}
\textcolor{set1red}{\rule{10pt}{10pt}} High School\\[0.5ex]
\textcolor{set1green}{\rule{10pt}{10pt}} Some College Education\\[0.5ex]
\textcolor{set1blue}{\rule{10pt}{10pt}} Bachelors Degree\\[0.5ex]
\textcolor{set1orange}{\rule{10pt}{10pt}} Graduate Degree\\[0.5ex]
\textcolor{set1purple}{\rule{10pt}{10pt}} Prefer Not to Say
\end{tabular} \\
(v)
\end{tabular}
\\[1ex]
\begin{tikzpicture}
\pie[
  radius=2.5,
  explode=0.0,
  color={set1red, set1blue},
  text=inside,
  after number=\%,
  style={very thin}
]{43.3/, 56.7/}
\end{tikzpicture} & 
\begin{tikzpicture}
\pie[
  radius=2.5,
  explode=0.0,
  color={set1red, set1blue},
  text=inside,
  after number=\%,
  style={very thin}
]{48.4/, 51.6/}
\end{tikzpicture}
& \\
\begin{tabular}{@{}l@{}}
\textcolor{set1red}{\rule{10pt}{10pt}} 18--24\\[0.5ex]
\textcolor{set1blue}{\rule{10pt}{10pt}} 25--34
\end{tabular}
& 
\begin{tabular}{@{}l@{}}
\textcolor{set1red}{\rule{10pt}{10pt}} Male\\[0.5ex]
\textcolor{set1blue}{\rule{10pt}{10pt}} Female
\end{tabular}
& \\
(i) & (ii) & \\[1ex]
\textbf{Native vs Non-native Speakers} & \textbf{Visual Aid Requirements} & \\
\begin{tikzpicture}
\pie[
  radius=2.5,
  explode=0.0,
  color={set1red, set1blue},
  text=inside,
  after number=\%,
  style={very thin}
]{38.7/, 61.3/}
\end{tikzpicture}
& 
\begin{tikzpicture}
\pie[
  radius=2.5,
  explode=0.0,
  color={set1blue, set1red, set1yellow},
  text=inside,
  after number=\%,
  style={very thin}
]{67.7/, 25.8/, 6.5/}
\end{tikzpicture} & \\
\begin{tabular}{@{}l@{}}
\textcolor{set1red}{\rule{10pt}{10pt}} Native Speakers\\[0.5ex]
\textcolor{set1blue}{\rule{10pt}{10pt}} Non-Native Speakers
\end{tabular}
& 
\begin{tabular}{@{}l@{}}
\textcolor{set1blue}{\rule{10pt}{10pt}} None\\[0.5ex]
\textcolor{set1red}{\rule{10pt}{10pt}} Contacts\\[0.5ex]
\textcolor{set1yellow}{\rule{10pt}{10pt}} Mild Myopia
\end{tabular}
& \\
(iii) & (iv) & 
\end{tblr}
}
    \caption{\textbf{Demographic statistics of the Columbus subset.} 
    An overview of the demographic distribution is shown: 
    (i)~Age Range, 
    (ii)~Gender, 
    (iii)~Native vs Non-native Speakers, 
    (iv)~Visual Aid Requirements, 
    and (v)~Education.}
    \label{fig:demographics}
\end{figure}
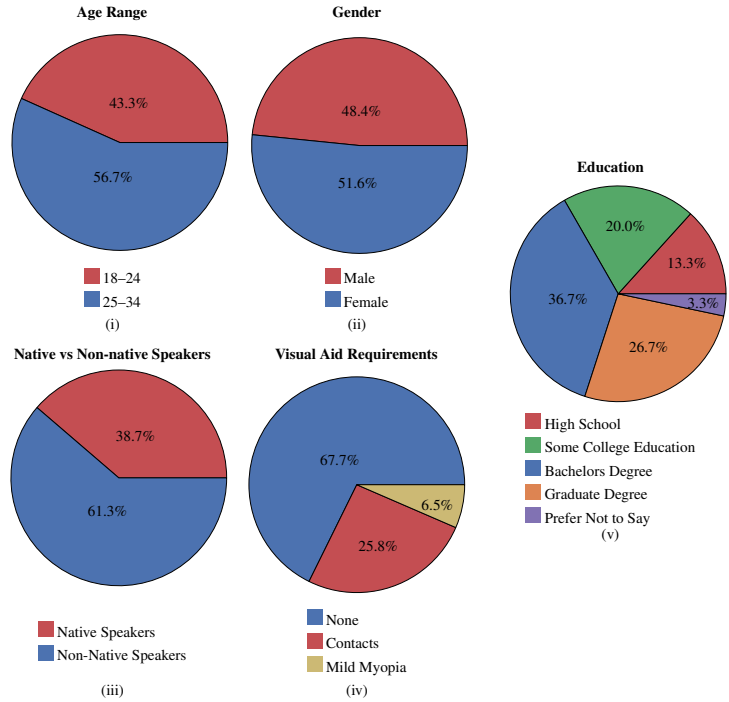

\subsection{Metadata}
The Columbus subset contains metadata related to demographics, tasks, and scenario details.

\begin{figure}[h]
\begin{mdframed}
\footnotesize
\begin{verbatim}
"name":"<name-of-recording>",
"is-reading": True,
"mode": "scanning",
"negative-type": "N/A"
"scenario": {
    "contains-non-text": True,
    "short-text-or-para": "Paragraphs",
    "platform": "Laptop",
    "mode": "scanning",
    "ExtraTags": [
        "Had contacts",
        ],
},
"subject": {
    "SubjectID": "12345",
    "AgeRange": "25-34",
    "Gender": "Male",
    "EducationLevel": "Graduate",
    "Corrective Lenses": "Contacts",
    "Specialization": "Computer Science",
    "Country": "USA"
}
\end{verbatim}
\end{mdframed}
\caption{\textbf{Example metadata for Columbus subset.} The metadata contained is slightly different than that of the Seattle subset, allowing for different directions to explore for further research.}
\label{tab:metadata_example_columbus}
\end{figure}

\subsection{Protocols}
For the Columbus subset, we follow a single protocol which was approved by IRB. We follow this protocol which involves both pre-session preparations, collection of data and final processing steps. The protocol steps are discussed as follows:
\subsubsection{Pre-session Briefing}
Before starting session, we brief the subject on the motivation and objective of the study, a description of the kind of reading material we will provide, an overview of all the sessions, and the nature of tasks like reading, hard negatives, mirror setup tasks, etc. We provide a consent form asking the participant to acknowledge consent for the study and their rights as a participant. Once the paperwork is done and the participant is briefed and ready, we begin data collection.

\subsubsection{Data collection}
The data collection process for this study consisted of three primary sessions, with a total duration of approximately one hour. The initial two sessions (digital medium session and print/object medium session) took place indoors inside a room, while the optional third session was conducted on corridors or balconies if time and participant availability permitted.  The collection protocol is discussed below.

\begin{flushleft}

\textbf{Pre-session Preparation} The process began with participants filling out a demographic questionnaire in a controlled indoor environment. Following this, an attendant explained how to calibrate the Aria glasses, and once participants were familiar with the procedure, calibration was performed. After this, we proceeded with the first session. 

\textbf{Digital Medium Session.} The session generally involved interacting with digital media: participants were directed via a Qualtrics survey to visit specific websites and complete tasks such as reading certain sections, searching for information, or browsing without reading. The tasks varied, and each participant received different instructions. Responses to these tasks were then collected to assess comprehension or adherence to instructions. It is important to note that any recordings where participants inadvertently read text, contrary to instructions, were discarded or trimmed. Throughout the session, each participant used one specific device (laptop, tablet, mobile, or desktop), although they were permitted to bring their own device. The survey recorded the completion time for each question, aiding in the segmentation of the recording for later analysis. The digital medium session was generally conducted first, but in a few cases, we started with the print and object medium session. 

\textbf{Print/Object Medium Session.} After a brief rest period of 5-10 minutes, the second session commenced, focusing on print and object mediums. Participants were provided with various printed materials, such as books, magazines, flyers, and instruction manuals, or objects with embedded text like product packaging. They were instructed to read specified sections, and the start and end times of their reading were manually logged by an attendant for segmenting the session later. Subsequently, participants answered questions to verify their engagement with the text. 

\textbf{Impromptu session} If there is time remaining, we then arrange the last session. This is a short session in which we ask the participants to do random tasks which include both reading and not reading. These tasks could be reading a wall-mounted poster or flyer, walking along the corridor reading room numbers or signs, or reading a wall-mounted map. Like the previous two sessions, participants may be instructed to either read or not read any text. 
\end{flushleft}

\subsubsection{Post-processing}
After the sessions, we generate draft previews of each session for segmentation. The previews are low quality RGB videos of the session with the gaze overlaid on the videos. We get the approximate segment break points from the survey data. For each session we manually adjust the start and end times of each segment. Additionally, we manually annotate medium, platform, content-length, content-type, and language for each segment at this time by looking at the previews. We also add extra tags we think may be of importance. We save the manually adjusted timing data and use it to segment and export the data into individual recordings containing only the relevant data, excluding the answering and instruction phases. We process these recordings using a cloud service to get the eye-tracking annotations and open-loop trajectory based on the SLAM data. We review the generated recordings and annotations and do another pass on the manual annotations to make sure the segmentation and metadata are correct.

\subsubsection{De-identification}
To conform with University policy and the guidelines approved by the IRB, we are required to de-identify any descriptions or data that may identify the participant. We deidentify any descriptions, text and filenames containing the participant's identifiable data. Despite precautions, some faces of the attendant and bystanders may be present in the video recordings. We use EgoBlur~\cite{raina2023egoblurresponsibleinnovationaria} to  blur out any such faces that may have been captured by the glasses in the segmented recordings.

\clearpage
\section{Supplementary Results on Columbus Subset}
\label{sec:columbus_supplementary_results}
\subsection{Main Results}

\begin{table}[h]
\centering
{\begin{tblr}{
  cells = {c},
  hline{1-2,5,8-9} = {-}{},
  vline{4,8} = {-}{}
}
\textbf{Gaze} & \textbf{RGB} & \textbf{IMU} & \textbf{Acc} & \textbf{F1} & \textbf{AUC} & \textbf{P\textsubscript{R=90.0}} & \textbf{T\textsubscript{R=90.0}} & \textbf{Acc\textsubscript{R=90.0}} & \textbf{F1\textsubscript{R=90.0}} \\
\checkmark &  &  & 77.1 & 84.0 & 95.3 & 84.1& 29.2 & 79.1 & 86.9\\
  & \checkmark &  & 76.7 & 84.5 & 92.1 & 83.4& 29.4 & 78.5 & 86.6\\
  &  & \checkmark & 71.4 & 82.2 & 81.9 & 78.5& 42.3 & 73.3 & 83.9\\
 \checkmark &  & \checkmark & 76.7 & 84.0 & 94.9 & 82.7& 26.6 & 77.8 & 86.2\\
  & \checkmark & \checkmark & 77.8 & 85.6 & 92.4 & 83.6& 36.9 & 78.6 & 86.7\\
\checkmark & \checkmark &  & 82.8 & 88.7 & \textbf{96.4} & \textbf{88.2}& 40.3 & \textbf{83.0} & \textbf{89.1}\\
 \checkmark & \checkmark & \checkmark & \textbf{82.9} & \textbf{88.8} & \textbf{96.4} & \textbf{88.2} & {42.4} & 82.9 & \textbf{89.1}
\end{tblr}
}
\vspace{2mm}
\caption{\textbf{Results on the Columbus subset.} The table includes different combinations of these modalities (using only single modality, using any combination of two modalities, and using all modalities). The first four metrics are: accuracy (Acc), F1 score (F1), area under the curve (AUC), and the Precision at Recall of 90\% (P\textsubscript{R=90.0}). Consistent with Seattle subset in the main paper, we show the usefulness of combining gaze and RGB. IMU becomes less useful with less daily activities. In the last three columns, the confidence threshold {T\textsubscript{R=90.0}} for binary classification is set where Recall is equal to 90\%. We report the accuracy metrics at this confidence threshold. We notice lowered threshold to reach such high recall, adversely affecting the precision in the process.}
\label{tab:main_results_columbus_zeroshot}
\end{table}


\begin{figure}[h]
\centering

\begin{tabular}{cc}
        \includegraphics[width=0.45\linewidth]{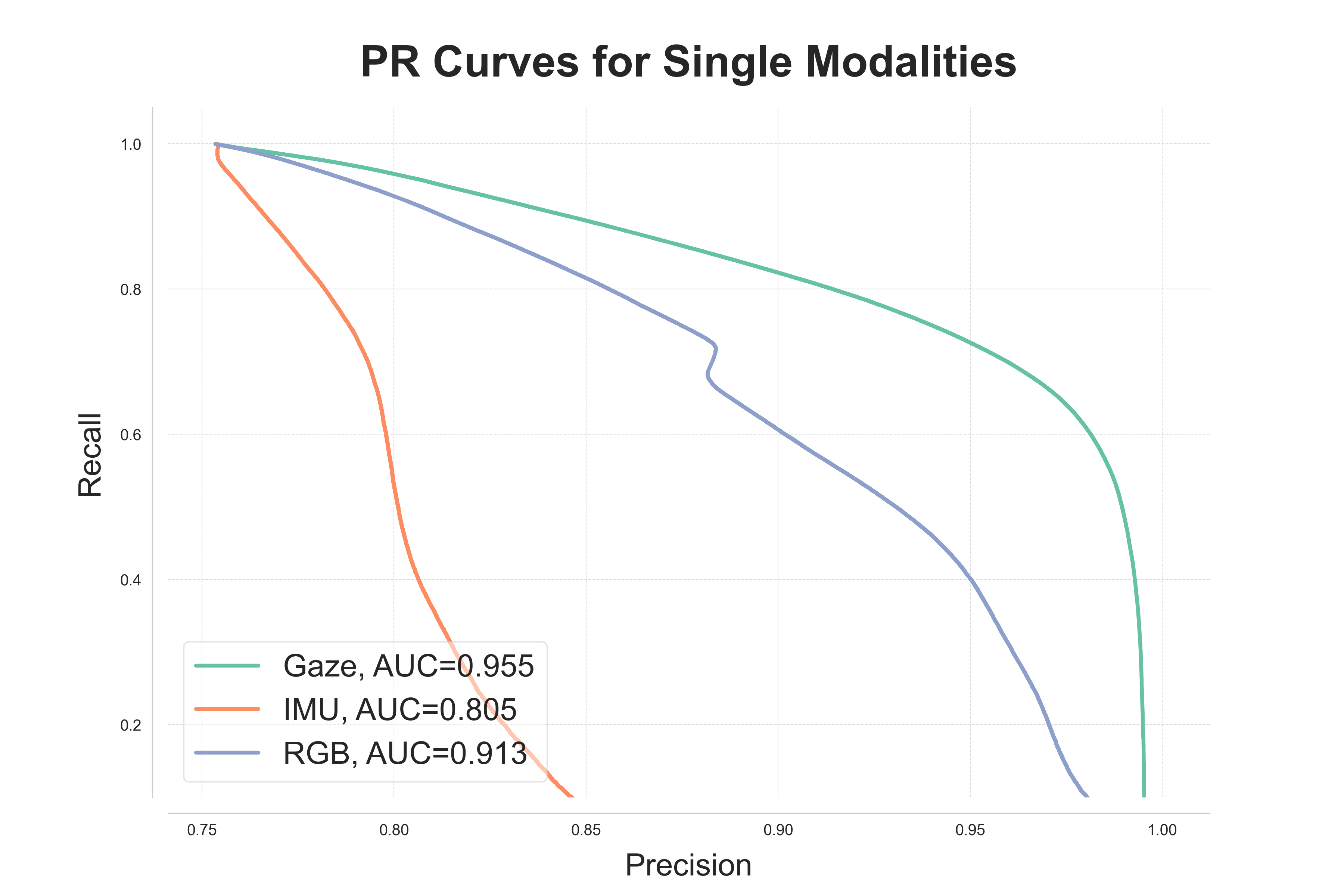}
 &     \includegraphics[width=0.45\linewidth]{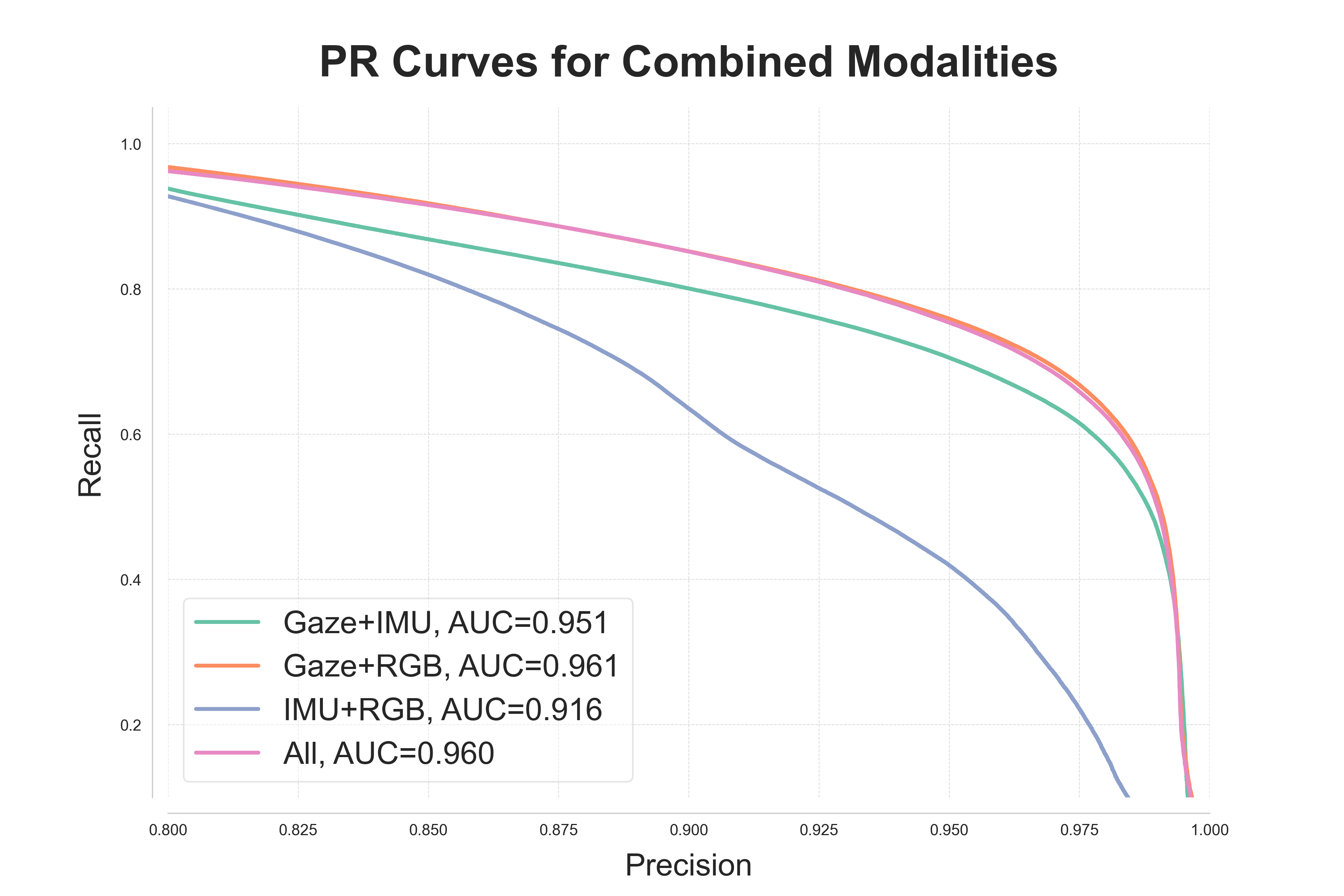}
 \\
   (a)  & (b)
\end{tabular}
    \caption{\textbf{PR Curves for different modalities} We compare the PR curves of different modalities. (a) shows the curves for individual modalities and (b) compares how combining different modalities influence performance}
    \label{fig:enter-label}
    \vspace{-8pt}
\end{figure}

The results presented in Table~\ref{tab:main_results_columbus_zeroshot} highlight the effectiveness of various modalities—Gaze, RGB, and IMU - in detecting reading activity on the Columbus Dataset. 
We make some key observations:

\textbf{Single Modality:} Firstly, we find that the Gaze alone achieves the highest performance among single-modality setups.
This is reasonable as gaze data provides the most discriminative features for detecting reading behavior. Given the larger proportion of hard negatives and smaller daily activities, we observe a drop in performance for RGB and IMU compared to the Seattle subset.


\textbf{Combined Modalities:} Consistent with the Seattle subset, we find that combining different modalities yield better results, with the Gaze+RGB+IMU model giving best performance.
However, we notice that the contribution of IMU data is relatively marginal in this subset when gaze and RGB features are already present. This is reasonable as the Columbus subset consists mostly of engaged reading tasks where the subject seldom moves their head. 

 \textbf{Confidence threshold tuning:} In contextual AI use cases, we expect the model to be prioritizing recall over precision to minimize false negatives while accepting some false positives. Hence, the last three columns of the table presents the threshold-tuned results for Gaze used individually and in combination with RGB and IMU modalities, where the confidence threshold for binary classification is set where recall is equal to 90\%. This results in some trade-off in precision in the process.






\subsection{Result breakdown}

\begin{figure}[h!]
    \centering
    \includegraphics[width=0.48\linewidth]{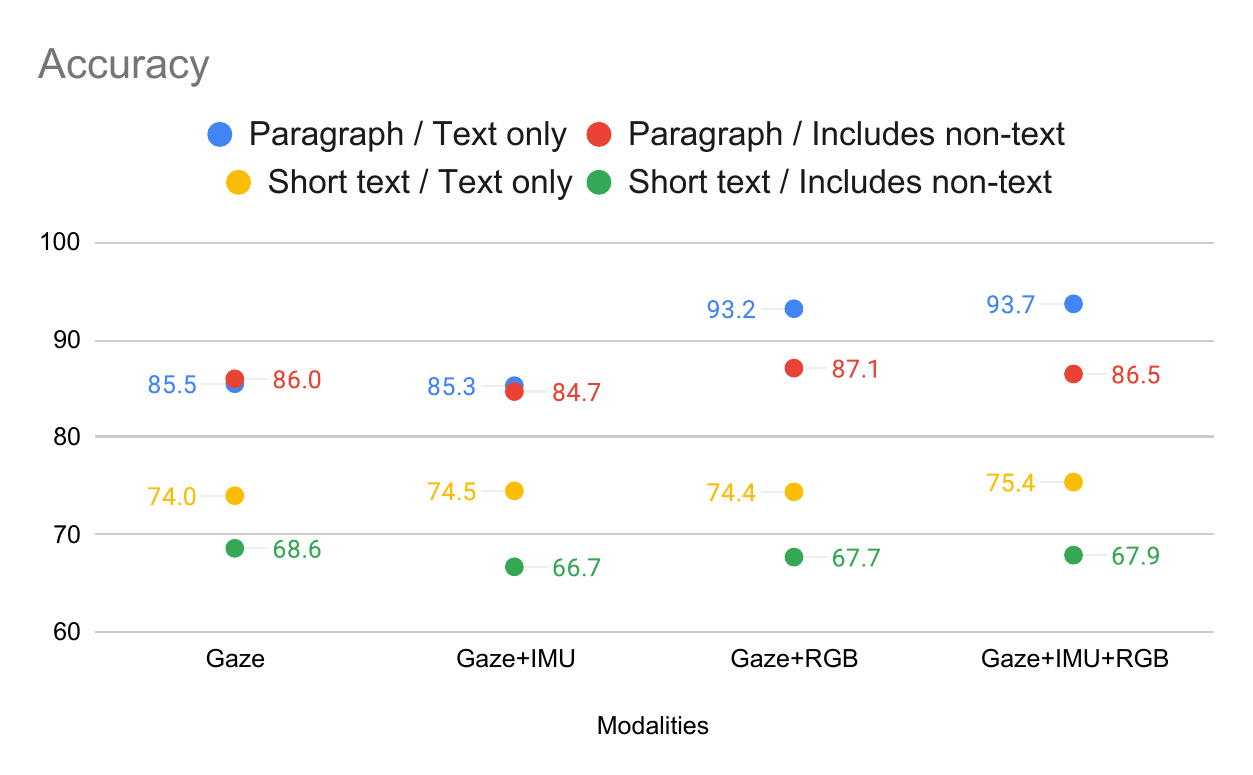}
    \includegraphics[width=0.48\linewidth]{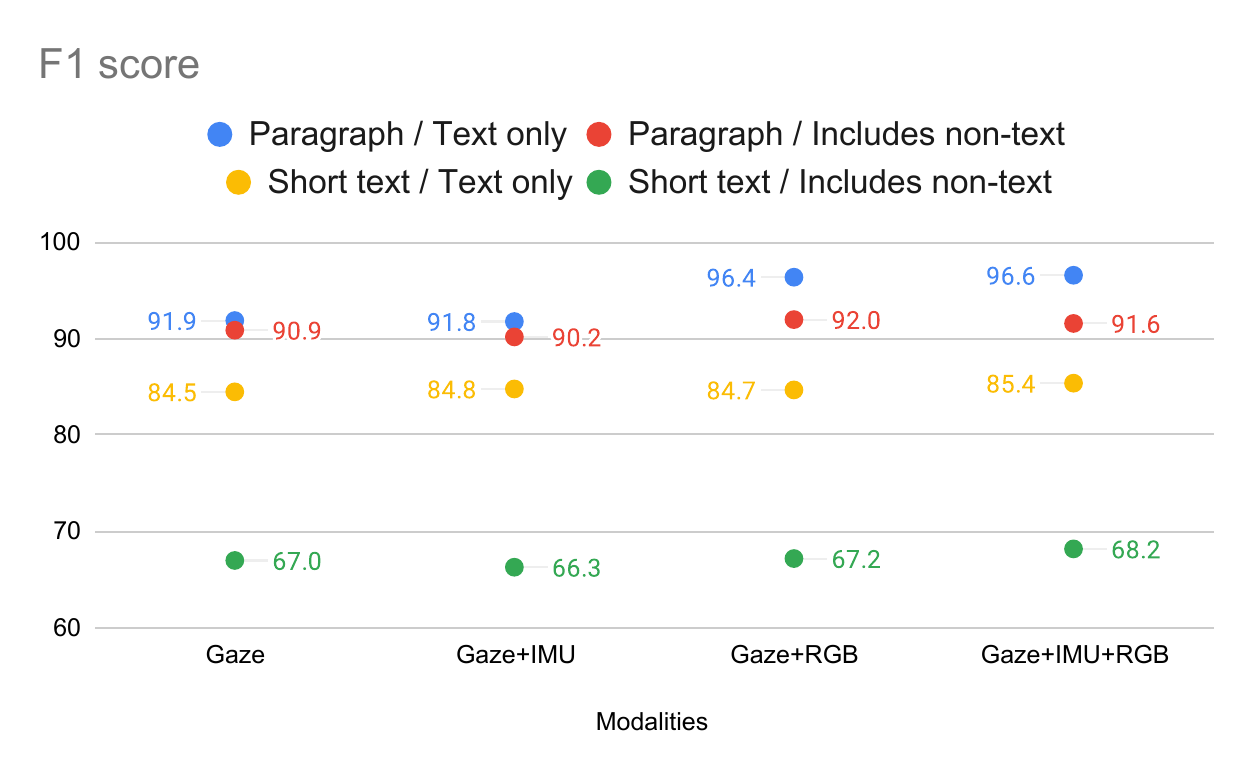}
\caption{\textbf{Breakdown by content length and content type in Columbus subset.} The figures show the Accuracy and F1 scores respectively of different combinations of content length (Paragraph vs Short text) and content type (Text only vs Includes non-text), across different modalities. We show that reading detection works better on paragraphs and text-only cases.}
\label{fig:breakdown_content_length}
\end{figure}

\textbf{Content type and content length.}
Figure~\ref{fig:breakdown_content_length} provides a detailed evaluation of the role of gaze as a primary modality for reading detection and how the addition of RGB and IMU data impacts performance across different content types (samples that have text only vs. those with non-text items) and content lengths (paragraphs vs. short texts). We make several observations:

First, in terms of text length, the performance for paragraphs (blue and red) is significantly better than short texts (yellow and green) across modality combinations and performance metrics. The scan pattern is more irregular in short texts making it harder for gaze data to detect reading behavior.

Second, we notice that text-only models performs better than ones that include non-text elements. We notice that this difference is more pronounced in models using RGB. This is sensible, as non-text elements introduce visual distractions, making it harder for the model to focus on reading cues.

Lastly, we notice the trend consistent with previous observations regarding different modality combinations, with all modalities performing best among all the options.

\begin{figure}[h]
    \centering
    \includegraphics[width=0.48\linewidth]{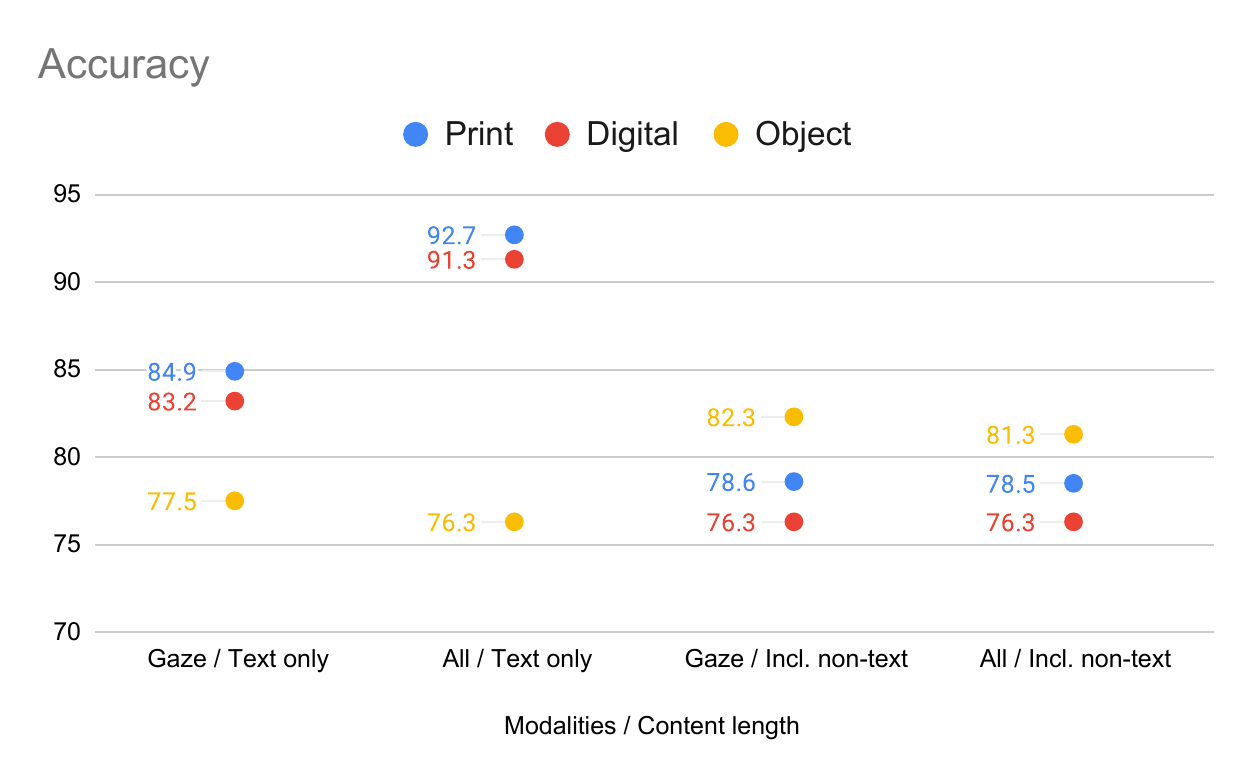}
    \includegraphics[width=0.48\linewidth]{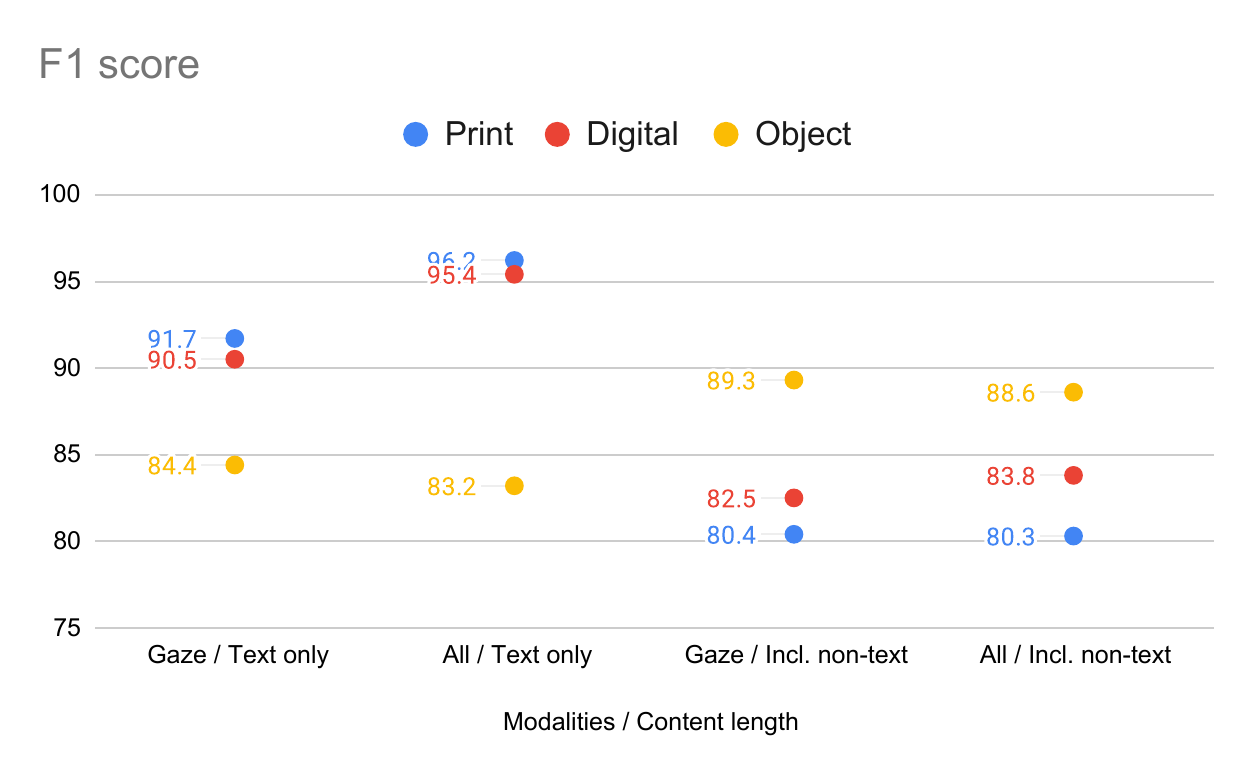}
\caption{\textbf{Breakdown by medium and content type in Columbus subset.} The figures show the Accuracy and F1 scores respectively of different combinations of medium (Print vs Digital vs Object) and content type (Text only vs Includes non-text), across different modalities. We show that reading detection works better on print and digital media.}
\label{fig:breakdown_medium}
\end{figure}

\noindent\textbf{Medium and content type.} Figure~\ref{fig:breakdown_medium} evaluates the performance of gaze as the primary modality and the contribution of RGB and IMU data across three mediums (Digital, Object, and Print) and two content types (Text Only and Contains Non-Text). The results reveal insights into the interaction of modality combinations with medium and content type.

For reading with text only, we notice that the performance for print and digital media are higher than that of object. The inverse is true for reading content that contains non-text. We think that this reflects the nature of the real-world reading content in the training data \emph{i.e.} print and digital media tend to be mostly containing texts, and objects tend to include non-text elements. 

Another interesting observation is that the performance tends to stagnate for both objects and other mediums including non-text despite more modalities. This may be because of the potential misleading cues provided by the RGB stream in these cases. For text-only print and digital media, the performance sees a signficant boost by adding more modalities.




\subsection{Cross-Language}

Although left-to-right (LTR) is the most common writing direction in modern writing systems and is used in popular languages like English, Hindi, Spanish, and Bengali, there are widely used languages that use alternate writing directions like right-to-left script (RTL) used in Arabic where text is written from right to left and top-to-bottom (TTB) used in Traditional Chinese, Japanese and Korean language where text is written top to bottom and right to left. In the Columbus subset, we include Arabic and Traditional Chinese to analyze the performance of reading detection for RTL and TTB using Arabic and Traditional Chinese respectively.

Table~\ref{tab:results_arabic} present the performance of the default model on Arabic text under two conditions: (1) the original gaze sequence and (2) the gaze sequence flipped. The results show the variations in performance based on modality combinations.

\begin{table}[h]
\footnotesize
\begin{minipage}{0.45\textwidth}
\centering
\begin{tblr}{
  cells = {c},
  cell{2}{1} = {r=3}{},
  cell{5}{1} = {r=3}{},
  hline{1-2,5,8} = {-}{},
}

 \textbf{Augment} & \textbf{Gaze} & \textbf{RGB} & \textbf{IMU} & \textbf{F1} & \textbf{Acc} \\
None & \checkmark &  &  & 23.8 & 21.0 \\
 & \checkmark & \checkmark &  & 82.5 & 70.8\\
 & \checkmark & \checkmark & \checkmark & 87.5 & 78.8\\
Flip & \checkmark &  &  & 63.8 & 51.5 \\
 & \checkmark & \checkmark &  & 88.4 & 79.6 \\
 & \checkmark & \checkmark & \checkmark & 91.5 & 85.0 \\
\end{tblr}
\vspace{2mm}
\caption{\textbf{Results on Arabic with and without augmentation.} We show that flipping the gaze horizontally allows right-to-left language to be read better. RGB models show good generalization despite only being trained in English.}
\label{tab:results_arabic}
\end{minipage}
\hfill
\begin{minipage}{0.45\textwidth}
\centering
\begin{tblr}{
  cells = {c},
  cell{2}{1} = {r=3}{},
  cell{5}{1} = {r=3}{},
  hline{1-2,5,8} = {-}{},
}
\textbf{Augment} & \textbf{Gaze} & \textbf{RGB} & \textbf{IMU} & \textbf{F1} & \textbf{Acc} \\
None & \checkmark &  &  & 51.6 & 35.5 \\
 & \checkmark & \checkmark &  & 77.4 & 63.4\\
 & \checkmark & \checkmark & \checkmark & 86.4 & 76.3\\
Rotate & \checkmark &  &  & 91.9 & 85.1 \\
 & \checkmark & \checkmark &  & 94.2 & 89.1 \\
 & \checkmark & \checkmark & \checkmark & 95.9 & 92.3 \\
\end{tblr}
\vspace{2mm}
\caption{\textbf{Results on Traditional Chinese with and without augmentation.} We show that rotating the gaze allows top-to-bottom language to be read better. Again, RGB models show good generalization despite only being trained in English.}
\label{tab:results_chinese}
\end{minipage}
\end{table}

In Table~\ref{tab:results_arabic}, which evaluates the original gaze sequence, gaze as a single modality achieves the lowest performance (F1 = 23.8\%). The performance is drastically affected since Arabic is read from right to left and the model was not trained on such data.
We also evaluate performance with the gaze sequence flipped. This is done by flipping the x-axis input. Here, the accuracy on Gaze only improves to 52.5\%. Other modalities are less affected by the reading direction.
Similarly, models with combined modalities using gaze 
show a noticeable improvement compared to the unaugmented original gaze sequence. This suggests that augmenting the gaze sequence to reflect the directionality of Arabic text better improves the effectiveness of gaze data in multimodal setups. 

Next, Table~\ref{tab:results_chinese} compares model performance on Traditional Chinese text with no augmentation vs when the gaze is rotated. Similar to Arabic, the gaze struggles to read vertical texts without augmentation (though not as much, as some vertical texts do exist even for English, especially in objects).
Still, we notice a significant performance boost when the gaze is rotated. Similarly, rotating the gaze also helps when the gaze is used with other modalities.


\subsection{Mirror Setups}

The performance for Mirror Setups are shown in Table~\ref{tab:results_miiror_setups}. Compared to the entire Columbus subset, we see a significant drop in performance for RGB modality. This is sensible, as this part of the dataset is designed so that the RGB appears difficult to distinguish. The performance of the other two modalities is similar to that of the whole Columbus dataset, as was expected, resulting in a slightly worse multimodal model when modalities are combined.


\begin{table}[h]
\centering
\resizebox{.5\linewidth}{!}{%
\begin{tblr}{
  cells = {c},
  hline{1,9} = {-}{0.08em},
  hline{2} = {-}{0.05em},
  hline{5,8} = {-}{},
}
  & \textbf{Gaze} & \textbf{RGB} & \textbf{IMU} & \textbf{F1} & \textbf{Acc} & \textbf{AUC} & \textbf{P\textsubscript{R=90.0}}\\
 & \checkmark &  &  & 82.4 & 78.9 & 92.7 & 75.7\\
Single &  & \checkmark &  & 69.1 & 66.1 & 82.1 & 67.6\\
 &  &  & \checkmark & 74.5 & 64.6 & 75.8 & 65.5\\
 & \checkmark &  & \checkmark & 80.8 & 76.6 & 91.8 & 71.9\\
Dual &  & \checkmark & \checkmark & 73.6 & 69.2 & 83.5 & 69.0\\
 & \checkmark & \checkmark &  & 80.6 & 77.4 & 92.1 & 73.8\\
All & \checkmark & \checkmark & \checkmark & 80.9 & 77.3 & 91.8 & 72.4
\end{tblr}
}
\vspace{2mm}
\caption{\textbf{Performance of model on mirror setups.} The performance for RGB is significantly lower, while other modalities remain similar.}
\label{tab:results_miiror_setups}
\end{table}

We also analyze the results for specific examples introduced in Section \ref{sec:mirror}.

In Figure~\ref{fig:mirror-comic}, the subject is asked to read a comic in the first setup and then look at the pictures in the comic without reading the second setup. The model accurately detects reading and not reading in both setups across all 3 frames using Gaze only, RGB Only, Gaze and RGB, and all modalities.

Figure~\ref{fig:mirror-corridor} presents a scenario where the subject is asked to read room numbers and signs in a corridor in the first task and then traverse the corridor without reading in the second task. Here, due to the small size of the reading material and the patterns of the eye gaze, none of the modality combinations taken individually or combined can accurately detect reading in any of the 3 frames. However, Gaze only and RGB and Gaze is able to detect not-reading in the second scenario correctly across all 3 frames. RGB only however detects the second scenario as reading across all 3 frames.

Figure~\ref{fig:mirror-search} explores a more fine-grained task differentiation where the subject is asked to read serial numbers on a circuit board versus search and count the number of resistors. Before the sessions subjects were briefed on how to identify resistors on the circuit board and what serial numbers to read. Here, all of Gaze only, RGB Only, Gaze+RGB, and Gaze+RGB+IMU  successfully detect reading in the reading task, however, RGB Only detects not reading on the search task. The other combinations successfully detect reading on the searching task too.

\subsection{Qualitative Results -- Partial Success}
We visualize some samples in the Columbus subset where different combinations succeed or fail. We note some interesting cases here.


\begin{figure}[ht]
    \centering
    \resizebox{\linewidth}{!}{%
    \begin{tabular}{cccccc}
    \includegraphics[width=0.26\linewidth]{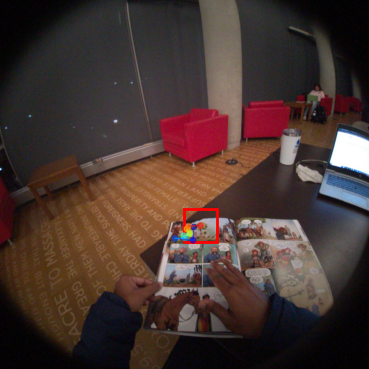} &
    \includegraphics[width=0.26\linewidth]{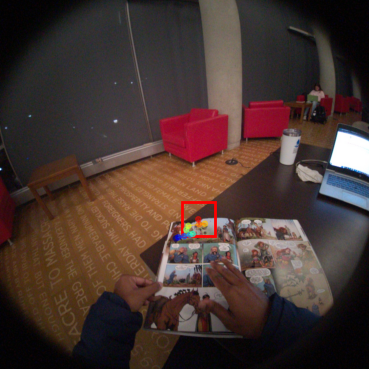} &
    \includegraphics[width=0.26\linewidth]{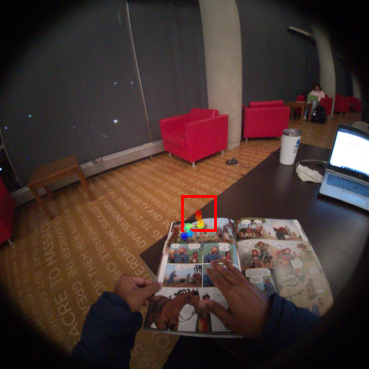} &
    \includegraphics[width=0.26\linewidth]{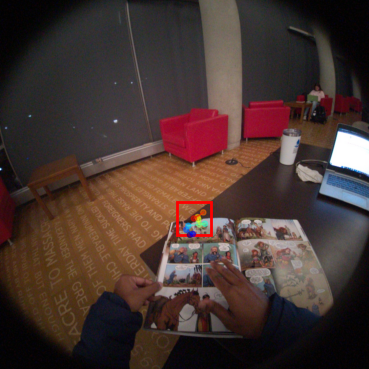} &
    \includegraphics[width=0.26\linewidth]{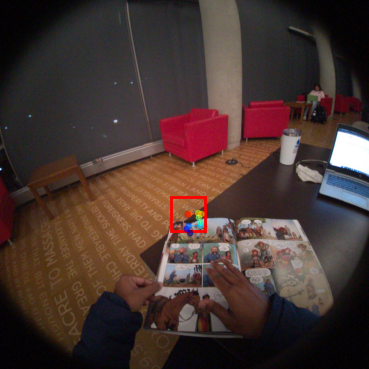} &
    \includegraphics[width=0.26\linewidth]{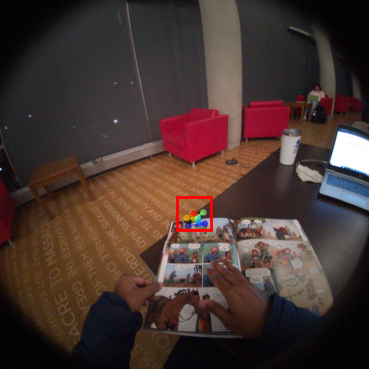} 
    \end{tabular}}
    \caption{\textbf{RGB fails but Gaze succeeds.} Failure case across 6 frames where RGB Fails but Gaze works. Notice that the RGB crop indicated in red has partial coverage of the reading material.}
    \label{fig:fail-rgb}
\end{figure}

In Figure~\ref{fig:fail-rgb}, we observe a case where RGB fails but Gaze succeeds. Here, the RGB crop, marked in red, provides only partial coverage of the reading material, likely leading to misclassification. Despite the suboptimal visual information, Gaze remains effective by directly capturing the subject's attention and focus. This case underscores the robustness of gaze pattern in reading detection.

\begin{figure}[ht]
    \centering
    \resizebox{\linewidth}{!}{%
    \begin{tabular}{cccccc}
    \includegraphics[width=0.26\linewidth]{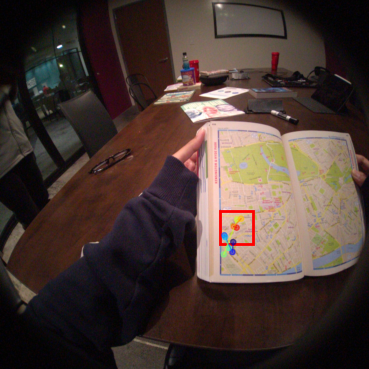} &
    \includegraphics[width=0.26\linewidth]{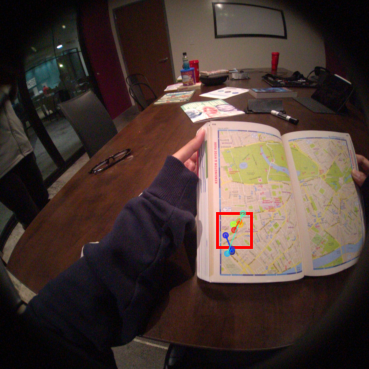} &
    \includegraphics[width=0.26\linewidth]{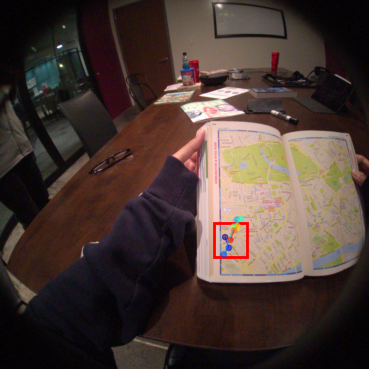} &
    \includegraphics[width=0.26\linewidth]{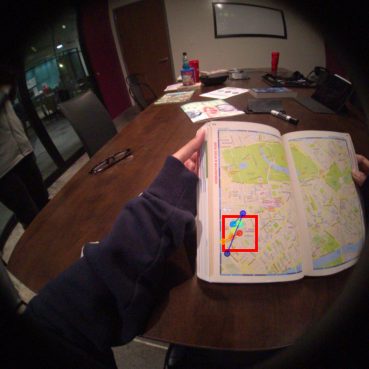} &
    \includegraphics[width=0.26\linewidth]{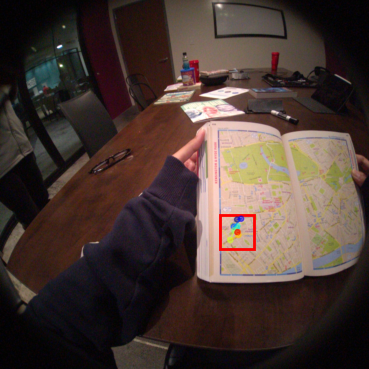} &
    \includegraphics[width=0.26\linewidth]{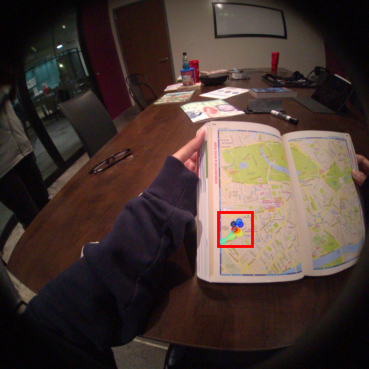}
    \end{tabular}}
    \caption{\textbf{Gaze fails but RGB succeeds.} Here, the subject is reading a map where text is irregularly placed across the field of view, making gaze patterns more sporadic.}
    \label{fig:fail-gaze}
\end{figure}

In contrast, Figure~\ref{fig:fail-gaze} presents a case where gaze fails but RGB works effectively. Here, the subject is reading a map. The subject is asked specifically to read all the names on the map along a particular route. The text being the names of streets and roads along the route does not follow a particular pattern. The irregular placement causes sporadic gaze patterns instead of the horizontal pattern seen in regular reading, making Gaze less reliable for reading detection. RGB, however, can capture the broader visual context of the map and successfully identify the reading behavior. This case along with the previous discussion of Figure~\ref{fig:fail-rgb} highlights the complementary nature of RGB, especially in scenarios with spatially scattered reading content.

\begin{figure}[ht]
    \centering
    \resizebox{\linewidth}{!}{%
    \begin{tabular}{cccccc}
    \includegraphics[width=0.26\linewidth]{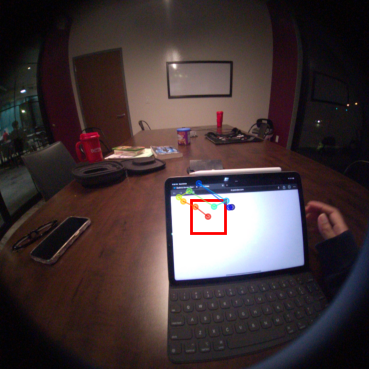} &
    \includegraphics[width=0.26\linewidth]{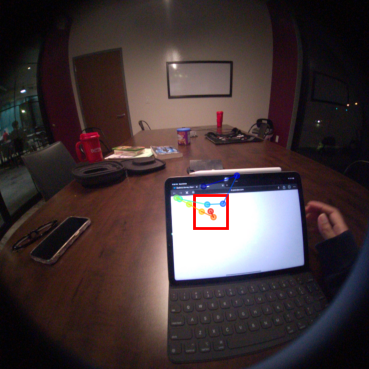} &
    \includegraphics[width=0.26\linewidth]{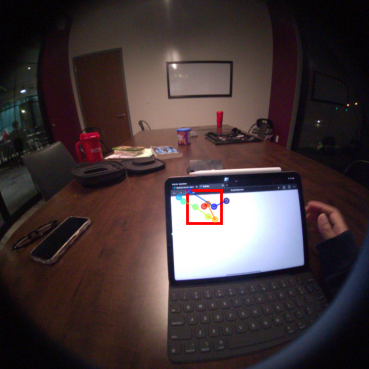} &
    \includegraphics[width=0.26\linewidth]{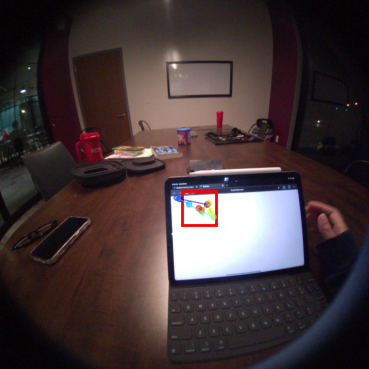} &
    \includegraphics[width=0.26\linewidth]{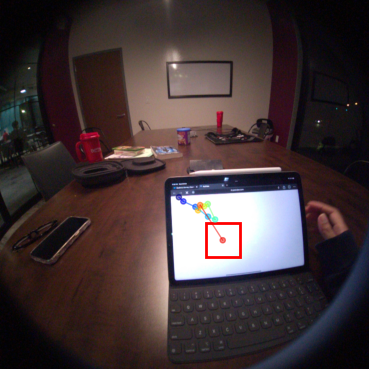} &
    \includegraphics[width=0.26\linewidth]{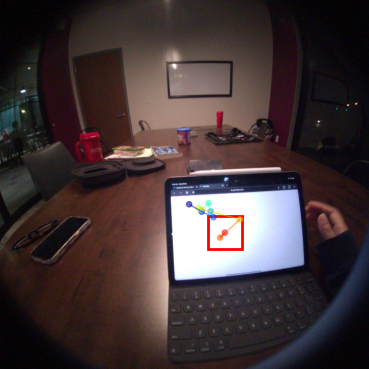}
    \end{tabular}}
    \caption{\textbf{Individual modality fails, combined modality succeeds.} Here, both gaze and RGB fail individually but succeed when combined.}
    \label{fig:gaze-fails-rgb+gaze-works}
\end{figure}

In Figure~\ref{fig:gaze-fails-rgb+gaze-works}, both Gaze and RGB alone fail to detect reading behavior, but their combination succeeds. The multimodal setup leverages gaze to pinpoint attention regions and RGB to provide contextual information, leading to accurate detection. Note that we have found no cases in the Columbus subset where both Gaze and RGB individually fail but Gaze and RGB taken together succeed.

\begin{figure}[ht]
    \centering
    \resizebox{\linewidth}{!}{%
    \begin{tabular}{cccccc}
    \includegraphics[width=0.26\linewidth]{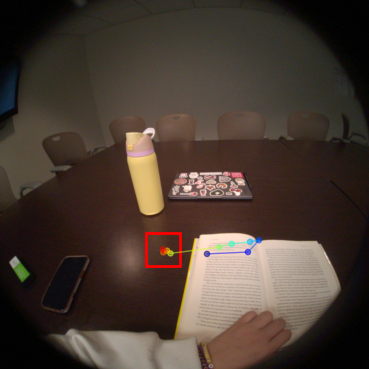} &
    \includegraphics[width=0.26\linewidth]{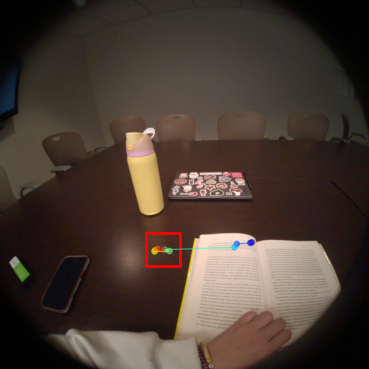} &
    \includegraphics[width=0.26\linewidth]{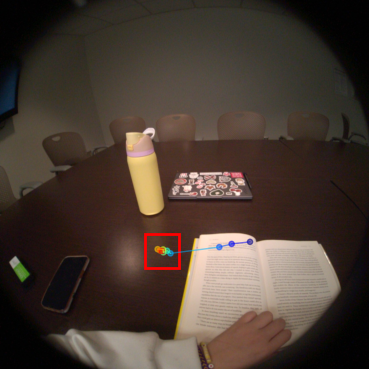} &
    \includegraphics[width=0.26\linewidth]{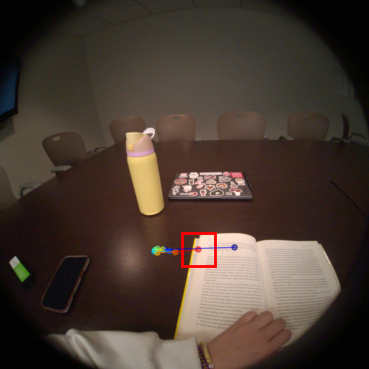} &
    \includegraphics[width=0.26\linewidth]{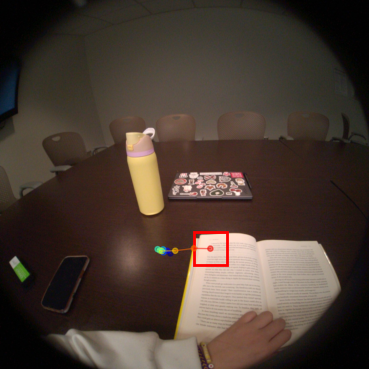} &
    \includegraphics[width=0.26\linewidth]{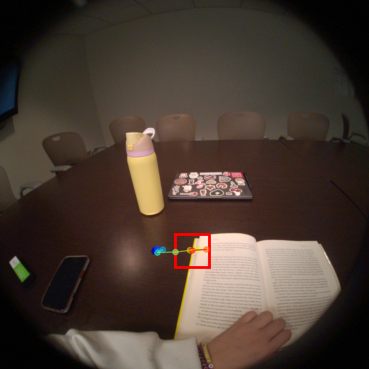}
    \end{tabular}}
    \caption{\textbf{Misleading RGB: Gaze measurement error.} Failure case across 6 frames where Gaze succeeds but RGB Only and Gaze with RGB fail. Note that here the eye gaze is offset due to measurement error, putting the RGB crop outside the reading material.}
    \label{fig:gaze-succeeds-rgb+gaze-fails}
\end{figure}

\begin{figure}[ht]
    \centering
    \resizebox{\linewidth}{!}{%
    \begin{tabular}{cccccc}
    \includegraphics[width=0.26\linewidth]{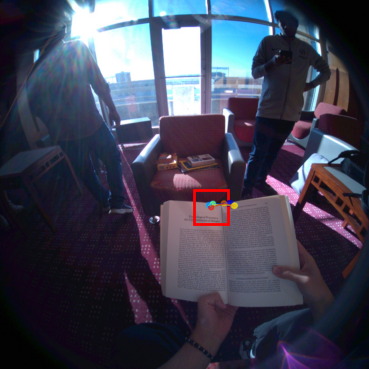} &
    \includegraphics[width=0.26\linewidth]{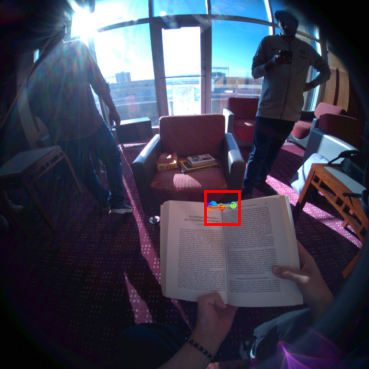} &
    \includegraphics[width=0.26\linewidth]{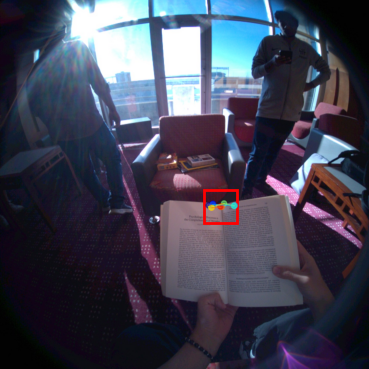} &
    \includegraphics[width=0.26\linewidth]{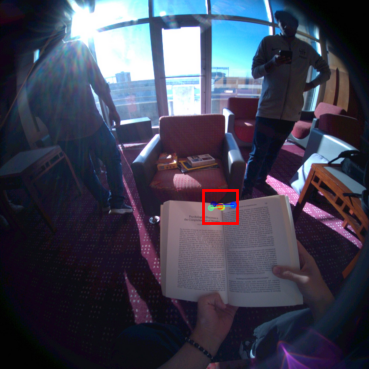} &
    \includegraphics[width=0.26\linewidth]{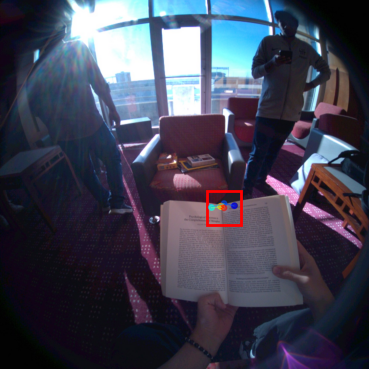} &
    \includegraphics[width=0.26\linewidth]{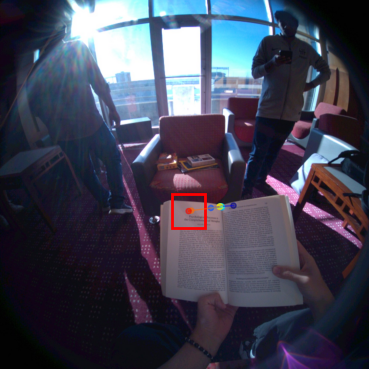}
    \end{tabular}}
    \caption{\textbf{Misleading RGB: Partial coverage.} Failure case across 6 frames where Gaze succeeds but RGB Only and Gaze with RGB fail. Notice that the RGB crop indicated in red has partial coverage of the reading material.}
    \label{fig:gaze-succeeds-rgb+gaze-fails-2}
\end{figure}

In Figure~\ref{fig:fail-rgb}, Figure~\ref{fig:fail-gaze} and Figure~\ref{fig:gaze-fails-rgb+gaze-works}, Gaze+RGB successfully detected reading. In contrast, Figures~\ref{fig:gaze-succeeds-rgb+gaze-fails} and~\ref{fig:gaze-succeeds-rgb+gaze-fails-2} illustrate cases where Gaze succeeds but RGB, either alone or in combination with Gaze, fails. Figure~\ref{fig:gaze-succeeds-rgb+gaze-fails} shows a measurement error where the gaze point is outside the reading material and so the RGB crop provides inaccurate contextual coverage, leading to misclassification. In Figure~\ref{fig:gaze-succeeds-rgb+gaze-fails-2}, the subject is reading at the top edge of a book and so the RGB crop is partially on the reading material. This is similar to the case in Figure~\ref{fig:fail-rgb} except, here, the RGB+Gaze combination fails.

\subsection{Qualitative Results -- Failure Cases}

\begin{figure}[ht]
    \centering
    \resizebox{\linewidth}{!}{%
    \begin{tabular}{cccccc}
    \includegraphics[width=0.26\linewidth]{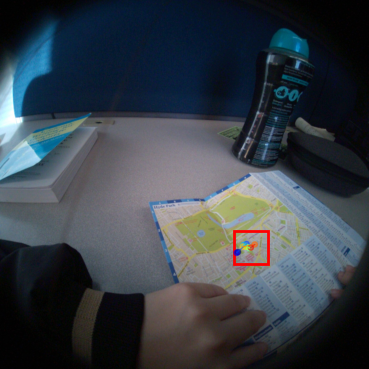} &
    \includegraphics[width=0.26\linewidth]{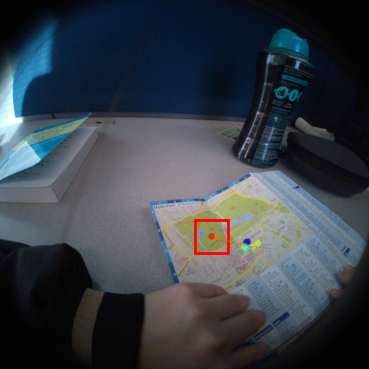} &
    \includegraphics[width=0.26\linewidth]{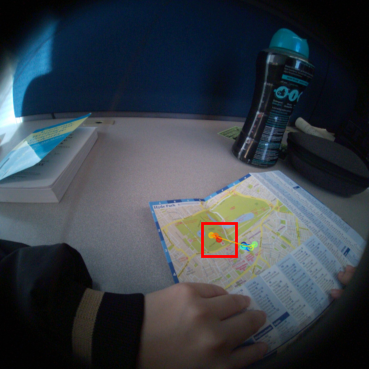} &
    \includegraphics[width=0.26\linewidth]{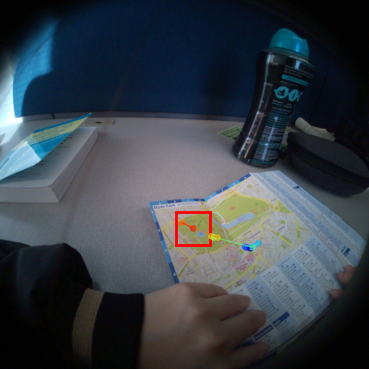} &
    \includegraphics[width=0.26\linewidth]{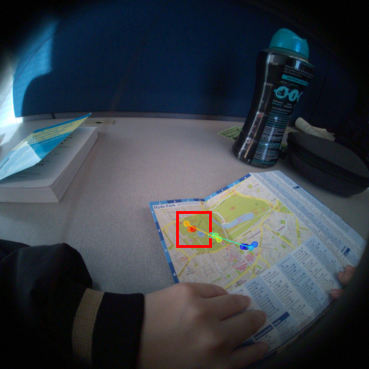} &
    \includegraphics[width=0.26\linewidth]{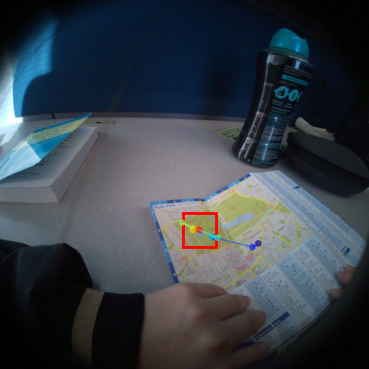}
    \end{tabular}}
    \caption{\textbf{All Modality Failure case: Searching on Map.} Failure case across 6 frames where all modalities and their combinations fail. Here the participant is searching for a particular name on map.}
    \label{fig:all-fail-1}
\end{figure}

\begin{figure}[ht]
    \centering
    \resizebox{\linewidth}{!}{%
    \begin{tabular}{cccccc}
    \includegraphics[width=0.26\linewidth]{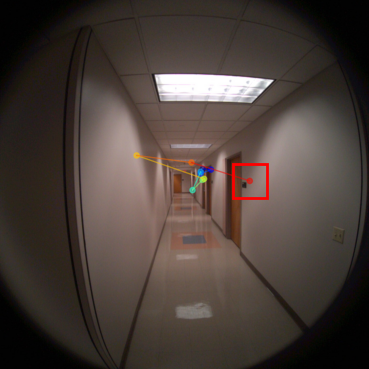} &
    \includegraphics[width=0.26\linewidth]{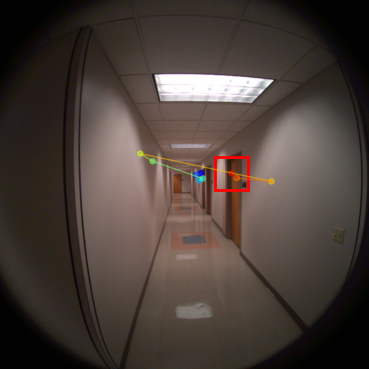} &
    \includegraphics[width=0.26\linewidth]{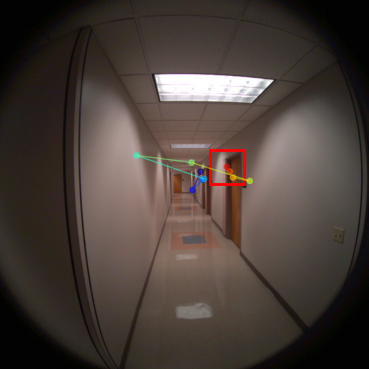} &
    \includegraphics[width=0.26\linewidth]{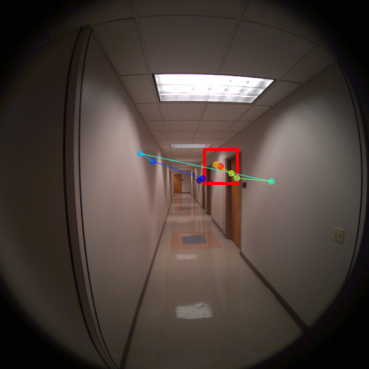} &
    \includegraphics[width=0.26\linewidth]{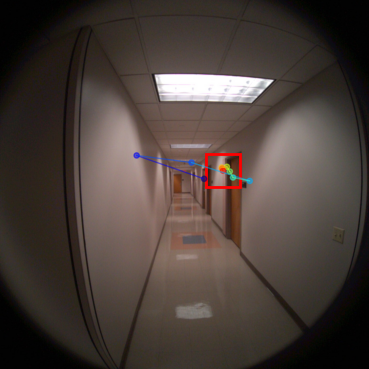} &
    \includegraphics[width=0.26\linewidth]{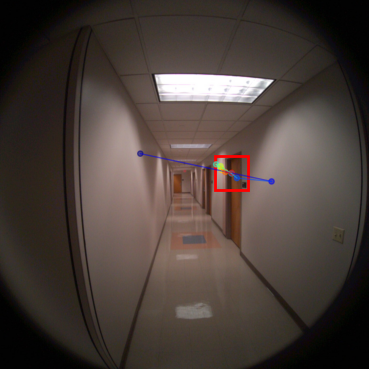}
    \end{tabular}}
    \caption{\textbf{All Modality Failure case: Reading room numbers while walking.} Failure case across 6 frames where all modalities and their combinations fail. Note that here the subject is asked to read room numbers while walking. The gaze pattern scans across the corridor before reading the room number. Note that the model was not trained on this kind of scenario.}
    \label{fig:all-fail-2}
\end{figure}

Figures~\ref{fig:all-fail-1} and~\ref{fig:all-fail-2} depict complete failure cases where all modalities and their combinations fail to detect reading behavior. Figure~\ref{fig:all-fail-1} is an extreme case of Figure~\ref{fig:fail-gaze}, where the subject is searching for a particular location in a map instead of reading individual street or road names along a route. Figure~\ref{fig:all-fail-2} highlights the challenge of detecting reading behavior while walking and reading room numbers. Note that although the model was trained on sequences where the subject was reading something in their hands while walking, it was not trained on reading stationary objects while walking, which potentially adds to the difficulty.

\clearpage
\section{Comparison with Other Methods}
\label{sec:compare}

In this section, we compare our method with other plausible approaches for this task, without considering power and compute constraints for always-on smart glasses. We include a discussion of alternative strategies, although we note that these methods are suboptimal both in terms of practical deployment and in how they handle multimodal inputs. 

\begin{itemize}
    \item \textbf{VLM-based methods}: This approach involves feeding each video frame into a vision-language model (VLM), with a large language model (LLM) backbone. Gaze can be incorporated in several ways such as by overlaying it onto RGB. However, this method is computationally intensive and currently impractical for on-device inference due to limited power and compute resources. Even when the VLM is not run on the device, streaming videos to a server still consumes significant power due to wireless communication cost.
    
    \item \textbf{Action recognition methods}: Reading has traditionally been addressed within the broader domain of action recognition, as discussed in the main paper. However, these models typically depend on sequences of full RGB and optical flow frames, whereas our approach uses a small crop of a single image patch processed by a much smaller model. This reliance on full-frame sequences results in significantly higher power consumption due to the increased sensing cost. 
    
    \item \textbf{Alternative model architectures with gaze-only inputs}: We experiment with a few other alternative model architectures using gaze inputs alone, such as RNNs, CNNs, MLPs. While lightweight and suitable for on-device execution, these methods work with only gaze. This lack of sensor fusion poses challenges for detecting reading short texts on signs and objects or reading text while moving past them.

    \item \textbf{Ours}: We design our method to be simple such that the sensor fusion is seamless, and the computational burden is minimal. It employs low-power sensors, including eye tracking, IMU, and foveated image patches (low-power RGB). Notably, the RGB sensor dominates the sensory cost, while eye tracking and IMU are relatively inexpensive. The lightweight nature of our model allows it to run on-device with minimal computational requirements, while eliminating wireless communication costs. While further engineering could enhance performance, we believe this approach offers a robust and straightforward baseline to highlight the contributions of different sensors.
\end{itemize}

It is important to note that our contribution is not the model architecture itself, which builds up on standard transformer encoders, but in the identification and effective combination of different sensor modalities as inputs. The proposed architecture serves as a minimal, baseline framework for sensor fusion, with the emphasis on simplicity and ease of integration rather than architectural novelty.

While these are \textbf{not practical solutions}, we include the quantitative experimental results in the following subsections for completeness. For these experiments, we use our Reading in the Wild - Columbus subset, labeled as \textbf{RiTW Columbus} for measuring the zero-shot performance. 
In addition, we also evaluate on the EGTEA~\cite{li2018eye} dataset for the models with limited zero-shot capabilities. We use the dataset's first training and test split. It is important to note that, \textbf{unlike our dataset, the EGTEA dataset only contains examples of reading long paragraphs from cooking recipes and lacks instances of reading short texts or hard negatives}, where text is present but not being read. This makes it easier for the existing models, as using RGB data alone is sufficient for most cases.

\begin{table}[h]
\setlength{\tabcolsep}{2.5pt}
\centering
\footnotesize
\begin{tabular}{l|cccc} \toprule
\multirow{ 2}{*}{\textbf{Method}}    & \multirow{ 2}{*}{VLM~\cite{grattafiori2024llama}} & \multirow{ 2}{*}{Action recognition~\cite{carreira2017quo}} & Alternative arch. & \multirow{ 2}{*}{\textbf{Ours}} \\ 
   &   &  & with gaze only &  \\ \midrule
\textbf{1. Enabled modalities} & & & & \\ 
Gaze                            & \xmark         & \xmark      & \checkmark           & \checkmark           \\
RGB                             & \checkmark     & \checkmark  & \xmark               & \checkmark           \\
IMU                             & \xmark         & \xmark      & \xmark               & \checkmark           \\
Fusion                          & \checkmark & \xmark      & \xmark               & \checkmark           \\ \midrule
\textbf{2. On-device Feasibility}       & \xmark         & \xmark      & \checkmark           & \checkmark           \\
Number of parameters                 & 11B            & 25M         & 1k                   & 130k                 \\
Sensing cost (power) & high                       & high               & low                           & low  \\
\hspace{3mm} RGB requirements & \multirow{ 2}{*}{full RGB }                        & \multirow{ 2}{*}{full RGB video}                 & \multirow{ 2}{*}{-}                          & foveated patch (5\textdegree~ FoV) \\
\hspace{3mm} (dominates sensing cost) &                          &                &                           & (optional) \\
Real-time                 & \xmark            & \xmark       & \checkmark                   & \checkmark                 \\
\hspace{3mm} Inference time (ms)               & 567.410  & 895.511 (incl. flow)     &      0.310    &   0.545 \\           
 \midrule
\textbf{3. Performance}    & & & \\ 
Zero-shot capability & \checkmark & \xmark & \xmark & \checkmark  \\
Acc / F1 on RiTW Columbus                & 76.7 / 65.6   & - & - & \textbf{82.9 / 88.8} \\ 
Acc / F1 on EGTEA dataset                & 89.6 / 61.5    & 88.8 / 65.8 & 85.8 / 62.8 & \textbf{89.6 / 70.6} \\ 

\bottomrule
\end{tabular} \vspace{5pt}
\caption{
\textbf{Comparison of alternative methods.} This table compares approaches for reading recognition, including (i) vision-language models (VLMs), (ii) action recognition models, and (iii) alternative architectures such as RNNs. Although all methods achieve reasonable performance on EGTEA (last row), these results are likely overestimated due to limited dataset diversity and the absence of hard negatives, leading to poor zero-shot generalization. Power and compute constraints of always-on smart glasses are not considered here—many of these models are impractical for on-device use. The high inference time of the action recognition model~\cite{carreira2017quo} is partly due to its reliance on optical flow computation.}
\label{tab:supp_alternative}
\end{table}

\subsection{VLM based methods (Gaze + RGB)}

To incorporate gaze into the input image, we simply overlay the gaze scanpath over the RGB image. We experiment using these settings.

\begin{itemize}
    \item VLM Model: meta-llama/Llama-3.2-11B-Vision-Instruct~\cite{grattafiori2024llama}
    \item Experiment: Given an image with 2s eye gaze trajectory overlaid on top, prompt llm to determine if reading or not. We fed images with the full image resolution of 640 x 480 to the VLM.
    \item Prompt: This image shows eye gaze trajectory represented by connected green lines and a red circle, indicating where a person was looking when the image was captured. Determine if the person is reading or not reading. Answer only with 'reading' or 'not reading'.
\end{itemize}

\begin{figure}[htbp]
    \centering
    \begin{minipage}{0.45\textwidth}
        \centering
        \includegraphics[width=\textwidth]{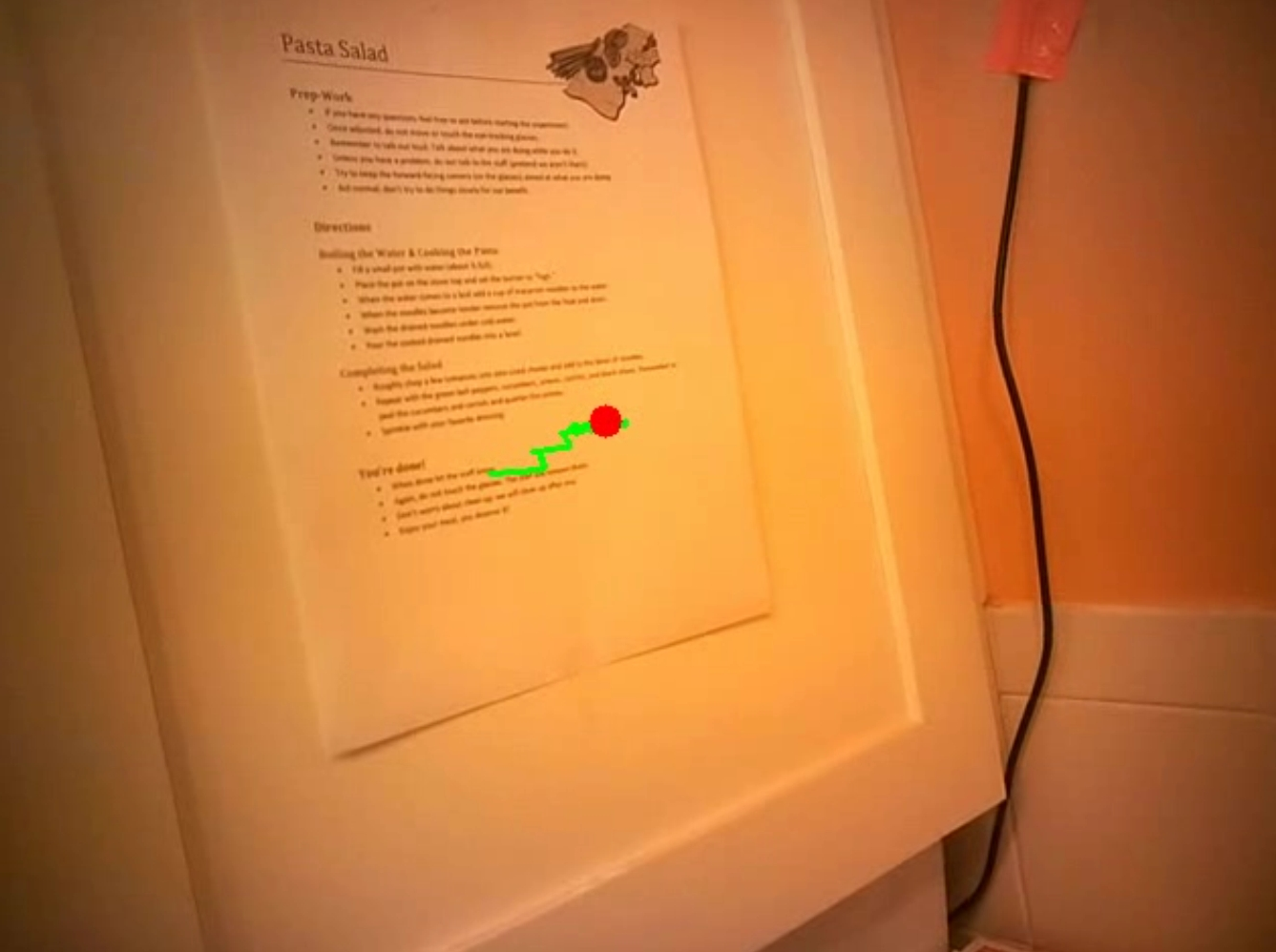}
        \caption{Reading}
        \label{fig:reading}
    \end{minipage}\hfill
    \begin{minipage}{0.45\textwidth}
        \centering
        \includegraphics[width=\textwidth]{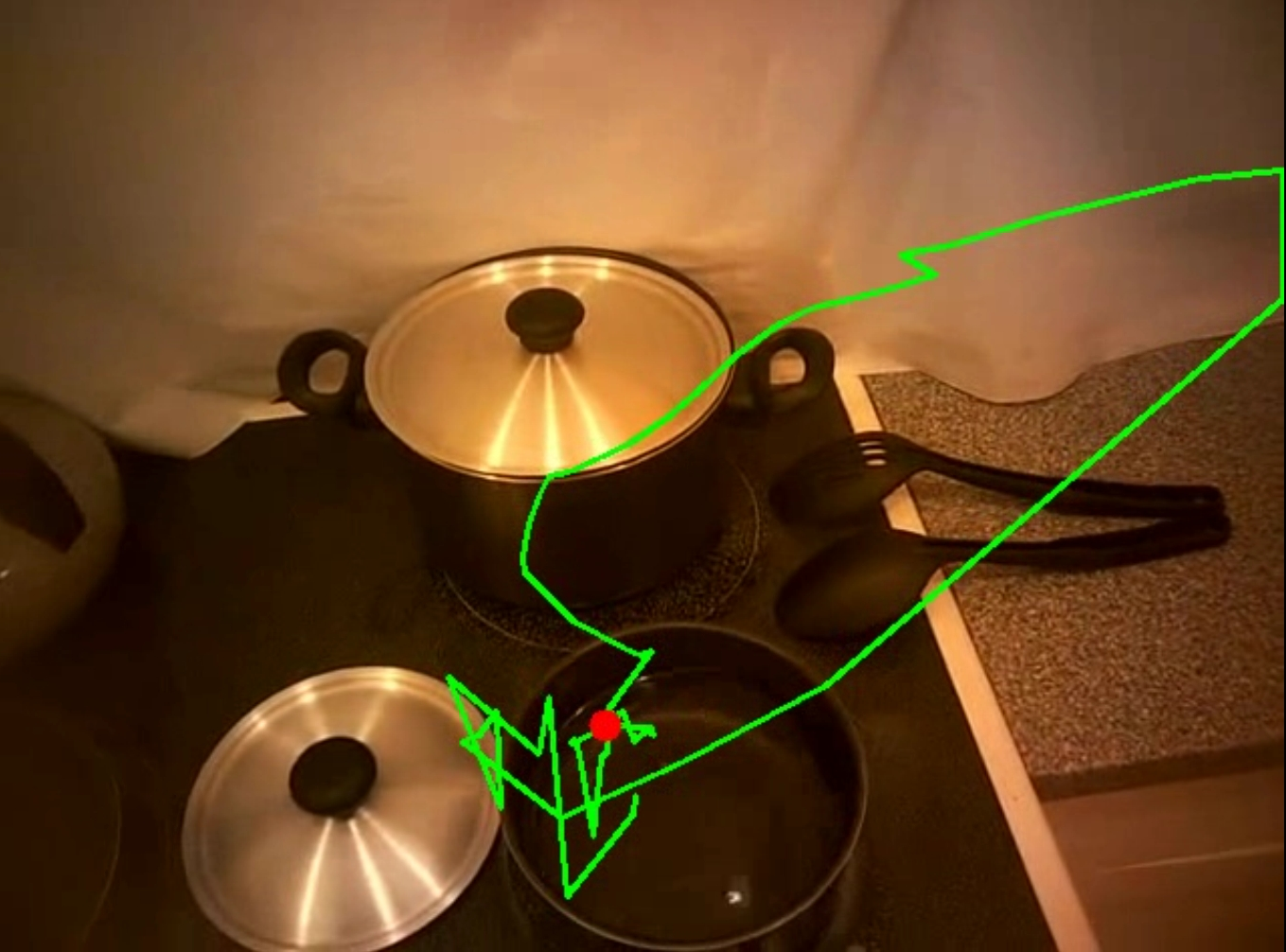}
        \caption{Not Reading}
        \label{fig:not_reading}
    \end{minipage}
    \caption{Example image inputs to the VLM (Llama-3.2-11B-Vision-Instruct~\cite{grattafiori2024llama})}
    \label{fig:vlm_examples}
\end{figure}

Our method outperforms the VLM approach on both the Reading in the Wild Columbus subset and the EGTEA dataset. While the results are decent, we note that this is an impractical solution in terms of being able to run on-device at all times.

We further experiment with the settings to match the settings used in our model (e.g. RGB crop and gaze only) on the EGTEA dataset.
\begin{itemize}
    \item RGB crop and gaze overlay on blank image: We use the same crop size and overlay the gaze onto a blank image to mimic our Gaze + RGB setting. The results are: F1 Score: 0.2648, Accuracy: 0.4622
\item Gaze overlay on blank image: We overlay the gaze pattern onto an empty image to mimic gaze-only settings. The results are: F1 Score: 0.2480, Accuracy: 0.2564.
\end{itemize}
Evidently, the VLM has not seen such settings before and are unable to generalize zero-shot onto such prompts, which is not unexpected.

\subsection{Action recognition based methods (RGB only)}

To establish a baseline using action recognition method, we use the base two-stream I3D network \cite{carreira2017quo}. The network has 25M parameters, is pretrained on Kinetics and fine-tuned on EGTEA (training set). We use both RGB and optical flow streams at full frames as input. Similar to before, we treat the class "Inspect/read recipe" as positive and the rest as negative. 

The results are similar on the EGTEA dataset, but we again note that this is not a practical solution towards our use case which is real-time, on-device processing given the input and compute requirements. Action recognition models are large (both in terms of input requirements and parameters), and computing optical flow adds to latency. We further note that there are some works that try to incorporate gaze into action recognition models \cite{min2021integrating}, but this further adds compute and goes further off tangent to our practical goal.

\subsection{Alternative model architectures with gaze-only inputs (Gaze only)}

Lastly, we also experiment with a few alternative model architectures with gaze inputs only, which we train on the EGTEA training split, including:

\begin{itemize}
    \item \textbf{RNN.} First, we treat this as time series classification, and us a single layer RNN by treating gaze as time series. We use a single-layer GRU (32 channel dimensions, 1k parameters) followed by a linear layer to perform binary classification and train it with cross-entropy loss. We optimize using Adam with learning rate 0.003. To handle class imbalance during training we subsample the negative examples so that the training set is roughly uniform. The results are reasonable, as shown in Table \ref{tab:supp_alternative}.
    \item \textbf{1D CNN.} We also experiment with using a 1D CNN (3 layers of convolutions each of 32 channel dimensions, followed by a linear layer; 11k params total). This setup is similar to our model with gaze only input, but without the transformer layers. We find it to yield similar results (85.7 / 65.0 on EGTEA).
    \item \textbf{MLP.} Finally, we experimented with a 4-layer MLP with fully connected layers. This model resulted in overfitting.
\end{itemize}

We note that, while these methods may work on-device, the usefulness of gaze-only methods can be rather limited in cases where other modalities are more useful, such as reading short texts on signs and objects, or reading a text while passing by. In these cases, these methods are unable to naturally incorporate other sensors (RGB, IMU).

\clearpage

\clearpage
\section{Discussion}
\label{sec:discussion}

\subsection{Reading medium classification using RGB on Seattle Subset}
We also present the results for reading medium classification using the Gaze+IMU+RGB (10° FoV) model, as a pseudo upper bound. Notice that other mediums hardly get misclassified as digital, but there is still some confusion due to (i) small model (ii) gaze inaccuracy.

\begin{table}[h]
\centering
\footnotesize
\begin{tabular}{lcccc}
            & (1)                          & (2)                          & (3)                          & (4)                          \\
(1) No read & \cellcolor[HTML]{6CC499}0.88 & \cellcolor[HTML]{FAFDFC}0.03 & \cellcolor[HTML]{FCFEFD}0.02 & \cellcolor[HTML]{F4FBF7}0.07 \\
(2) Print   & \cellcolor[HTML]{F9FDFB}0.04 & \cellcolor[HTML]{83CDA9}0.74 & \cellcolor[HTML]{FEFFFE}0.01 & \cellcolor[HTML]{DCF1E7}0.21 \\
(3) Digital & \cellcolor[HTML]{F7FCFA}0.05 & \cellcolor[HTML]{EDF8F3}0.11 & \cellcolor[HTML]{85CEAA}0.73 & \cellcolor[HTML]{EDF8F3}0.11 \\
(4) Objects & \cellcolor[HTML]{F7FCFA}0.05 & \cellcolor[HTML]{CAEADA}0.32 & \cellcolor[HTML]{FEFFFE}0.01 & \cellcolor[HTML]{97D5B7}0.62
\end{tabular}
\vspace{2mm}
\caption{\textbf{Reading medium classification} using RGB.}
\end{table}

\subsection{Limitations and Future Work}

\textbf{Generalization and personalization}. While our aim is to build a generalizable model that can capture the way people read, it does not take away the fact that different people read in different ways. Most prominently, we notice the reading speed varies greatly with language and the reader's fluency, and the model sometimes fails when the reading becomes too slow.

For example, the average gaze velocity in the case of Chinese script is about two-thirds that of English. We did some simple personalization augmentations by simply scaling all velocity values to match the English case, and notice some improvements. This is a promising first step towards not only building reading detection models but also one that can be optimized towards a user.

\noindent\textbf{Exploring eye tracking sensors.} We have identified three modalities (gaze, RGB, and head pose), though we think that with eye tracking sensors gaze is not the only available information. Particularly, we also know that (i) the pupils dilate more (ii) people blink less when they are reading, and incorporating these cues may be interesting in the future.

\noindent\textbf{Deployment and efficiency.} We have shown that our model performs best by using all three combined modalities, but the model using gaze or RGB alone is also able to perform to good accuracy. In practical scenarios in wearable devices, power constraints sometimes restrict the use of all three modalities at all times. An interesting question here would be to know when to turn on/off each modality to maximize overall efficiency while still maintaining performance. 

\noindent\textbf{Gaze and human perception.} Reading is one of the subset of activities where the eye gaze pattern clearly associates with the activity. More generally, it will be interesting to expand the questions we asked for reading towards more general human perception and ask: what does it mean to `look' at something?

Imagine a scenario in which a person looks at a painting. The gaze can tell whether or not a person is looking at the painting, but is the person just glancing over it, or are they inspecting it to great detail? What aspect of the painting is the person looking at (color, shape, texture)? Are they examining critically or in awe, or just staring into blank space in introspection? These are questions that simply projecting where the eye is looking at onto the image does not yield a satisfactory solution, and will be interesting to investigate in the future.

\end{document}